\documentclass[twocolumn,runningheads]{svjour3}

\usepackage[latin1]{inputenc}
\usepackage{times}

\usepackage{psfrag,amsmath, amssymb, amscd, amsfonts, latexsym, epsf, graphicx,amscd}

\def\bbbr{{\mathbb R}}

\newcommand{\orth}{\bot}

\journalname{arXiv preprint}

\begin{document}

\titlerunning{Provably scale-covariant continuous hierarchical networks}
\title{\bf Provably scale-covariant continuous hierarchical networks
  based on scale-normalized differential expressions coupled in cascade%
\thanks{In: Journal of Mathematical Imaging and Vision,
  doi:10.1007/s10851-019-00915-x. The support from the Swedish Research Council 
              (contract 2018-03586) is gratefully acknowledged. }}
\author{Tony Lindeberg}

\institute{Computational Brain Science Lab,
        Division of Computational Science and Technology,
        KTH Royal Institute of Technology,
        SE-100 44 Stockholm, Sweden.
       \email{tony@kth.se}}

\date{}

\maketitle

\begin{abstract}
\noindent
This article presents a theory for constructing hierarchical networks in
such a way that the networks are guaranteed to be provably scale covariant.
We first present a general sufficiency argument for obtaining scale
covariance, which holds for a wide class of networks defined from linear and non-linear
differential expressions expressed in terms of scale-normalized
scale-space derivatives.
Then, we present a more detailed development of one example of such a
network constructed from  
a combination of mathematically derived models of receptive
fields and biologically inspired computations. 
Based on a functional model of complex cells in terms of an oriented
quasi quadrature combination of first- and second-order directional 
Gaussian derivatives, we couple such primitive computations in cascade
over combinatorial expansions over image orientations.
Scale-space properties of the computational primitives are analysed
and we give explicit proofs of how the resulting representation allows
for scale and rotation covariance. 
A prototype application to texture analysis is developed
and it is demonstrated that a simplified mean-reduced representation of the
resulting QuasiQuadNet leads to promising
experimental results on three texture datasets.

\keywords{Hand-crafted \and Structured \and Deep \and Hierarchical \and Network \and Scale
  covariance \and Scale invariance \and Differential
  expression \and Quasi quadrature \and Complex cell
                  \and Feature detection \and Texture analysis \and
                  Texture classification \and
                  Scale space \and Deep learning \and Computer vision}

\end{abstract}

\section{Introduction}
\label{sec-intro}

The recent progress with deep learning architectures 
\cite{KriSutHin12-NIPS,SimZis15-ICLR,LeCBenHin15-Nature,SzeLiuJiaSerReeAngErhVanRab15-CVPR,HeZhaRenSun16-CVPR,RedDivGirFar16-CVPR,HuaLiyMaaWei17-CVPR,XieGIrDolTuHe17-CVPR,RenHeGirSun17-PAMI,SabFroHin17-NIPS}
has demonstrated that hierarchical feature representations over multiple layers have higher potential compared to
approaches based on single layers of receptive fields.

Although theoretical and empirical advances are being made 
\cite{CohShaSha15-ConfLearnTheor,MahVed15-CVPR,TisZas15-ITW,Mal16-RoySoc,LinTegRol17-StatPhys,VidBruGirSoa17-arXiv,WiaBol18-IT,GolBerGreMelNguKinPol18-arXiv},
we currently lack a comparable understanding of the non-linearities in
deep networks in the way that scale-space theory provides a deep
understanding of early visual receptive fields.
Training deep networks is still very much of an art \cite{AchRovSoa17-arXiv}.
Moreover, deep nets sometimes perform serious errors.
The observed problem with adversarial examples
\cite{SzeZarSutBruErhGooFer13-arXiv,NguYoClu15-CVPR,TanGri16-arXiv,AthSut17-arXiv,SuVarKou17-arXiv,MooFawFawFro17-CVPR,BakLuErlKel-CompBiol}
can be taken as an indication that current deep nets may not solve the
same type of problem as one might at first expect them to do.
For these reasons, it is of interest to develop theoretically
principled approaches to capture non-linear hierarchical relations
between image structures at different scales as an extension of the
regular scale-space paradigm.

A specific limitation of current deep nets is that they are not truly scale covariant.
A deep network constructed by repeated application of compact 
$3 \times 3$ or $5 \times 5$ kernels, such as AlexNet \cite{KriSutHin12-NIPS},
VGG-Net \cite{SimZis15-ICLR} or
ResNet \cite{HeZhaRenSun16-CVPR},
implies an implicit assumption of a preferred size in the image
domain as induced by the discretization in terms of local $3 \times 3$ 
or $5 \times 5$ kernels of a fixed size.
Spatial max pooling over image neighbourhoods of fixed size, such as
over $2 \times 2$ neighbourhoods over multiple layers, also implies that
non-linearities are applied relative to a fixed grid spacing.
Thereby, due to the non-linearities in the deep net, the output from
the network may be qualitatively different
depending on the specific size of the object in the image domain,
as varying because of {\em e.g.\/}\ different distances between the
object and the observer.
To handle this lack of scale covariance, approaches have been developed such as
spatial transformer networks \cite{JadSimZisKav15-NIPS}, using sets of
subnetworks in a multi-scale fashion \cite{CaiFanFerVas16-ECCV} or by combining deep nets with image
pyramids \cite{LinDolGirHeHarBel17-CVPR}.
Since the size normalization performed by a spatial transformer
network is not guaranteed to be truly scale covariant, and since
traditional image pyramids imply a loss of image information
that can be interpreted as corresponding to undersampling,
it is of interest to
develop continuous approaches for deep networks that guarantee true scale covariance or better
approximations thereof.

An argument that we want to put forward in this article is that truly
scale-covariant deep networks with their associated extended notion of
truly scale-invariant networks may be conceptually much easier to achieve if we
set aside the issues of spatial sampling in the first modelling stage and
model the transformations between adjacent layers in the deep network as continuous translation-covariant 
operators as opposed to discrete filters.
Specifically, we will propose to combine concepts from hierarchical
families of CNNs with
scale-space theory to define continuous families of hierarchical networks, with 
each member of the family being a rescaled copy of the base network, 
in a corresponding way as an input image is embedded into a
one-parameter family of images, with scale as the parameter, within the regular scale-space framework.
Then, a structural advantage of a 
continuous model as compared to a discrete
model is that it can guarantee provable scale covariance in the following way:
If the computational primitives that are used for defining a hierarchical
network are defined in a multi-scale manner, {\em e.g.\/} from Gaussian
derivatives and possibly non-linear differential expressions
constructed from these, and if the scale
parameters of the primitives in the higher layers are proportional to the scale parameter
in the first layer, then if we define a multi-scale hierarchical
network over all the scale parameters in the first layer, the multi-scale
network is guaranteed to be truly scale covariant. 

This situation is in
contrast to the way most deep nets are currently constructed, as a
combination of discrete primitives whose scales are instead proportional to the
grid spacing. That in turn implies a preferred scale of the computations and
which will violate scale covariance unless the image data are
resampled to multiple rescaled copies of the input image prior to being used as input to a deep net. If using such
spatial resampling to different levels of resolution, then, however, 
it may be harder to combine information between
different multi-scale channels compared to using a continuous model
that preserves the same spatial sampling in the input data.
Rescaling of the image data prior to later stage processing may also introduce sampling artefacts.

The subject of this article is to first present a general sufficiency
argument for constructing provably scale-covariant hierarchical
networks based on a spatially continuous model of the transformations
between adjacent layers in the hierarchy. This sufficiency result
holds for a very wide class of possible continuous hierarchical networks. Then, we will develop in more
detail one example of such a continuous network for capturing
non-linear hierarchical relations between features over multiple
scales. 

Building upon axiomatic modelling of visual
receptive fields in terms of Gaussian derivatives and affine
extensions thereof, which can serve as idealized models of simple
cells in the primary visual cortex
\cite{KoeDoo92-PAMI,Lin10-JMIV,Lin13-BICY,Lin13-PONE},
we will propose a functional model for complex cells in terms 
of an oriented quasi quadrature measure, which combines first- and
second-order directional affine Gaussian derivatives according to an energy
model \cite{AdeBer85-JOSA,Hee92-VisNeuroSci,Lin97-IJCV,Lin18-SIIMS}.
Compared to earlier approaches of related types
\cite{Fuk80-BICY,RiePog99-Nature,SerWolBilRiePog07-PAMI,BruMal13-PAMI,PogAns16-book},
our quasi quadrature model has the conceptual advantage that
it is expressed in terms of scale-space theory in addition to well reproducing
properties of complex cells as reported by
\cite{AdeBer85-JOSA,TouFelDan05-Neuron,CarHee12-NatureRevNeuroSci,GorSimMov15-Neuron}.
Thereby, this functional model of complex cells allows for a conceptually easy integration with
transformation properties, specifically truly provable scale covariance,
or a generalization to affine covariance provided that the receptive field responses are
computed in terms of affine Gaussian derivatives as opposed to regular
Gaussian derivatives.

Then, we will combine such oriented quasi quadrature measures in cascade, building upon
the early idea of Fuku\-shima \cite{Fuk80-BICY} of using  
Hubel and Wiesel's findings regarding
receptive fields in the primary visual cortex \cite{HubWie59-Phys,HubWie62-Phys,HubWie05-book}
to build a hierarchical neural network from repeated application
of models of simple and complex cells.
This will result in a hand-crafted network, termed quasi quadrature network, with
structural similarities to the scattering network proposed by Bruna and Mallat
\cite{BruMal13-PAMI}, although expressed in terms of Gaussian
derivatives instead of Morlet wavelets.

We will show how the scale-space properties of the quasi
quadrature primitive in this representation can be theoretically
analysed and how the resulting hand-crafted network becomes provably
scale covariant and rotation covariant, in such a way that the multi-scale and
multi-orientation network commutes with scaling transformations and
rotations in the spatial image domain. 

As a proof of concept that the proposed methodology can lead to
meaningful results, we will experimentally investigate a prototype application
to texture classification based on a substantially simplified
representation that uses just the average values over image space of
the resulting QuasiQuadNet. It will be demonstrated that the resulting
approach leads to competitive results compared to classical texture
descriptors as well as to other hand-crafted networks.

Specifically, we will demonstrate that in the presence of substantial
scaling transformations between the training data and the test data,
true scale covariance substantially improves the ability to perform
predictions or generalizations beyond the variabilities that are spanned by the training data.

\subsection{Structure of this article}

Section~\ref{sec-rel-work} begins with an overview of related work, with
emphasis on related scale-space approaches, deep learning approaches
somehow related to scale, rotation-covariant deep networks, biologically
inspired networks, other hand-crafted or structured networks including other hybrid
approaches between scale space and deep learning.

As a general motivation for studying hierarchical networks that are based on primitives that
are continuous over image space, 
Section~\ref{sec-sc-cov-cont-hier-nets} then presents a general
sufficiency argument that guarantees provable scale covariance for a very
wide class of networks defined from layers of scale-space operations
coupled in cascade.

To provide an additional theoretical basis for a subclass of such
networks that we shall study in more detail in this article, based on
functional models of complex cells coupled in cascade,
Section~\ref{sec-quasi-quad-1D} describes a quasi quadrature measure
over a purely 1-D signal, which measures the energy of first- and
second-order Gaussian derivative responses.
Theoretical properties of this entity are analysed with regard to
scale selectivity and scale selection properties, and we show how free
parameters in the quasi quadrature measure can be determined from
closed-form calculations.

In Section~\ref{sec-ori-quasi-quad-measure}, an oriented extension of
the 1-D quasi quadrature measure is presented over multiple orientations in
image space and is proposed as a functional model that mimics some of
the known properties of complex cells, while at the same time being
based on axiomatically derived affine Gaussian derivatives that well
model the functional properties of simple cells in the primary visual cortex.

In Section~\ref{sec-hier-quasi-quad-net} we propose to couple such quasi quadrature
measures in cascade, leading to a class of hierarchical networks based
on scale-space operations that we term quasi quadrature networks.
We give explicit proofs of scale covariance and rotational covariance
of such networks, and show examples of the type of information that
can be captured in different layers in the hierarchies.

Section~\ref{sec-appl-texture-anal} then outlines a prototype application to
texture analysis based on a substantially mean-reduced version of such
a quasi quadrature network, with the feature maps in the different
layers reduced to just their mean values over image space.
By experiments on three datasets for texture classification, we show
that this approach leads to promising results that are comparable or better
than other hand-crafted networks or more
dedicated hand-crafted texture descriptors.
We do also present experiments of scale prediction or scale
generalization, which quantify the performance over scaling
transformations for which the variabilities in the testing data are
not spanned by corresponding variabilities in the training data.

Finally, Section~\ref{sec-summ-disc} concludes with a summary and discussion.

\subsection{Relations to previous contribution}

This paper constitutes a substantially extended version of a
conference paper presented at the 
SSVM 2019 conference \cite{Lin19-SSVM} and with substantial additions
concerning:
\begin{itemize}
\item
  the motivations underlying the developments of this work
  and the importance of scale covariance for deep networks 
  (Section~\ref{sec-intro}),
\item
  a wider overview of related work (Section~\ref{sec-rel-work}),
\item
  the formulation of a general sufficiency result to guarantee scale
  covariance of hierarchical networks constructed from computational
  primitives (linear and non-linear filters) formulated based on 
  scale-space theory (Section~\ref{sec-sc-cov-cont-hier-nets}),
\item
  additional explanations regarding the quasi quadrature measure
  (Section~\ref{sec-quasi-quad-1D}) and its oriented affine extension
  to model functional properties of complex cells
  (Section~\ref{sec-ori-quasi-quad-measure}),
\item
  better explanation of the quasi quadrature network constructed by
  coupling oriented quasi quadrature measures in cascade, including 
  a figure illustration of the network architecture, 
  details of discrete implementation, issues of exact {\em vs.\/}
  approximate covariance or invariance in a practical implementation and experimental results showing
  examples of the type of information that is computed in different layers of the
  hierarchy (Section~\ref{sec-hier-quasi-quad-net}),
\item
  a more extensive experimental section showing the results of applying a mean-reduced
  QuasiQuadNet for texture classification, including additional
  experiments demonstrating the importance of scale covariance and
  better overall descriptions about the experiments that could not be given
  in the conference paper because of the space limitations
  (Section~\ref{sec-appl-texture-anal}).
\end{itemize}
In relation to the SSVM 2019 paper, this paper therefore gives a more
general treatment about the notion of scale covariance of more general
validity for continuous hierarchical networks, presents more experimental
results regarding the prototype application to texture classification
and gives overall better descriptions of the subjects treated in the paper,
including more extensive references to related literature.

\section{Related work}
\label{sec-rel-work}

In the area of scale-space theory, theoretical results have been
derived showing that Gaussian kernels and Gaussian derivatives
constitute a canonical class of linear receptive fields for an
uncommitted vision system
\cite{Iij62,Iij63-StudInfCtrl,Wit83,Koe84,BWBD86-PAMI,YuiPog86-PAMI,KoeDoo92-PAMI,Lin93-Dis,Lin94-SI,Lin96-ScSp,Flo97-book,WeiIshImi99-JMIV,Haa04-book,DuiFloGraRom04-JMIV,Lin10-JMIV}.
The conditions that specify this uniqueness property are basically
linearity, shift invariance and regularity properties combined with
different ways of formalizing the notion that new structures should
not be created from finer to coarser scales in a multi-scale
representation.

The receptive field responses obtained by convolution with such
Gaussian kernels and Gaussian derivatives are truly scale
covariant---a property that has been
used for designing a large number of scale-covariant
and scale-invariant feature detectors and image descriptors
\cite{Lin97-IJCV,Lin98-IJCV,BL97-CVIU,ChoVerHalCro00-ECCV,MikSch04-IJCV,Low04-IJCV,BayEssTuyGoo08-CVIU,TuyMik08-Book,Lin14-EncCompVis,Lin15-JMIV}.
With the generalization to affine covariance and affine invariance
based on the notion of affine scale-space
\cite{Iij63-StudInfCtrl,Lin93-Dis,LG96-IVC,Bau00-CVPR,MikSch04-IJCV}, these
theoretical developments served as a conceptual
foundation that opened up for a very successful track of methodology development for
image-based matching and recognition in classical computer vision.

In the area of deep learning, approaches to tackle the notion of scale have been
developed in different ways. By augmenting the training images with
multiple rescaled copies of each training image or by randomly resizing the training images
over some scale range (scale jittering), the robustness of a deep net
can usually be extended to moderate scaling factors
\cite{BarCas91-TNN,SimZis15-ICLR}.
Another basic data-driven approach consists of training a module to estimate
spatial scaling factors from the data by a spatial transformer
network \cite{JadSimZisKav15-NIPS,LinLuc19-CVPR}.
A more structural approach consists of applying deep networks to
multiple layers in an image pyramid
\cite{LinDolGirHeHarBel17-CVPR,LinGoyGirHeDol17-ICCV,HeKiDolGir17-ICCV,HuRam17-CVPR}, 
or using some other type of
multi-channel approach where the input image is rescaled to different
resolutions, possibly combined with
interactions or pooling between the layers \cite{RenHeGirZhaSun16-PAMI,NahKimLee17-CVPR,ChePapKokMurYui17-PAMI,SinDav18-CVPR}.
Variations or extensions of this approach include scale-dependent
pooling \cite{YanChoLin16-CVPR}, using sets of
subnetworks in a multi-scale fashion \cite{CaiFanFerVas16-ECCV},
using dilated convolutions
\cite{YuKol15-arXiv,YuKolFun17-CVPR,MehRasCasShaHaj18-ECCV},
scale-adaptive convolutions \cite{ZhaTanZhaLiYan17-ICCV} or
adding additional branches of down-samplings and/or up-samplings in each
layer of the network \cite{WanKemFarYuiRas19-CVPR,CheFanXuYanKalRohYanFen19-arXiv}.

A more specific approach to designing a scale-covariant network is by
spatially warping the image data priori to image filtering by a
log-polar transformation \cite{HenVed17-ICML,EstAllZhoDan18-ICLR}.
Then, the scaling and rotation transformations are mapped to mere
translations in the transformed domain, although this property only
holds provided that
the origin of the log-polar transformation can be preserved between
the training data and the testing data.
Specialized learning approaches for scale-covariant or affine-covariant feature detection
have been developed for interest point
detection \cite{LenVed16-ECCV,ZhaYuKarCha17-CVPR}.

There is a large literature on approaches to achieve
rotation-covariant networks
\cite{DieFauKav16-ICML,LapSavBuhPol16-CVPR,WorGarTurBro17-CVPR,ZhoYeQiuJia17-CVPR,MarVolKomTui17-ICCV,CohWel17-ICLR,WeiGeiWelBooCoh18-NIPS,WeiHamSto18-CVPR,WorBro18-ECCV,CheHanZhoXu18-IP}
with applications to different domains including astronomy
\cite{DieWilDam15-RoyAstro},
remote sensing \cite{CheZhoHan17-GeoRemoteSens},
medical image analysis \cite{WanZheYanJinCheYin17-TBiomedHealth,BekLafVetEppPluDui18-MICCAI} and
texture classification \cite{AndDep18-arXiv}.
There are also approaches to invariant networks based on formalism from
group theory \cite{PogAns16-book,CohWel16-ICML,KonTri18-arXiv}.

In the context of more general classes of image transformations, it is
worth noting that beyond the classes of spatial scaling
transformations and spatial affine transformations (including
rotations), the framework of generalized axiomatic scale-space theory
\cite{Lin13-ImPhys,Lin17-arXiv-norm-theory-RF} does also allow for covariance and/or invariance
with regard to temporal scaling transformations \cite{Lin16-JMIV},
Galilean transformations and local multiplicative intensity
transformations \cite{Lin13-BICY,Lin13-PONE}.

Concerning biologically inspired neural networks,
Fuku\-shima \cite{Fuk80-BICY}
proposed to build upon Hubel and Wiesel's findings regarding
receptive fields in the primary visual cortex
(see \cite{HubWie05-book})
to construct a hierarchical neural network from repeated application
of models of simple and complex cells.
Poggio and his co-workers built on this idea and
constructed hand-crafted networks based on two layers of such
models expressed in terms of Gabor functions
\cite{RiePog99-Nature,SerWolBilRiePog07-PAMI,JhuSerWolPog07-ICCV}. 

The approach of scattering convolution networks \cite{BruMal13-PAMI,SifMal13-CVPR,OyaMal15-CVPR}
is closely related,
where directional odd and even wavelet responses are computed 
and combined with a non-linear modulus (magnitude) operator
over a set of different orientations in the image domain and
over a hierarchy over a dyadic set of scales.

Other types of hand-crafted or structured networks have been constructed by applying
principal component analysis in cascade 
\cite{ChaJiaGaoLuZenMa15-TIP} or by using Gabor functions as
primitives to be modulated by learned filters \cite{LuaCheZhaHanLiu18-PAMI}.

Concerning hybrid approaches between scale space and deep learning, 
Jacobsen {\em et al.\/}\ \cite{JacGemLouSme16-CVPR} construct a
hierarchical network from learned linear combinations of Gaussian
derivative responses.
Shelhamer {\em et al.\/}\ \cite{SheWanDar19-arXiv} compose free-form filters with affine Gaussian
filters to adapt the receptive field size and shape to the image data.

Concerning the use of a continuous model of the transformation from the input data to the
output data in a hierarchical computation structure, which we will
here develop for deep networks from motivations of making it possible
for the network to fulfil
geometric transformation properties in spatial input data, such a
notion of continuous transformations from the input to the output has
been proposed as a model for neural networks prior to the deep
learning revolution by Le~Roux and Bengio \cite{LeRBen07-AISTATS}
from the viewpoint of an uncountable number of hidden units and
suggesting that that makes it possible for the network to represent some smooth functions
more compactly.

For an overview of texture classification, which we shall later use as
an application domain, we refer to the recent survey by
Liu {\em et al.\/} \cite{LiuCheFieZhaChePie19-IJCV} and the references
therein.

In this work, we aim towards a conceptual bridge between scale-space
theory and deep learning, with specific emphasis on handling the variability in image data
caused by scaling transformations. We will show that it is possible to design a wide class of
possible scale-covariant networks by coupling linear or
non-linear expressions in terms of Gaussian derivatives in cascade.
As a proof-of-concept that such a construction can lead to meaningful
results, we will present a specific example of such a network, based on
a mathematically and biologically motivated model of complex cells,
and demonstrate that it is possible to get quite promising performance
on texture classification, comparable or better than many
classical texture descriptors or other hand-crafted networks.
Specifically, we will demonstrate how the notion of scale covariance improves the ability to
perform predictions or generalizations to scaling variabilities in the
testing data that
are not spanned by the training data.

We propose that this opens up for studying other hybrid approaches
between scale space theory and deep learning to incorporate explicit
modelling of image transformations as a prior in hierarchical networks.

\section{General scale covariance property for continuous hierarchical networks}
\label{sec-sc-cov-cont-hier-nets}

For a visual observer that views a dynamic world, the size of 
objects in the image domain can vary substantially, because of variations in
the distance between the objects and the observer and because of objects
having physically different size in the world.
If we rescale an image pattern by a uniform scaling factor, we would
in general like the perception of objects in the underlying scene to
be preserved.%
\footnote{When rescaling an object in the image domain, there are three main scaling effects occurring:
  (i)~how large the object will be in the image domain, 
  (ii)~how large the image structures of the object will be relative to the resolution
  of the image sensor and (iii)~how large the object will be relative to the
  outer dimensions (the size) of the image sensor.
In this article, we focus primarily on the first effect, to design
mechanisms for achieving
scale covariance and scale invariance under variations of the apparent
size of objects in the image domain, assuming that the resolution as
well as the size of
the image sensor is sufficient to sufficiently resolve the interesting
image structures over the scale range we are interested in covering.
In a practical implementation, the resolution of the image data will
additionally imply a lower bound on how fine scale levels can be computed
(the inner scale). The size of the image sensor will also impose an additional
upper bound on how large objects can be captured (the outer scale).
While such effects may also be highly important with regard to a specific
application, the topic of this article concerns how to handle the
essential geometric effects of the image
transformations, leaving more detailed issues of image sampling and
handling of image boundaries for future work.}
A natural precursor to achieving such a
scale-invariant perception of the world is to have a scale-covariant
image representation.
Specifically, a scale-covariant image representation can often be used
as a basis for constructing scale-invariant image descriptors and/or
scale-invariant recognition schemes.

In the area of scale-space theory
\cite{Iij62,Wit83,Koe84,KoeDoo92-PAMI,Lin93-Dis,Lin94-SI,Flo97-book,Haa04-book},
theoretically well-founded approaches have
been developed to handle the notion of scale in image data and to
construct scale-covariant and scale-invariant image representations
\cite{Lin97-IJCV,Lin98-IJCV,BL97-CVIU,ChoVerHalCro00-ECCV,MikSch04-IJCV,Low04-IJCV,BayEssTuyGoo08-CVIU,TuyMik08-Book,Lin13-ImPhys,Lin15-JMIV}.
In this section, we will present a general argument of how these
notions can be extended to construct provably scale-covariant hierarchical
networks, based on continuous models of the image operations between
adjacent layers.

Given an image $f(x)$, consider a multi-scale representation 
$L(x;\; s)$ constructed by Gaussian convolution and then from this
scale-space representation defining a family of
scale-parameter\-ized possibly non-linear operators ${\cal D}_{1,s_1}$
over a continuum of scale
parameters $s_1$ 
\begin{equation}
  F_1(\cdot;\; s_1) = ({\cal D}_{1,s_1} \, f)(\cdot),
\end{equation}
where the effect of the Gaussian smoothing operation is incorporated 
in the operator ${\cal D}_{1,s_1}$.

Within the framework of Gaussian scale-space representation
\cite{Iij62,Wit83,Koe84,KoeDoo92-PAMI,Lin93-Dis,Lin94-SI,Flo97-book,Haa04-book}, we could consider
these operators as being formed from sufficiently homogeneous possibly
non-linear combinations of Gaussian derivative operators, such that
they under a rescaling of the input domain $x' = S x$ by a factor of $S$
are guaranteed to obey the scale covariance property
\begin{equation}
  \label{eq-scale-covariance-prop-F1}
  F'_1(x';\; s'_1) = S^{\alpha_1} F_1(x;\; s_1) 
\end{equation}
for some constant $\alpha_1$ and some transformation of the scale
parameters $s'_1 = \phi_1(s_1)$. In other words, for any image
representation computed over the original image domain $x$ at scale
$s_1$, it should be possible to find a corresponding representation over the
transformed domain $x' = S x$ at scale $s_1'$ with a possibly transformed
magnitude as determined by the relative amplification factor
$S^{\alpha_1}$.

\begin{figure}[hbtp]
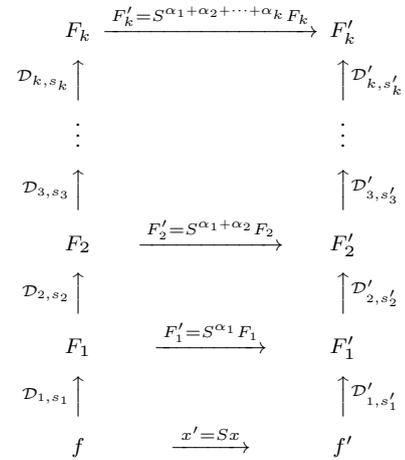

  \begin{center}
    \[
      \hspace{25mm}\begin{CD}
       F_k @>{F'_k= S^{\alpha_1+\alpha_2 + \dots + \alpha_k} F_k}>> F_k' \\
       @A{{\cal D}_{k,s_k}}AA @AA{{\cal D}'_{k,s'_k}}A \\
      \vdots @. \vdots \\
        @A{{\cal D}_{3,s_3}}AA @AA{{\cal D}'_{3,s'_3}}A \\ 
       F_2 @>{F'_2= S^{\alpha_1+\alpha_2} F_2}>> F_2' \\
        @A{{\cal D}_{2,s_2}}AA @AA{{\cal D}'_{2,s'_2}}A \\ 
       F_1 @>{F'_1= S^{\alpha_1} F_1}>> F_1' \\
        @A{{\cal D}_{1,s_1}}AA @AA{{\cal D}'_{1,s'_1}}A \\
        f @>{x'=Sx}>> f'
    \end{CD}
    \]
  \end{center}
  \caption{Commutative diagram for a scale-covariant hierarchical network
    constructed according to the presented sufficiency result. Provided that the
    individual differential operators ${\cal D}_{k,s_k}$ between
    adjacent layers are scale covariant, which for example holds for
    the class of homogeneous differential expressions of the form
    (\ref{eq-sc-cov-hom-diff-expr-general-gamma}) as well as
    self-similar compositions of such operations that additionally
    satisfy corresponding homogeneity requirements, it follows
    that it will be possible to perfectly match the corresponding
    layers $F_k$ and $F_k'$ under a scaling transformations of the
    underlying image domain $f'(x') =
    f(x)$ for $x' = Sx$, provided that
the scale parameter $s_k$ in layer $k$ is proportional to the scale
parameter $s_1$ in the first layer, $s_k = r_k^2 \, s_1$, for some
scalar constants $r_k$. For such a network constructed from scale-space
operations based on the Gaussian scale-space theory framework, the scale parameters
in the two domains should be related according to $s_k' = S^2 s_k$.}
  \label{fig-comm-diag-hier-network}
\end{figure}

\begin{figure}[hbt]
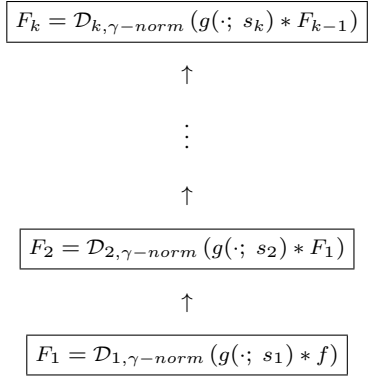

\begin{center}
\framebox{$F_ k ={\cal D}_{k,\gamma-norm} \, (g(\cdot;\; s_k) * F_{k-1})$} \\
$\,$ \\ $\uparrow$\\ $\,$ \\
$\vdots$ \\
$\,$ \\ $\uparrow$\\ $\,$ \\
\framebox{$F_ 2 ={\cal D}_{2,\gamma-norm} \, (g(\cdot;\; s_2) * F_1)$} \\
$\,$ \\ $\uparrow$\\ $\,$ \\
\framebox{$F_ 1 ={\cal D}_{1,\gamma-norm} \, (g(\cdot;\; s_1) * f)$}
\end{center}
\caption{A hierarchical network defined by coupling scale-covariant differential
  expressions formulated within the continuous scale-space framework
  will be guaranteed to be provably scale covariant provided that the scale
  parameters in higher layers $s_k$ for $k \geq 2$ are proportional to
  the scale parameter $s_1$ in the first layer.
  If the scale normalization parameter $\gamma$ in the
  scale-normalized derivative expressions is equal to one, then
  general differential expressions in terms of such derivatives can be
used based on the transformation property
(\ref{eq-sc-transf-gammanorm-ders-gamma-eq-1}). If the scale
normalization parameter $\gamma$ is not equal to one, then one can
take homogeneous polynomial differential expressions of the form
(\ref{eq-sc-cov-hom-diff-expr-general-gamma}) as well as self-similar
transformations of such expressions.
(In this schematic illustration, the arguments of the layers $F_1$,
$F_2$, $F_{k-1}$ and $F_k$, which should be $F_ 1(\cdot;\; s_1)$, $F_ 2(\cdot;\;
s_1, s_2)$, $F_{k-1}(\cdot;\; s_1, s_2, \dots, s_{k-1})$ and 
$F_k(\cdot;\; s_1, s_2, \dots, s_{k-1}, s_k)$, respectively, 
have been suppressed to simplify the
notation. The argument of the input data $f$ should be $f(\cdot)$.)}
\label{fig-sc-cov-network-scsp}
\end{figure}

By concatenating a set of corresponding scale-parameter\-ized
differential operators ${\cal D}_{k,s_k}$
\begin{equation}
  F_k(\cdot;\; s_1, \dots, s_{k-1}, s_k) 
  = {\cal D}_{k,s_k} \, F_{k-1}(\cdot;\; s_1, \dots, s_{k-1})
\end{equation}
that obey similar scale covariance properties such that
\begin{equation}
  \label{eq-scale-covariance-prop-Fk}
  F'_k(x';\; s'_1, \dots, s'_k) 
  = S^{\alpha_1} \dots \, S^{\alpha_{k}}  \, F_k(x;\; s_1, \dots, s_k),
\end{equation}
it follows that the combined hierarchical network is guaranteed to be provably
scale covariant, see Figures~\ref{fig-comm-diag-hier-network} and~\ref{fig-sc-cov-network-scsp} for 
schematic illustrations. Specifically, it is natural to choose the scale
parameters $s_k$ in the higher layers proportional to the scale
parameter $s_1$ in the first layer to guarantee scale covariance.

More generally, we could also consider constructing scale-covariant networks from other types of scale-covar\-iant
operators that obey similar scaling properties as in
Equations~(\ref{eq-scale-covariance-prop-F1}) and (\ref{eq-scale-covariance-prop-Fk}), for example, expressed in
terms of a basis of rescaled Gabor functions or a family of continuously rescaled wavelets.
Then, however, the information reducing properties from finer to
coarser scales in the representation computed by Gaussian convolution
and Gaussian derivatives are, however, not guaranteed to hold.
As mentioned above, the Gaussian kernel and the Gaussian derivatives can be
uniquely determined from different ways of formalizing the requirement that they should not introduce
new image structures from finer to coarser scales in a multi-scale
representation \cite{Iij62,Iij63-StudInfCtrl,Wit83,Koe84,BWBD86-PAMI,YuiPog86-PAMI,KoeDoo92-PAMI,Lin93-Dis,Lin94-SI,Lin96-ScSp,Flo97-book,WeiIshImi99-JMIV,Haa04-book,DuiFloGraRom04-JMIV,Lin10-JMIV}.

In this overall structure, there is a large flexibility in how to
choose the operators ${\cal D}_{k,s_k}$. Within the family of operators
defined from a scale-space representation, we could
consider a large class of differential expressions and differential
invariants in terms of scale-normalized Gaussian derivatives
\cite{Lin97-IJCV} that guarantee provable scale covariance. 

For example, if we choose to express the first
differential operator ${\cal D}_{1,s_1}$ in a basis in terms of
scale-normalized derivatives \cite{Lin97-IJCV} (here with the
multi-index notation $\partial_x^n = \partial_{x_1^{n_1} ... x_D^{n_D}}$ for
the partial derivatives in $D$ dimensions and $|n| = n_1 + \dots + n_D$)
\begin{equation}
  \label{eq-sc-norm-ders}
  \partial_{\xi^n} 
  = \partial_{x^n,\gamma-norm} = s^{|n| \gamma/2} \, \partial_x^n
\end{equation}
computed from a scale-space
representation of the input signal
\begin{equation}
  \label{eq-scsp-repr-of-f}
  L_1(\cdot;\; s_1) 
  = g(\cdot;\; s_1) * f(\cdot)
\end{equation}
by convolution with Gaussian kernels
\begin{equation}
   g(x;\; s) = \frac{1}{(2 \pi s)^{D/2}} e^{-\frac{|x|^2}{2s}}
\end{equation}
and with $s$ in (\ref{eq-sc-norm-ders}) determined from $s_1$ in (\ref{eq-scsp-repr-of-f}),
it then follows that under a rescaling of the image domain $f'(x') = f(x)$ for $x' = S \, x$
the scale-normalized derivatives transform
according to \cite[Eq.~(20)]{Lin97-IJCV} 
\begin{equation}
  \partial_{\xi'^n} L'_1(x';\; s'_1) = S^{|n|(\gamma-1)} \, \partial_{\xi^n} L_1(x;\; s_1)
\end{equation}
provided that the scale parameters are matched according to $s'_1 = S^2 \, s_1$.
Specifically, in the special case of choosing $\gamma = 1$, the
scale-normalized derivatives will be equal
\begin{equation}
  \label{eq-sc-transf-gammanorm-ders-gamma-eq-1}
  \partial_{\xi'^n} L'_1(x';\; s'_1) = \partial_{\xi^n} L_1(x;\; s_1).
\end{equation}
This implies that any scale parameterized differential operator ${\cal D}_{1,s_k}$ 
that can be expressed as a sufficiently regular function $\psi$
\begin{equation}
  {\cal D}_{1,s_1} \, f = \psi({\cal J}_{N,s_1} L_1)
\end{equation}
of the scale-normalized
$N$-jet, ${\cal J}_{N,s_1} L_1$, of the scale-space
representation $L_1$ of the input image, which is the union of all
partial derivatives up to order $N$
\begin{equation}
  {\cal J}_{N,s_1} L_1 = \cup_{1 \leq |n| \leq N} \, \partial_{\xi^n} L_1,
\end{equation}
will satisfy the scale covariance property
(\ref{eq-scale-covariance-prop-F1}) for $\alpha_1 = 0$. More generally, it
is not necessary that all the derivatives are computed at the same scale,
although such a choice could possibly be motivated from conceptual simplicity.

In the less specific case of choosing $\gamma \neq 1$, we can consider
homogeneous polynomials of scale-normalized derivatives of the form
\begin{equation}
  \label{eq-sc-cov-hom-diff-expr-general-gamma}
  {\cal D}_{s_1} f = \sum_{i=1}^I c_i \prod_{j = 1}^J L_{1,x^{\beta_{ij}}}
\end{equation}
for which the sum of the orders of differentiation in a certain term
\begin{equation}
  \sum_{j=1}^J |\beta_{ij}| = M
\end{equation}
does not depend on the index $i$ of that term. The corresponding
scale-normalized expression with the regular spatial derivatives
replaced by $\gamma$-normalized derivatives is
\begin{equation}
  \label{eq-sc-norm-self-sim-diff-expr}
  {\cal D}_{\gamma-norm,s_1} f = s_1^{M\gamma/2} \, {\cal D}_{s_1} f 
\end{equation}
and transforms
according to \cite[Eq.~(25)]{Lin97-IJCV} 
\begin{equation}
  {\cal D}'_{\gamma-norm,s'_1} f' =  S^{M(\gamma-1)} {\cal D}_{\gamma-norm,s_1} f
\end{equation}
under any scaling transformation $f'(x') = f(x)$ for $x' = S \; x$
provided that the scale levels are appropriately matched $s'_1 = S^2 \, s_1$.
Such a self-similar form of scaling transformation
will also be preserved under self-similar transformations $z \mapsto
z^{\delta}$ of such expressions as well as for a rich family of
polynomial combinations as well as rational expressions of
such expressions as long as the scale covariance property
(\ref{eq-scale-covariance-prop-F1}) is preserved.

A natural complementary argument to constrain such self-similar compositions is to
preserve the dimensionality of the image data, such that each layer 
$F_k$ has the same dimensionality $[\mbox{intensity}]$ as the input
image $f$. If a polynomial is used for constructing a composed non-linear
differential expression ${\cal D}_{comp,s_1} f$ by combinations of
differential expressions of the form
(\ref{eq-sc-norm-self-sim-diff-expr})
and if this composed polynomial is a homogeneous polynomial of order $P$
relative to the underlying partial derivatives $\partial_{\xi^n}$ in
the $N$-jet, in the sense that under a rescaling of the magnitude of
the original image data $f$ by a factor of $\beta$ such that $f'(x') = \beta f(x)$
the differential expression transforms according to
\begin{equation}
  \label{eq-hom-req}
   {\cal D}'_{comp,s_1'} f' = \beta^P {\cal D}_{comp,s_1} f,
\end{equation}
we should then transform that differential expression
by the power $1/P$ to preserve the dimensionality of
$[\mbox{intensity}]$.
A similar argument applies to differential entities formed from
rational expressions of differential expressions of the form
(\ref{eq-sc-norm-self-sim-diff-expr}) as long as the scale covariance property
(\ref{eq-scale-covariance-prop-F1}) is preserved.

Corresponding reasoning as done here regarding the transformation from the input image $f$ to
the first layer $F_1$ can be performed regarding the transformations
${\cal D}_{k,s_k}$ between any pairs of adjacent layers $F_{k-1}$ and $F_k$, implying that
if the differential operators ${\cal D}_{k,s_k}$ are chosen from
similar families of differential operators as described above
regarding the first differential operator ${\cal D}_{1,s_1}$, 
then the entire layered hierarchy will be scale covariant, provided that
the scale parameter $s_k$ in layer $k$ is proportional to the scale
parameter $s_1$ in the first layer, $s_k = r_k^2 \, s_1$, for some
scalar constants $r_k$
(see Figure~\ref{fig-comm-diag-hier-network}).  This opens up
for a large class of provably scale-covariant continuous hierarchical networks based
on differential operators defined from the scale-space framework,
where it remains to be determined which of these possible networks
lead to desirable properties in other respects.
In the following, we will develop one specific way of defining
such a scale-covariant continuous network, by
choosing these operators based on functional models of complex cells
expressed within the Gaussian scale-space paradigm.

\section{The quasi quadrature measure over a 1-D signal}
\label{sec-quasi-quad-1D}

Consider the scale-space representation \cite{Iij62,Wit83,Koe84,Lin93-Dis,Lin94-SI,Flo97-book,Haa04-book} 
\begin{equation}
  L(\cdot;\; s) = g(\cdot;\; s) * f(\cdot)
\end{equation}
of a 1-D signal $f(x)$ defined by
convolution with Gaussian kernels 
\begin{equation}
   g(x;\; s) = \frac{1}{\sqrt{2\pi s}} \, e^{-\frac{x^2}{2s}}
\end{equation}
and with 
scale-normalized derivatives according to \cite{Lin97-IJCV}
\begin{equation}
  \partial_{\xi^n} = \partial_{x^n,\gamma-norm} = s^{n \gamma/2} \, \partial_x^n.
\end{equation}
In this section, we will describe a quasi quadrature entity that measures the
local energy in the first- and second-order derivatives in the scale-space
representation of a 1-D signal and analyse its behaviour to image structures over
multiple scales.
Later in Section~\ref{sec-ori-quasi-quad-measure}, an oriented
extension of this measure to two-dimensional image space will be used
for expressing a functional model of complex cells that reproduces
some of the known properties of complex cells.

\begin{figure*}[hbt]
  \begin{center}
    \begin{tabular}{ccc}
        {\footnotesize\em $g(x;\ s)$\/} 
           & {\footnotesize\em $g_x(x;\ s)$\/} 
           & {\footnotesize\em $g_{xx}(x;\ s)$\/} \\
        \includegraphics[width=0.25\textwidth]{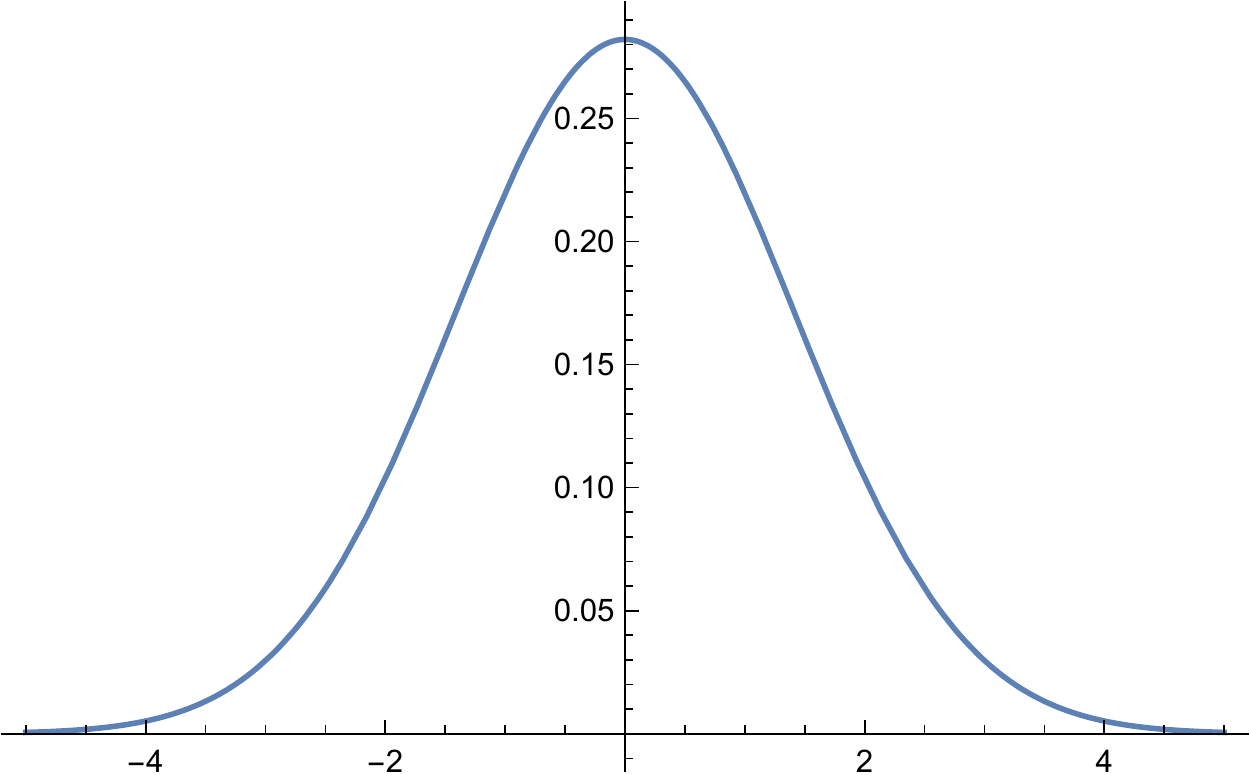}
           & \includegraphics[width=0.25\textwidth]{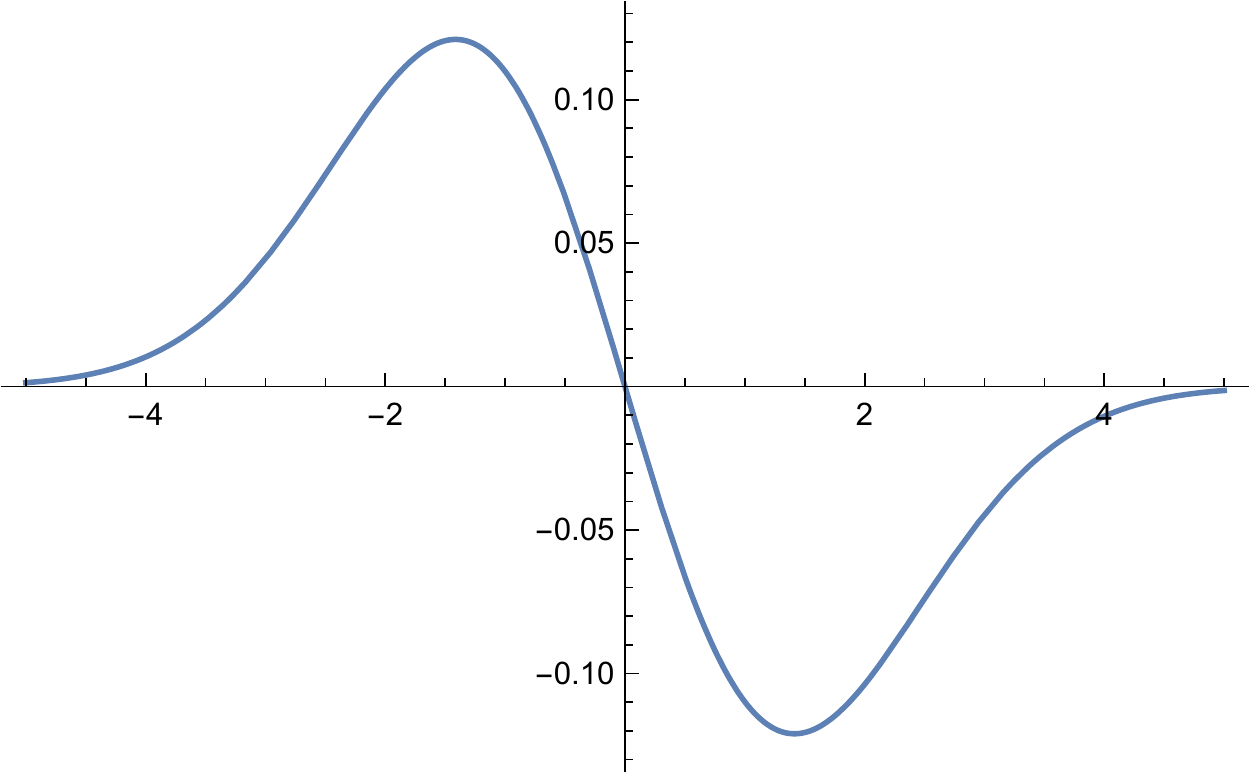} 
           & \includegraphics[width=0.25\textwidth]{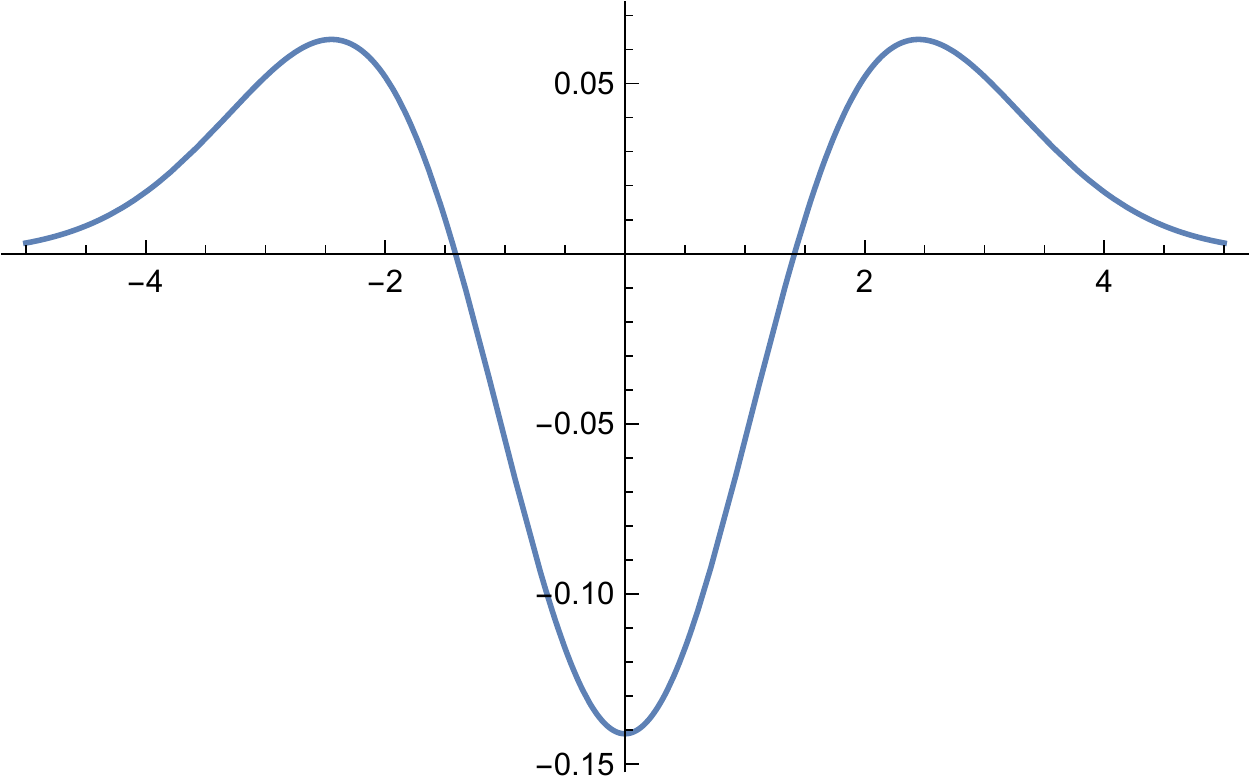} 
           \medskip \\
        {\footnotesize\em ${\cal Q}_{x,norm} L$}
          & {\footnotesize\em ${\cal Q}_{x,norm} L$} 
          & {\footnotesize\em ${\cal Q}_{x,norm} L$} \\
        \includegraphics[width=0.25\textwidth]{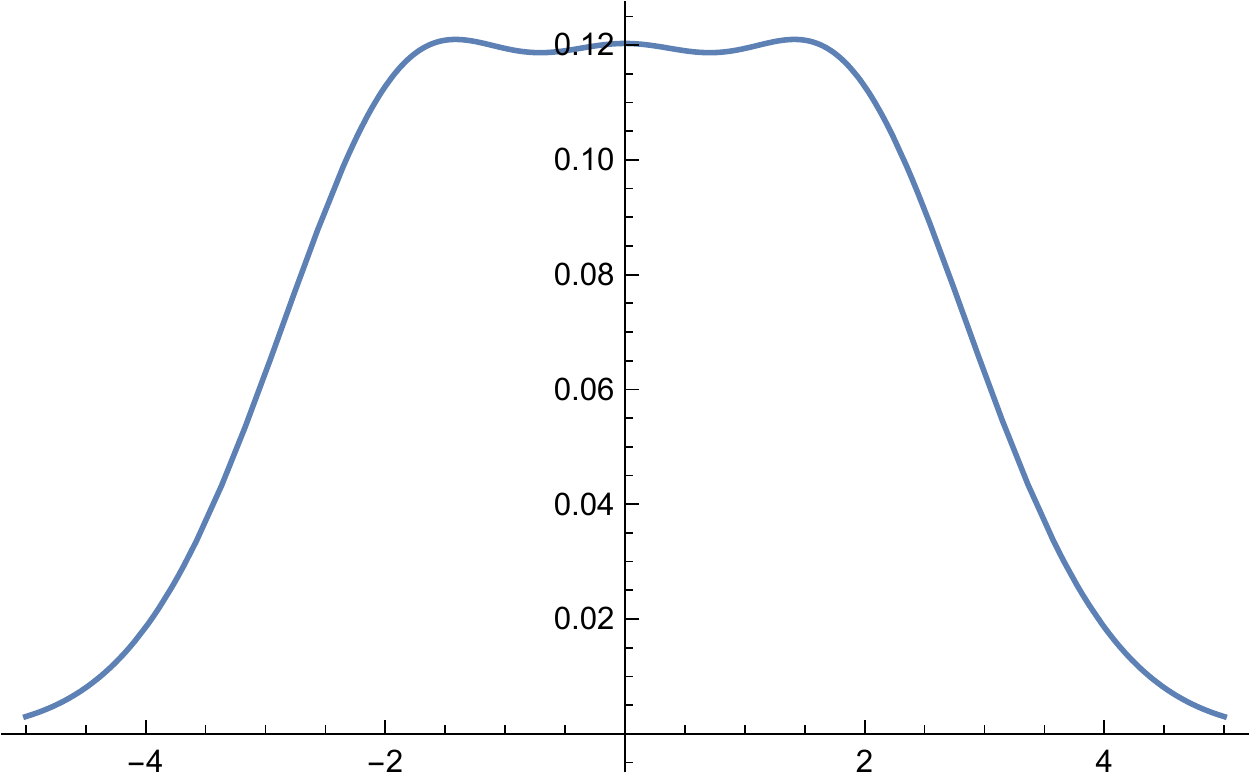} 
          & \includegraphics[width=0.25\textwidth]{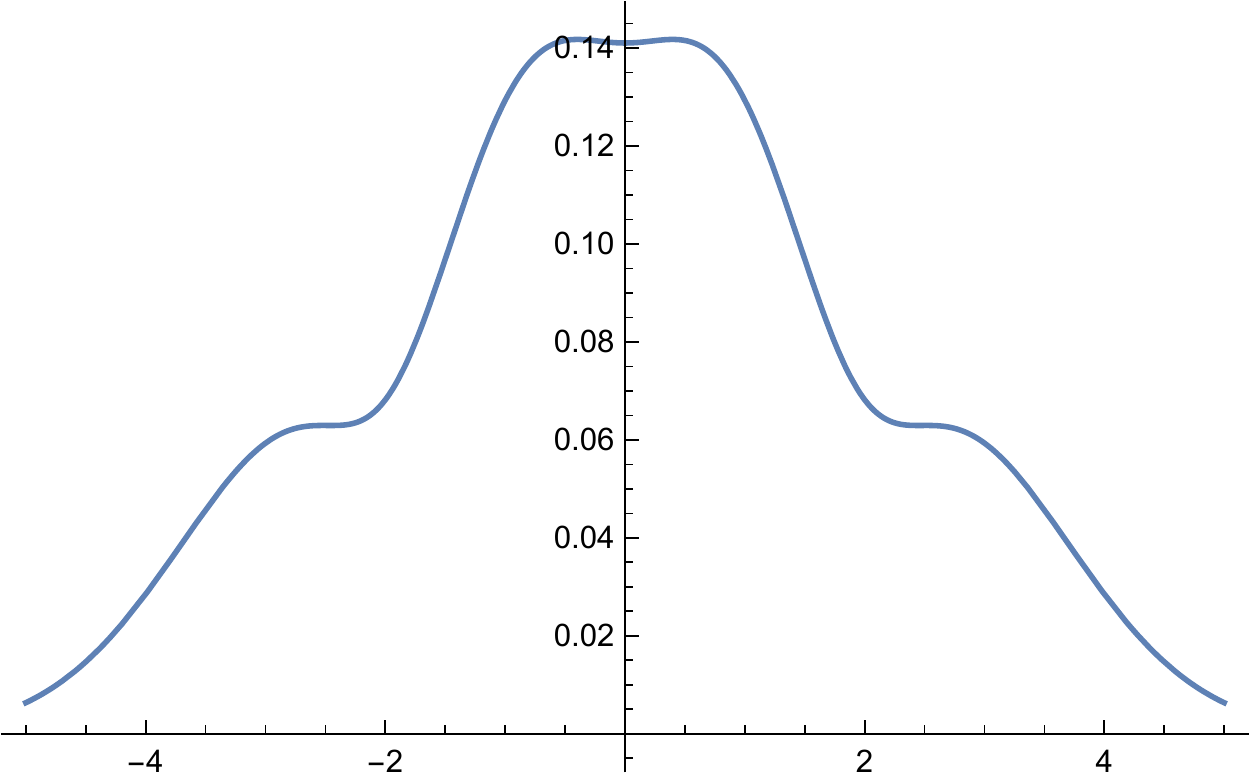} 
          & \includegraphics[width=0.25\textwidth]{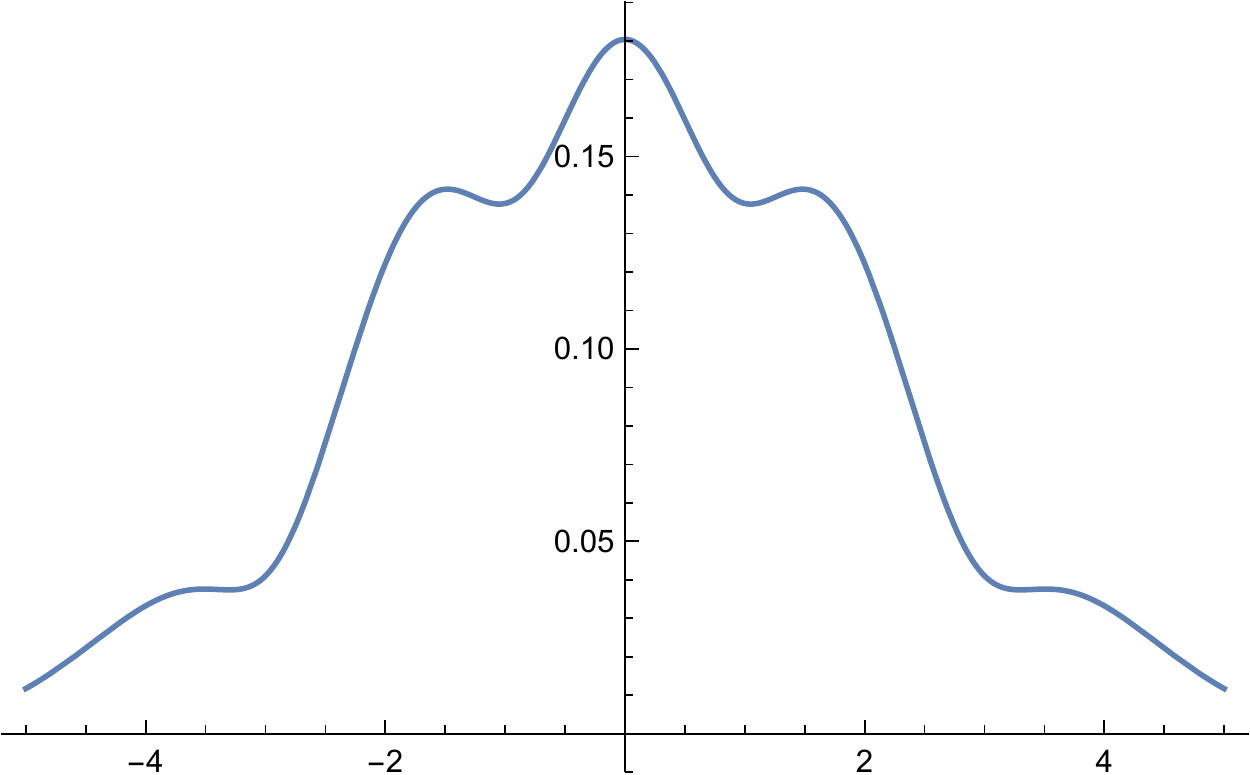} 
          \medskip \\
    \end{tabular} 
  \end{center}
  \caption{1-D Gaussian derivatives up to orders 0, 1 and 2 for $s_0 = 1$ with the
    corresponding 1-D quasi quadrature measures
    (\ref{eq-quasi-quad-1D}) computed from them at
    scale $s = 1$ for $C = 8/11$.
   (Horizontal axis: $x \in [-5, 5]$.)}
  \label{fig-quasiquad-gaussders-1D}
\end{figure*}

\subsection{Quasi quadrature measure in 1-D}

Motivated by the fact that the first-order derivatives primarily
respond to the locally odd component of the signal, whereas the
second-order derivatives primarily respond to the locally even
component of a signal, it is natural to aim at a differential feature
detector that combines locally odd and even components in a
complementary manner. By specifically combining the first- and
second-order scale-normalized derivative responses in a 
Euclidean way, we obtain a quasi quadrature measure of the form
\begin{equation}
  \label{eq-quasi-quad-1D}
  {\cal Q}_{x,norm} L 
  = \sqrt{\frac{s \, L_x^2 + C \, s^2 \, L_{xx}^2}{s^{\Gamma}}}
\end{equation}
as a modification%
\footnote{Compared to the previous work on quasi quadrature measures in
 \cite{Lin97-IJCV,Lin18-SIIMS}, we here transform use the previous 1-D
 quasi quadrature measure by a square root function to maintain the same dimensionality as the input
signal, which is a useful property when defining hierarchical networks
by coupling quasi quadrature measures in cascade.} 
of the quasi quadrature measures previously proposed and studied in \cite{Lin97-IJCV,Lin18-SIIMS},
with the scale normalization parameters $\gamma_1$ and $\gamma_2$
of the first- and second-order derivatives coupled according to
$\gamma_1 = 1 - \Gamma$ and $\gamma_2 = 1 - \Gamma/2$ to enable
scale covariance by adding derivative expressions of different orders
only for the scale-invariant choice of $\gamma = 1$.
This differential entity can be seen as an approximation%
\footnote{For a true quadrature pair, the Euclidean norm of the two
  feature responses should be constant for sine waves of all
  frequencies and thus insensitive to the local phase of the signal. Due to the restriction of filters to first- and
  second-order Gaussian derivatives only, this property cannot hold for
  sine waves of all frequencies at all scales simultaneously. Near the scale levels
  that are determined by applying scale selection to a sine wave of a
  given frequency, the phase dependency in the response will, however,
be moderately low, as described in \cite{Lin97-IJCV,Lin18-SIIMS}.
Since the Euclidean norm of the first- and second-order Gaussian
derivative responses tries to mimic these properties of a quadrature
pair, although not being able to
obey them fully because of the restriction of the filter basis to 
the square responses of the first- and second-order Gaussian derivatives only, this entity is termed quasi quadrature.}
of the notion
of a quadrature pair of an odd and even filter \cite{Gab46} as more traditionally
formulated based on a Hilbert transform \cite[p.~267-272]{Bra99}, while confined within
the family of differential expressions based on Gaussian derivatives.

Intuitively, this quasi quadrature operator is intended as a measure of the amount of
local changes in the signal, not specific to whether the dominant
response comes from odd first-order derivatives or even second-order
derivatives, and with additional scale selective properties as will be
described later in Section~\ref{sc-sel-props-quasi-quad}.

If complemented by spatial integration, the components of the quasi
quadrature measure are specifically related to the following
class of energy measures over the frequency domain
(Lindeberg \cite[App.~A.3]{Lin97-IJCV}):
\begin{align}
  \begin{split}
  E_{m,\gamma-norm}
  & = \int_{x \in \bbbr} s^{m\gamma} \, L_{x^m}^2 \, dx
  \end{split}\nonumber\\
  \begin{split}
  & = \frac{s^{m\gamma}}{2 \pi}
      \int_{\omega \in \bbbr} | \omega|^{2 m} \, \hat{g}^2(\omega;\; s)
      \, d\omega.
  \end{split}
\end{align}
For the specific choice of $C = 1/2$ and $\Gamma = 0$, the square of the quasi quadrature measure
(\ref{eq-quasi-quad-1D}) coincides
with the proposals by Loog \cite{Loo07-SSVM} and Griffin
\cite{Gri07-PAMI} to define a metric of the $N$-jet in scale space,
which can specifically been seen as an approximation of the
variance of a signal using a Gaussian spatial weighting function.%
\footnote{To understand the relationship between the proposed metric
  of the $N$-jet with the variance of the signal, which has been previously
  described by Griffin \cite{Gri07-PAMI}  and Loog
  \cite{Loo07-SSVM}, consider a 1-D signal that is approximated by its
  second-order Taylor expansion $L(x) = c_0 + c_1 \, x + c_2 \, x^2/2$
  around $x = 0$ at some scale level in scale space, where $c_0 = L(0)$, $c_1 = L_x(0)$ and $c_2 = L_{xx}(0)$.
  The variance of this signal with a Gaussian weighting
function $g(x) = \exp(-x^2/2s)/\sqrt{2 \pi s}$ around $x = 0$ is $V = \int_{x \in \bbbr} (L(x))^2 \, g(x) \,
 dx - (\int_{x \in \bbbr} L(x) \, g(x) \, dx)^2 = M_2 - M_1^2$.
Solving these integrals gives $M_2 = c_0^2+c_0 \, c_2
s+c_1^2 s+ 3 c_2^2 s^2/4$ and $M_1 =
c_0+c_2 s/2$, from which we obtain $V = s \, c_1^2 +s^2 \, c_2^2/2
= s \, L_x(0)^2 + s^2 \, (L_{xx}(0))^2/2 = ({\cal Q}_{x,norm} L)^2$ at
$x = 0$ for $C = 1/2$ and
$\Gamma = 0$. A similar result holds if we instead determine a
preferred representative of the class of possible signals that has similar
2-jets (the metamer) by its metamery
class norm minimizer $L(x) = (c_0 - c_2 s/2) + c_1 \, x + c_2 \, x^2/2$
according to Griffin \cite[Sect.~2.1]{Gri07-PAMI}.}

Figure~\ref{fig-quasiquad-gaussders-1D} shows the result of computing
this quasi quadrature measure for a Gaussian peak as well as its
first- and second-order derivatives. As can be seen, the quasi
quadrature measure is much less sensitive to the position of the peak
compared to {\em e.g.\/}\ the first- or second-order
derivative responses. Additionally, the quasi quadrature measure also has some
degree of spatial insensitivity for a first-order Gaussian
derivative (a local edge model) and a second-order Gaussian derivative.

\subsection{Determination of the parameter $C$}

To determine the weighting parameter $C$ between local second-order
and first-order information, let us consider a Gaussian blob $f(x) = g(x;\; s_0)$
with spatial extent given by $s_0$ as input model signal.

By using the semi-group property of the Gaussian kernel
$g(\cdot;\; s_1) * g(\cdot;\; s_2) = g(\cdot;\; s_1 + s_2)$,
it follows that the scale-space representation is given by
$L(x;\; s) = g(x;\; s_0+s)$ and that the
first- and second-order derivatives of the scale-space
representation are 
\begin{align}
  \begin{split}
      L_x = g_x(x;\; s_0+s) = -\frac{x}{(s_0+s)} \, g(x;\; s_0+s),
  \end{split}\\
  \begin{split}
      L_{xx} = g_{xx}(x;\; s_0+s) = \frac{\left(x^2 - (s_0 + s)\right)}{(s_0+s)^2} \, g(x;\; s_0+s),
  \end{split}
\end{align}
from which the quasi quadrature measure (\ref{eq-quasi-quad-1D}) can be computed in closed form
\begin{multline}
   {\cal Q}_{x,norm} L
   = \\ \frac{s^{\frac{1-\Gamma}{2}} e^{-\frac{x^2}{2(s+s_0)}} 
       \sqrt{x^2 (s+s_0)^2 + C s \left(s+s_0-x^2\right)^2+2}}
              {\sqrt{2 \pi} \,  (s+s_0)^{5/2}}.
\end{multline}
By determining the weighting parameter $C$ such that it minimizes the
overall ripple in the squared quasi quadrature measure for a Gaussian input
\begin{equation}
   \hat{C} = \operatorname{argmin}_{C \geq 0}
                      \int_{x=-\infty}^{\infty} 
                         \left( \partial_x({\cal Q}^2_{x,norm} L) \right)^2 \, dx,
\end{equation}
which is one way of quantifying the desire to have a stable response
under small spatial perturbations of the input, we obtain
\begin{equation}
  \hat{C} = \frac{4 (s+s_0)}{11 s},
\end{equation}
which in the special case of choosing $s = s_0$ corresponds to $C = 8/11 \approx 0.727$.
This value is very close to the value $C = 1/\sqrt{2} \approx 0.707$
derived from an equal contribution condition in
\cite[Eq.~(27)]{Lin18-SIIMS} for the special case of choosing $\Gamma = 0$.

\subsection{Scale selection properties}
\label{sc-sel-props-quasi-quad}

To analyse how the quasi quadrature measure selectively responds to image
structures of different size, which is important when computing the quasi
quadrature entity at multiple scales, we will in this section analyse
the scale selection properties of this entity. 

Let us consider the result of using Gaussian derivatives of
orders 0, 1 and 2 as models of different types of local input signals, {\em i.e.\/},
\begin{equation}
f(x) = g_{x^n}(x;\; s_0)
\end{equation}
for $n \in \{ 0, 1, 2 \}$.
For the zero-order Gaussian kernel, the scale-normalized quasi
quadrature measure at the origin is given by
\begin{equation}
  \left. {\cal Q}_{x,norm} L \right|_{x=0,n=0}
  = \frac{\sqrt{C} s^{1-\Gamma/2}}{2 \pi (s+s_0)^2}.
\end{equation}
For the first-order Gaussian derivative kernel, the scale-norm\-alized quasi
quadrature measure at the origin is 
\begin{equation}
  \left. {\cal Q}_{x,norm} L \right|_{x=0,n=1}
  = \frac{s_0^{1/2} s^{(1-\Gamma)/2}}{2 \pi (s+s_0)^2},
\end{equation}
whereas for the second-order Gaussian derivative kernel,
the scale-normalized quasi quadrature measure at the origin is
\begin{equation}
  \left. {\cal Q}_{x,norm} L \right|_{x=0,n=2}
  = \frac{3 \sqrt{C} s_0 s^{1-\Gamma/2}}{2 \pi  (s+s_0)^3}.
\end{equation}
By differentiating these expressions with respect to the scale
parameter $s$, we find
that for a zero-order Gaussian kernel the maximum response over scale
is assumed at
\begin{equation}
   \left. \hat{s} \right|_{n=0} 
   = \frac{s_0 \, (2 -\Gamma)}{2+\Gamma},
\end{equation}
whereas for the first- and second-order derivatives, respectively, the
maximum response over scale is assumed at
\begin{align}
  \begin{split}
     \left. \hat{s} \right|_{n=1} 
     = \frac{s_0 \; (1 -\Gamma)}{3+\Gamma}, 
  \end{split}\\
  \begin{split}
     \left. \hat{s} \right|_{n=2} 
     = \frac{s_0 \, (2 - \Gamma)}{4+\Gamma}.
  \end{split}
\end{align}
In the special case of choosing $\Gamma = 0$, these scale estimates
correspond to 
\begin{align}
  \begin{split}
  \label{eq-scsel-quasiquad-0-der}
     \left. \hat{s} \right|_{n=0} 
     = s_0, \quad\quad
  \end{split}\\
  \begin{split}
  \label{eq-scsel-quasiquad-1-der}
     \left. \hat{s} \right|_{n=1} 
     = \frac{s_0}{3}, \quad\quad
  \end{split}\\
  \begin{split}
  \label{eq-scsel-quasiquad-2-der}
     \left. \hat{s} \right|_{n=2} 
     = \frac{s_0}{2}.
  \end{split}
\end{align}
Thus, for a Gaussian input signal, the selected scale level will for
the most scale-invariant choice of using $\Gamma = 0$ reflect the
spatial extent $\hat{s} = s_0$ of the blob, whereas if we would like
the scale estimate to reflect the scale parameter of first- and
second-order derivatives, we would have to choose $\Gamma = -1$.
An alternative motivation for using finer scale levels for the
Gaussian derivative kernels is to regard the positive and negative lobes
of the Gaussian derivative kernels as substructures of a more complex
signal, which would then warrant the use of finer scale levels to
reflect the substructures of the signal 
((\ref{eq-scsel-quasiquad-1-der}) and (\ref{eq-scsel-quasiquad-2-der})).

\subsection{Spatial sensitivity of the quasi quadrature measure}

Due to the formulation of the quasi quadrature measure in terms of Gaussian
derivatives from the $N$-jet, the spatial sensitivity (phase
dependency) of this entity can be estimated from the first-order
component in the local Taylor expansion
\begin{equation}
  \frac{\sqrt{s} \, \partial_x ({\cal Q}_{x,norm} L)}{{\cal Q}_{x,norm} L}
   = \frac{s \, L_{xx} \, (s^{1/2} \, L_x + C \, s^{3/2} \, L_{xxx})}
              {s \, L_x^2 + C \, s^2 \, L_{xx}^2},
\end{equation}
where have expressed this entity in terms
of scale-normalized derivatives for $\gamma = 1$ to emphasize the
scale-invariant form of the scale-normalized perturbation measure
$s^{1/2} \, \partial_x ({\cal Q}_{x,norm} L)$.
Notably, this entity is zero at inflection points where $L_{xx} = 0$.

\subsection{Post-smoothed quasi quadrature measure}

To reduce the spatial sensitivity of the quasi quadrature measure, 
the definition in equation~(\ref{eq-quasi-quad-1D}) can be
complemented by spatial post-smoothing
\begin{equation}
  (\overline{\cal Q}_{x,norm} L)(\cdot;\; s, r^2 s) = g(\cdot;\; r^2 s) * ({\cal Q}_{x,norm} L)(\cdot;\; s),
\end{equation}
where the parameter $r$ is referred to as the relative post-smoothing
scale. When coupling quasi quadrature measures in cascade,
this amount of post-smoothing $r^2 s$ will represent the amount of
additional Gaussian smoothing before computing derivatives in the
next layer in the hierarchical feature representation.

This spatial post-smoothing operation serves as a scale-covariant
spatial pooling operation, notably with the support region, as determined by
the integration scale $r^2 s$, proportional to the current scale level $s$,
as opposed to the standard application of spatial pooling over
neighbourhoods of fixed size in most CNNs, which would then imply
violations of scale covariance.

\begin{figure*}[hbtp]
    \begin{center}
     \begin{tabular}{ccc}
        & & {\small $\partial_{\varphi}g(x, y;\; \Sigma)$} \\
       \includegraphics[height=0.17\textheight]{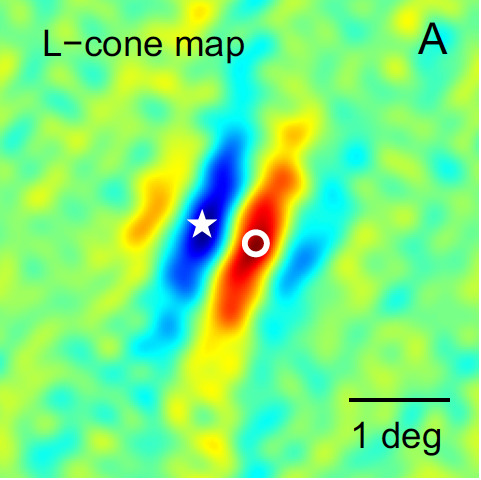}
       & \includegraphics[height=0.17\textheight]{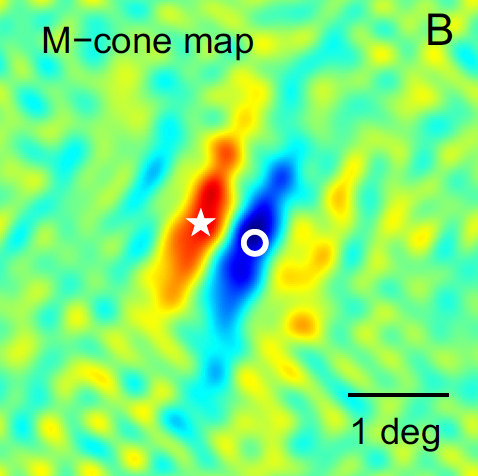}
       & \includegraphics[height=0.17\textheight]{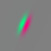}
     \end{tabular}
   \end{center}
  \caption{Example of a colour-opponent receptive field profile 
           for a double-opponent simple cell in the primary visual
           cortex (V1) as measured by Johnson {\em et al.\/}\
           \cite{JohHawSha08-JNeuroSci}.
          (left) Responses to L-cones corresponding to long wavelength
          red cones, with positive weights
          represented by red and negative weights by blue. 
          (middle) Responses to M-cones corresponding to medium wavelength
          green cones, with positive weights
          represented by red and negative weights by blue. 
          (right) Idealized model of the receptive field
          from a first-order directional derivative of an affine
          Gaussian kernel $\partial_{\varphi}g(x, y;\; \Sigma)$ 
          according to (\ref{eq-spat-RF-model}) for $\sigma_1 = \sqrt{\lambda_1} = 0.6$,
         $\sigma_2 = \sqrt{\lambda_2} = 0.2$ in units of
           degrees of visual angle, $\alpha = 157~\mbox{degrees}$ and with positive
           weights for the red-green colour-opponent channel
           $U = R-G$ with positive values represented by red and
           negative values by green.}
  \label{fig-simple-cell-aff-gauss-model-col-opp}
\end{figure*}

\begin{figure*}[hbtp]
   \begin{center}
     \begin{tabular}{c}
       \includegraphics[width=0.60\textwidth]{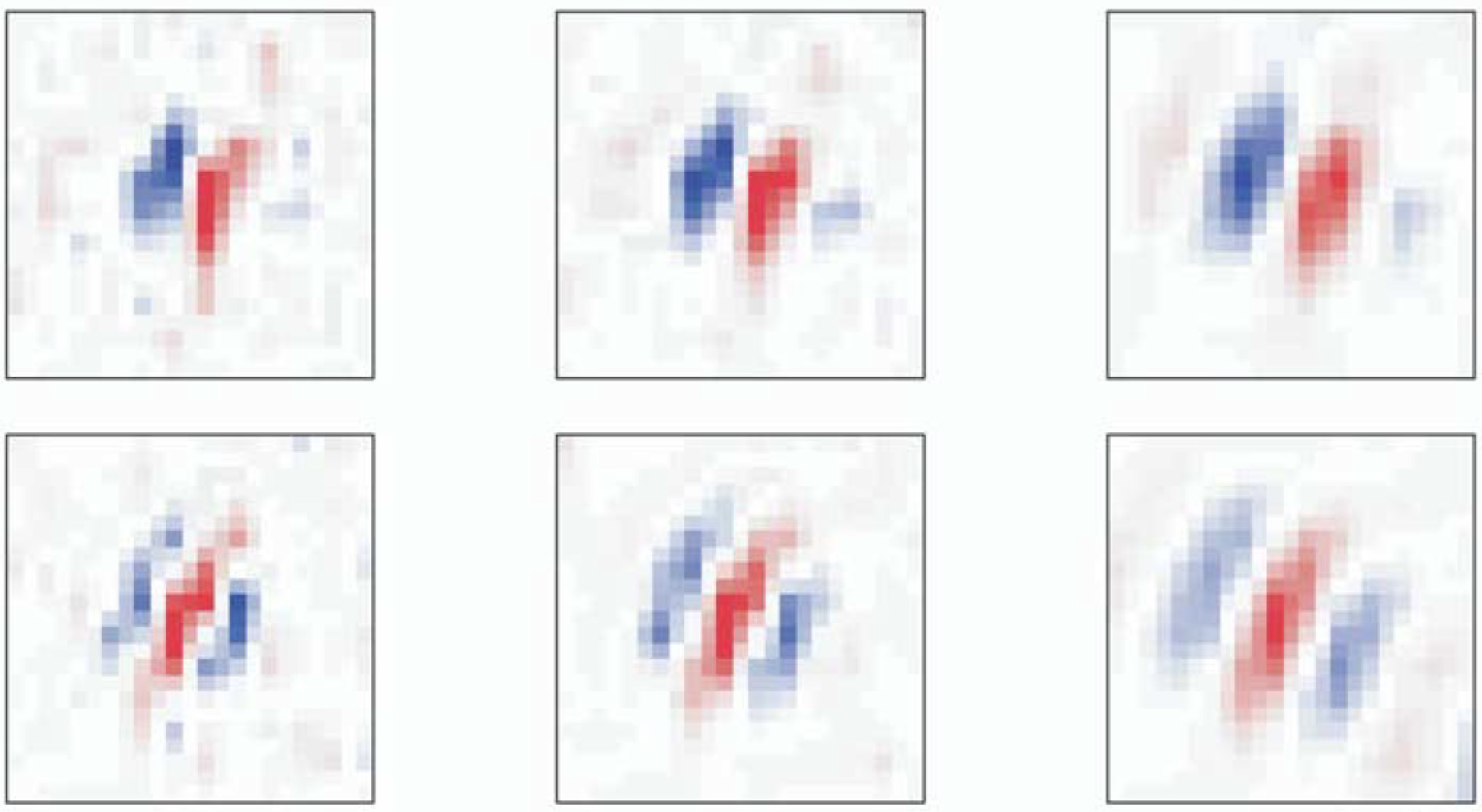}
     \end{tabular}
   \end{center}
   \caption{Significant eigenvectors of a complex cell in the cat
     primary visual cortex, as
     determined by Touryan {\em et al.\/}\
\cite{TouFelDan05-Neuron} from the response properties of the cell to a set of natural image
   stimuli, using a spike-triggered covariance method (STC) that
   computes the eigenvalues and the eigenvectors of a second-order
   Wiener kernel using three different parameter settings (cutoff
   frequencies) in the system
   identification method (from left to right).
   Qualitatively, these kernel shapes agree well with the shapes of
   first- and second-order affine Gaussian derivatives.}
  \label{fig-Touryan-eigenvectors-complex-cell}
\end{figure*}

\section{Oriented quasi quadrature modelling of complex cells}
\label{sec-ori-quasi-quad-measure}

In this section, we will consider an extension of the 1-D quasi
quadrature measure (\ref{eq-quasi-quad-1D}) into an oriented quasi
quadrature measure over 2-D image space of the form
\begin{equation}
  \label{eq-quasi-quad-dir}
  {\cal Q}_{\varphi,norm} L 
  = \sqrt{\frac{\lambda_{\varphi} \, L_{\varphi}^2 + C \, \lambda_{\varphi}^2 \, L_{\varphi\varphi}^2}{s^{\Gamma}}},
\end{equation}
where $L_{\varphi}$ and $L_{\varphi\varphi}$ denote directional
derivatives of an affine Gaussian scale-space representation
\cite{Iij63-StudInfCtrl} \cite[ch.~15]{Lin93-Dis} 
\begin{equation}
   L(\cdot;\; s, \Sigma) = g(\cdot;\; s, \Sigma) * f(\cdot)
\end{equation}
of the form 
\begin{align}
  \begin{split}
    \label{eq-1st-dir-der-scsp}
    L_{\varphi} 
     = \cos \varphi \, L_{x_1} + \sin \varphi \, L_{x_2},
  \end{split}\\ 
  \begin{split}
    \label{eq-2nd-dir-der-scsp}
     L_{\varphi\varphi} 
     = \cos^2 \varphi \, L_{x_1x_1} + 2 \cos \varphi \, \sin \varphi \, L_{x_1x_2} + \sin^2 \varphi \, L_{x_2x_2},
  \end{split}
\end{align}
and with
$\lambda_{\varphi}$ denoting the variance of the affine
Gaussian kernel (with $x = (x_1, x_2)^T$)
\begin{equation}
  \label{eq-aff-gauss-2D}
   g(x;\; s, \Sigma)  = \frac{1}{2 \pi s \sqrt{\det\Sigma}} e^{-x^T \Sigma^{-1} x/2s} 
  \end{equation}
in direction $\varphi$, preferably with the orientation $\varphi$
aligned with the direction $\alpha$ of either of the eigenvectors of the
composed spatial covariance matrix $s \, \Sigma$, with 
\begin{align}
  \begin{split}
  \label{eq-aff-cov-mat-2D}
  \Sigma =
  & \frac{1}{\max(\lambda_1, \lambda_2)} \times
  \end{split}\nonumber\\
  \begin{split}
  & \left(
    \begin{array}{ccc}
      \lambda_1 \cos^2 \alpha + \lambda_2 \sin^2 \alpha \quad
        &  (\lambda_1 - \lambda_2) \cos \alpha \, \sin \alpha 
        \\
      (\lambda_1 - \lambda_2) \cos \alpha \, \sin \alpha \quad
        & \lambda_1 \sin^2 \alpha + \lambda_2 \cos^2 \alpha
    \end{array}
  \right)
  \end{split}
\end{align}
normalized such that the main eigenvalue is equal to one.

\subsection{Affine Gaussian derivative model for linear receptive
  fields}

According to the normative theory for visual receptive fields in
Lindeberg \cite{Lin10-JMIV,Lin13-BICY,Lin13-PONE,Lin17-arXiv-norm-theory-RF},
directional derivatives of affine Gaussian kernels
constitute a canonical model for visual receptive fields over a 2-D
spatial domain.
Specifically, it was proposed that simple cells in the primary visual
cortex (V1) can be modelled by directional derivatives of affine Gaussian
kernels, termed {\em affine Gaussian derivatives\/}, of the form
\begin{equation}
  \label{eq-spat-RF-model}
   T_{{\varphi}^{m}}(x_1, x_2;\; s, \Sigma)  
  = \partial_{\varphi}^{m} 
      \left( g(x_1, x_2;\; s, \Sigma) \right).
\end{equation}
Figure~\ref{fig-simple-cell-aff-gauss-model-col-opp} shows an example of the
spatial dependency of a colour-opponent simple cell that can be well modelled by a
first-order affine Gaussian derivative over an R-G colour-opponent
channel over image intensities. 
Corresponding modelling results for non-chromatic receptive fields can be found in \cite{Lin10-JMIV,Lin13-BICY,Lin13-PONE}.

\subsection{Affine quasi quadrature modelling of complex cells}

Figure~\ref{fig-Touryan-eigenvectors-complex-cell} shows functional properties of a complex cell as
determined from its response properties to natural images, 
using a spike-triggered covariance method (STC),
which computes the eigenvalues and the eigenvectors of
a second-order Wiener kernel (Touryan {\em et al.\/}\
\cite{TouFelDan05-Neuron}).
As can be seen from this figure, the shapes of the eigenvectors 
determined from the non-linear Wiener kernel model of the complex cell
do qualitatively agree very well with the shapes of corresponding
affine Gaussian derivative kernels of orders 1 and 2.

Motivated by this property, that mathematical modelling of functional
properties of a biological complex cell in terms of a second-order energy
model reveals computational primitives similar to affine Gaussian derivatives,
combined with theoretical and experimental
motivations for modelling receptive field profiles of simple cells by
affine Gaussian derivatives, we propose to model complex cells by a
possibly post-smoothed oriented quasi quadrature measure of the
form (\ref{eq-quasi-quad-dir})
\begin{multline}
  \label{eq-quasi-quad-dir-smooth}
  (\overline{\cal Q}_{\varphi,norm} L)(\cdot;\; s_{loc}, s_{int}, \Sigma_{\varphi})
  = \\ \sqrt{g(\cdot;\; s_{int}, \Sigma_{\varphi}) 
               * ({\cal Q}^2_{\varphi,norm} L)(\cdot;\; s_{loc}, \Sigma_{\varphi})} 
\end{multline}
where $s_{loc} \,\Sigma_{\varphi}$ represents an affine covariance matrix 
in direction $\varphi$ for computing directional
derivatives and $s_{int} \, \Sigma_{\varphi}$ represents an affine covariance
matrix in the same direction for integrating pointwise affine quasi quadrature
measures over a region in image space.

The pointwise affine quasi quadrature measure in this expression
$({\cal Q}_{\varphi,norm} L)(\cdot;\; s_{loc}, \Sigma_{\varphi})$
can be seen as a Gaussian derivative based analogue
of the energy model for complex cells as proposed by 
Adelson and Bergen \cite{AdeBer85-JOSA} and Heeger \cite{Hee92-VisNeuroSci}.
It is closely related to a proposal by 
Koenderink and van Doorn \cite{KoeDoo90-BC} of summing up the squares of
first- and second-order derivative responses and
nicely compatible with results by 
De~Valois {\em et al.\/}\ \cite{ValCotMahElfWil00-VR}, who showed that
first- and second-order receptive fields typically occur in pairs that
can be modelled as approximate Hilbert pairs. 

Specifically, this pointwise differential
entity mimics some of the known properties of complex cells in the
primary visual cortex
as discovered by Hubel and Wiesel \cite{HubWie05-book}
in the sense of: (i)~being independent of the polarity of the stimuli,
(ii)~not obeying the superposition principle and 
(iii)~being rather insensitive to the phase of the visual stimuli.
The primitive components of the quasi quadrature measure (the directional
derivatives) do in turn mimic some of the known properties of simple cells in the
primary visual cortex in terms of: (i)~precisely localized ``on'' and ``off''
subregions with (ii)~spatial summation within each subregion,
(iii)~spatial antagonism between on- and off-subregions and
(iv)~whose visual responses to stationary or moving spots can be
predicted from the spatial subregions.

The addition of a complementary post-smoothing stage in (\ref{eq-quasi-quad-dir-smooth}) as
determined by the affine Gaussian weighting function $g(\cdot;\; s_{int}, \Sigma_{\varphi})$
is closely related to recent results by West{\"o} and May \cite{WesMay18-JNeuroPhys},
who have shown that complex cells are better modelled as a combination
of two spatial integration steps than a single spatial integration.
This spatial post-smoothing stage, which serves as a spatial pooling
operation, does additionally decrease the spatial sensitivity of the
pointwise quasi quadrature measure and makes it more robust to local
spatial perturbations.

By choosing these spatial smoothing
and weighting functions as affine Gaussian kernels, we ensure an affine-covariant model of the
complex cells, to enable the computation of affine invariants at higher
levels in the visual hierarchy.

\begin{figure}[!h]
   \begin{center}
     \begin{tabular}{c}
       \includegraphics[width=0.40\textwidth]{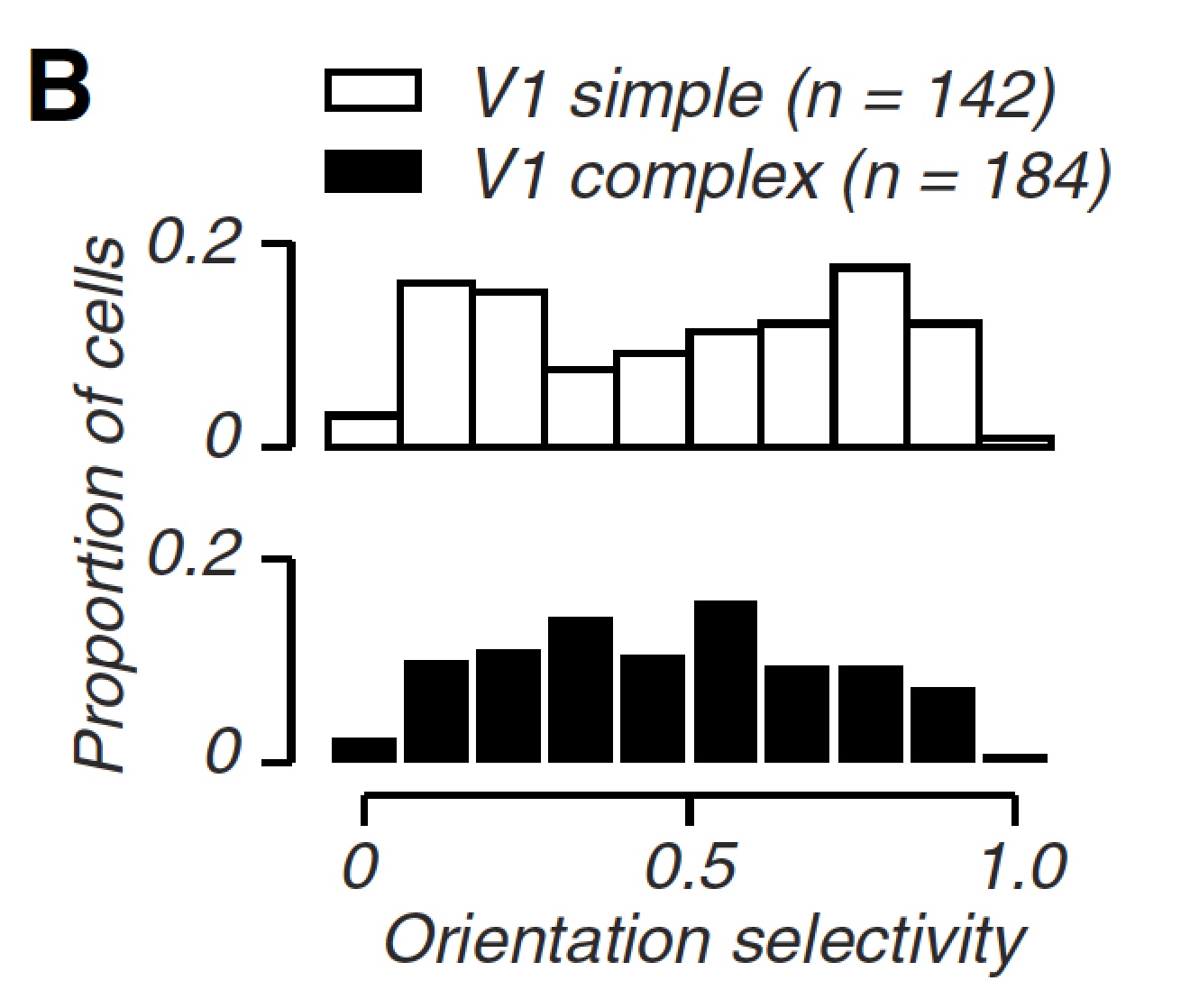}
     \end{tabular}
   \end{center}
   \caption{Statistics of the orientation selectivity of simple cells
     and complex cells in the primary visual cortex of the Macaque monkey
    as reported by Goris {\em et al.\/}\ \cite{GorSimMov15-Neuron}.
    With respect to the affine Gaussian derivative model for the
    receptive fields of simple
    and complex cells, the large variability in orientation selectivity
    reported from these biological measurements implies that we should
    consider derivatives of affine Gaussian kernels for a large
    variability in the eccentricity of their shapes, as can be
    parameterized by {\em e.g.\/}\ the ratio between the eigenvalues
    $\lambda_1$ and $\lambda_2$ of the affine covariance matrix $s \,
    \Sigma$. (A highly eccentric affine Gaussian derivative kernel
    will have more narrow orientation selectivity.)}
  \label{fig-DeValois-orient-select-stat}
\end{figure}

The use of multiple affine receptive fields over different shapes of the affine
covariance matrices $\Sigma_{\varphi,loc}$ and $\Sigma_{\varphi,int}$ can
be motivated by results by Goris {\em et al.\/}\
\cite{GorSimMov15-Neuron}, who show that there is a large variability
in the orientation selectivity of simple and complex cells
(see Figure~\ref{fig-DeValois-orient-select-stat}). 
With respect
to this model, this means that we can think of affine covariance
matrices of different eccentricities as being present from isotropic to
highly eccentric. By considering the full family of positive definite
affine covariance matrices, we obtain a fully affine-covariant
image representation able to handle local linearizations of the perspective
mapping for all possible views of any smooth local surface patch.

\begin{figure*}[!p]
   \begin{center}
     \begin{tabular}{c}
       \includegraphics[height=0.80\textheight]{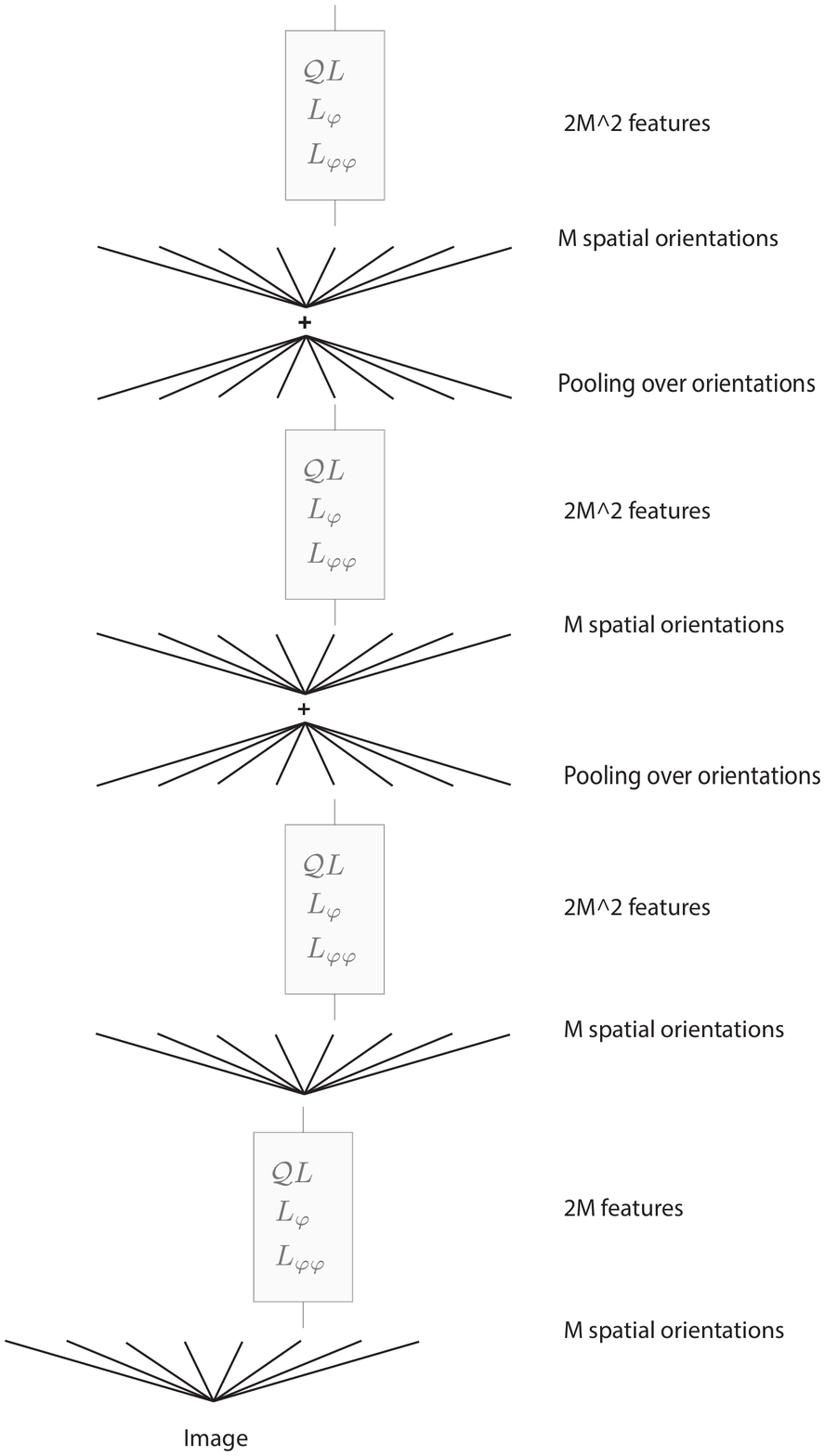}
     \end{tabular}
   \end{center}
   \caption{Schematic illustration of how the quasi quadrature network
   is constructed from an image, here with a total number of 4 layers.
   In the first layer, there is an expansion over all $M = 8$
   orientations, leading to a total number of $2 M$ independent
   features $L_{\varphi}$ and $L_{\varphi\varphi}$ over all $M$ image
   orientations from which the dependent feature ${\cal Q} L$ is then computed
   according to (\ref{eq-quasi-quad-dir}).
   In the second layer, the maps of ${\cal Q} L$ for all the $M$ image
   orientations are used for another expansion over image
   orientations, such that a total number of $2 M^2$ independent features 
   $L_{\varphi}$ and $L_{\varphi\varphi}$ is
   computed over all pairs of image orientations.
   To delimit the complexity of the features in higher layers, there
   is a pooling stage over image orientations by summing up the quasi
   quadrature responses over all the image orientations before further
   expansions over image orientations are perform at layer $K =
   3$. Thereby, the number of independent features in
   these layers is delimited by $2 M^2$ instead of $2 M^3$. By a
   corresponding pooling stage before layer 4, the number of
   independent features
   in this layer is also delimited by $2 M^2$.
   (The grey boxes, which show the independent features
   $L_{\varphi}$ and $L_{\varphi\varphi}$ and the dependent feature
   ${\cal Q} L$ that are computed in every layer in the hierarchy, are
   here only shown for one of the several possible paths through the
   hierarchy. The combinatorial expansion in layer 2 is also only
   shown for one of the $M$ orientations in layer 1.)}
   \label{fig-quasiquadnet}
\end{figure*}

With respect to computational modelling of biological vision, the proposed
affine quasi quadrature model constitutes a novel functional model of
complex cells as previously studied in biological vision by
Hubel and Wiesel \cite{HubWie59-Phys,HubWie62-Phys,HubWie05-book},
Movshon {\em et al.\/}\ \cite{MovThoTol78-JPhys}, 
Emerson {\em et al.\/}\ \cite{EmeCitVauKle87-JNeuroPhys}, 
Touryan {\em et al.\/}\ \cite{TouLauDan02-JNeuroSci,TouFelDan05-Neuron}
and Rust {\em et al.\/}\ \cite{RusSchMovSim05-Neuron},
and modelled computationally by Adelson and Bergen
\cite{AdeBer85-JOSA}, Heeger \cite{Hee92-VisNeuroSci},
Serre and Riesenhuber \cite{SerRie04-AIMemo},
Einh{\"a}user {\em et  al.\/} \cite{EinKayKoeKoe02-EurJNeurSci},
Kording {\em et al.\/}\ \cite{KorKayWinKon04-JNeuroPhys},
Merolla and Boahen \cite{MerBoa04-NIPS},
Berkes and Wiscott \cite{BerWis05-JVis},
Carandini \cite{Car06-JPhys} 
and Hansard and Horaud \cite{HanHor11-NeurComp}.
A conceptual novelty of our model, which emulates
several of the known properties of complex cells although our
understanding of the non-linearities of complex cells is still
limited, is that it is fully expressed based on the mathematically derived
affine Gaussian derivative model for visual receptive fields
\cite{Lin13-BICY} and therefore possible to relate to natural image
transformations as modelled by affine transformations over the
spatial domain.

In the following, we will use this quasi quadrature model
of complex cells for constructing continuous hierarchical networks.

\section{Hierarchies of oriented quasi quadrature measures}
\label{sec-hier-quasi-quad-net}

Let us in this first study henceforth for simplicity disregard the
variability due to different shapes of
the affine receptive fields for different eccentricities and assume
that $\Sigma = I$.

This restriction enables covariance to scaling transformations and rotations,
whereas a full treatment of affine quasi quadrature measures over all
positive definite covariance matrices for the underlying affine
Gaussian smoothing operation
would enable full affine covariance.

An approach that we shall pursue is to build feature
hierarchies by coupling oriented quasi quadrature measures
(\ref{eq-quasi-quad-dir}) or (\ref{eq-quasi-quad-dir-smooth}) in
cascade%
\footnote{If using raw quasi quadrature measures of the form
  (\ref{eq-quasi-quad-dir}) when constructing the hierarchical
  representation, the Gaussian spatial smoothing operation, underlying
  the computation of the Gaussian derivatives from which the quasi
  quadrature measure is computed, implies that a certain amount of
  spatial integration (spatial pooling) is guaranteed to be performed
  in the transformation between successive layers. If the
  post-smoothed quasi quadrature measure
  (\ref{eq-quasi-quad-dir-smooth}) is instead used for constructing
  the feature hierarchy, then the spatial post-smoothing operation in
  the post-smoothed quasi quadrature measure guarantees that a certain
  amount of spatial integration (spatial pooling) is also guaranteed
  in the quasi quadrature measure computed in any layer.}
\begin{align}
  \begin{split}
    \label{eq-hier-quasi-quad-line1}
     & F_1(x, \varphi_1) = ({\cal Q}_{\varphi_1,norm} \, L)(x) 
  \end{split}\\
  \begin{split}
    \label{eq-hier-quasi-quad-line2}
      & F_k(x, \varphi_1, ..., \varphi_{k-1}, \varphi_k) = 
      ({\cal Q}_{\varphi_k,norm} \, F_{k-1})(x, \varphi_1, ..., \varphi_{k-1}),
   \end{split}
\end{align}
where we have suppressed the notation for the scale
levels assumed to be distributed such that the scale
parameter at level $k$ is $s_k = s_0 \, r^{2(k-1)}$ for some 
$r > 1$, {\em e.g.}, $r = 2$. 
Assuming that the initial scale-space representation $L$ is
computed at scale $s_0$, such a network can in turn be initiated for
different values of $s_0$, also distributed according to a geometric distribution.

This construction builds upon an early proposal by Fuku\-shima \cite{Fuk80-BICY}
of building a hierarchical neural network from repeated application
of models of simple and complex cells \cite{HubWie59-Phys,HubWie62-Phys,HubWie05-book},
which has later been explored in a hand-crafted network based on
Gabor functions by Riesenhuber and Poggio \cite{RiePog99-Nature} and Serre {\em et al.\/}\ \cite{SerWolBilRiePog07-PAMI}
and in the scattering convolution networks by Bruna and Mallat
\cite{BruMal13-PAMI}. This idea is also consistent with a proposal by
Yamins and DiCarlo \cite{YamDiC16-NatNeuroSci} of using repeated application 
of a single hierarchical convolution layer for explaining the
computations in the mammalian cortex.
With this construction, we obtain a way to
define continuous networks that express a corresponding hierarchical
architecture based on Gaussian derivative based models of simple and
complex cells within the scale-space framework.

Each new layer in this model implies an expansion of combinations
of angles over the different layers in the hierarchy. For example, if
we in a discrete implementation discretize the angles $\varphi \in [0, \pi[$ into $M$ discrete
spatial orientations, we will then obtain $M^k$ different features at
level $k$ in the hierarchy. To keep the complexity down at higher
levels, we will for $k \geq K$ in a corresponding way as done by
Hadji and Wildes \cite{HadWil17-ICCV} introduce a pooling stage 
over orientations 
\begin{equation}
  \label {eq-orient-pooling}
  ({\cal P}_k F_{k})(x, \varphi_1, ..., \varphi_{K-1}) 
   = \sum_{\varphi_k} F_k(x, \varphi_1, ..., \varphi_{K-1}, \varphi_k),
\end{equation}
which sums up the responses for all the orientations in the current layer,
before the next successive layer is instead defined by applying
oriented quasi quadrature measures to the pooled responses
\begin{multline}
     F_k(x, \varphi_1, ..., \varphi_{k-2}, \varphi_{K-1}, \varphi_k) = \\
     ({\cal Q}_{\varphi_k,norm} \, {\cal P}_{k-1} F_{k-1})(x, \varphi_1, ..., \varphi_{K-1}).
\end{multline}
In this way, the number of features at any level will be limited to maximally
$M^{K-1}$. The proposed hierarchical feature representation 
is termed QuasiQuadNet.

Figure~\ref{fig-quasiquadnet} gives a schematic illustration of the structure
of such a resulting hierarchy using an expansion over $M = 8$ spatial
orientations in the image domain over a total number of 4 layers with
the combinatorial expansion over image
orientations delimited from layer $K = 3$.

\begin{figure*}[hbtp]
  \begin{center}
    \begin{tabular}{cccccc}
        {\footnotesize\em image\/} 
           & {\footnotesize\em $\varphi_1 = 0$\/} 
           & {\footnotesize\em $\varphi_1 = \tfrac{\pi}{4}$\/} 
           & {\footnotesize\em $\varphi_1 = \tfrac{\pi}{2}$\/}
           & {\footnotesize\em $\varphi_1 = \tfrac{3 \pi}{4}$\/} \\
        \includegraphics[width=0.15\textwidth]{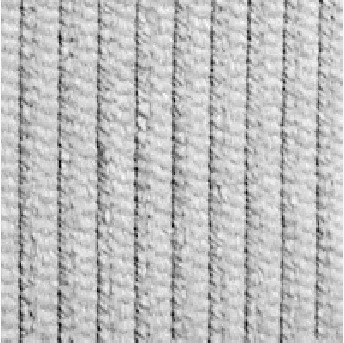}
           & \includegraphics[width=0.15\textwidth]{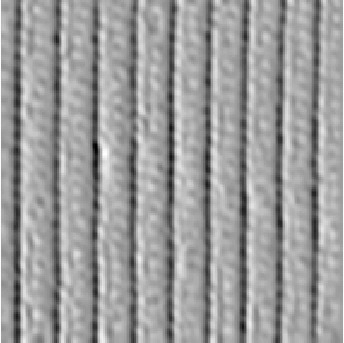}
           & \includegraphics[width=0.15\textwidth]{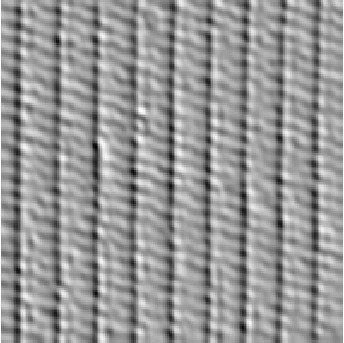}
           & \includegraphics[width=0.15\textwidth]{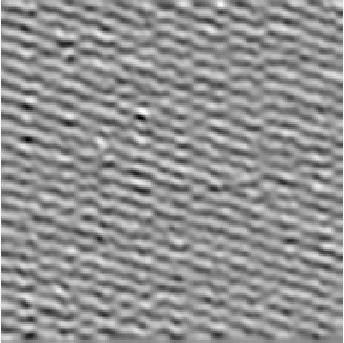}
           & \includegraphics[width=0.15\textwidth]{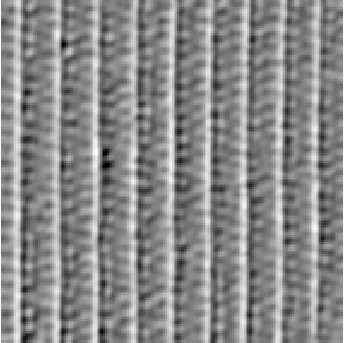} 
           & $L_{\varphi_1}$ \\
       & \includegraphics[width=0.15\textwidth]{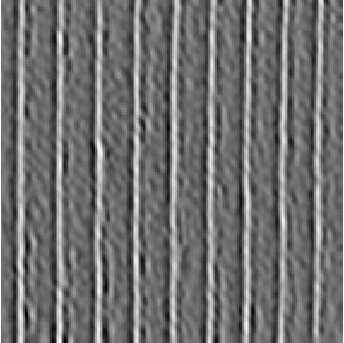}
           & \includegraphics[width=0.15\textwidth]{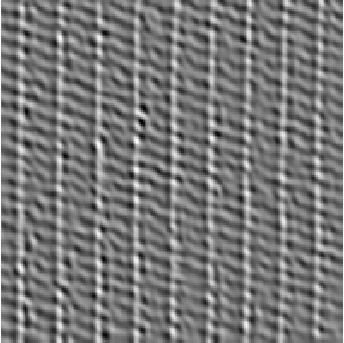}
           & \includegraphics[width=0.15\textwidth]{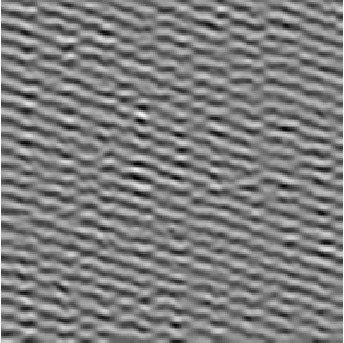}
           & \includegraphics[width=0.15\textwidth]{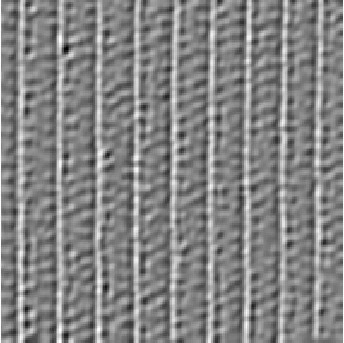} 
           & $L_{\varphi_1\varphi_1}$ \\
       & \includegraphics[width=0.15\textwidth]{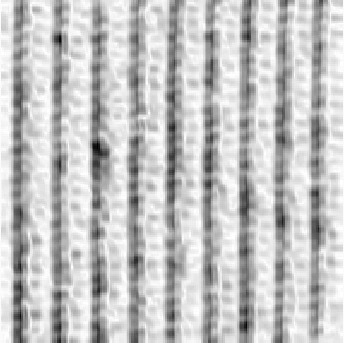}
           & \includegraphics[width=0.15\textwidth]{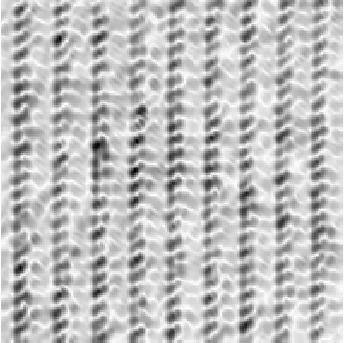}
           & \includegraphics[width=0.15\textwidth]{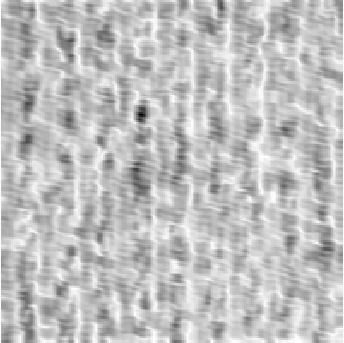}
           & \includegraphics[width=0.15\textwidth]{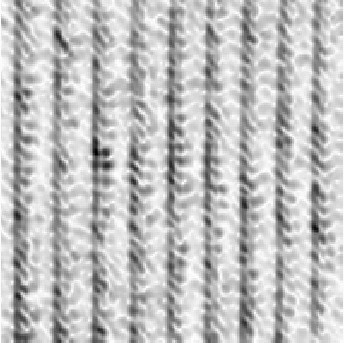} 
           & ${\cal Q}_{\varphi_1} L$ \\
       & {\footnotesize\em $\varphi_1 = 0$, $\varphi_2 = 0$\/} 
           & {\footnotesize\em $\varphi_1 = 0$, $\varphi_2 = \tfrac{\pi}{4}$\/} 
           & {\footnotesize\em $\varphi_1 = 0$, $\varphi_2 = \tfrac{\pi}{2}$\/}
           & {\footnotesize\em $\varphi_1 = 0$, $\varphi_2 = \tfrac{3 \pi}{4}$\/} \\
       & \includegraphics[width=0.15\textwidth]{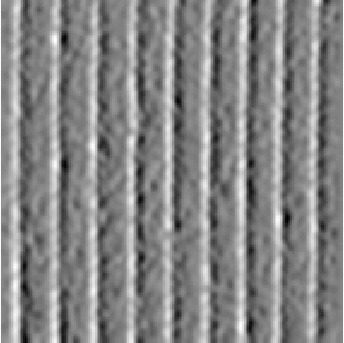}
           & \includegraphics[width=0.15\textwidth]{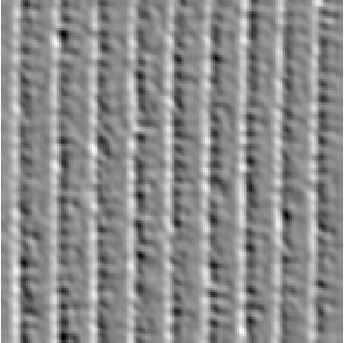}
           & \includegraphics[width=0.15\textwidth]{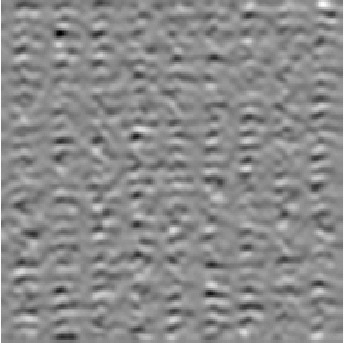}
           & \includegraphics[width=0.15\textwidth]{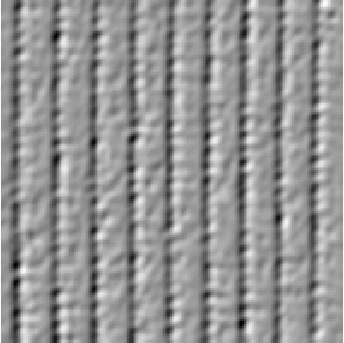} 
           & $\overline{\partial}_{\varphi_2} {\cal Q}_{\varphi_1} L$ \\
       & \includegraphics[width=0.15\textwidth]{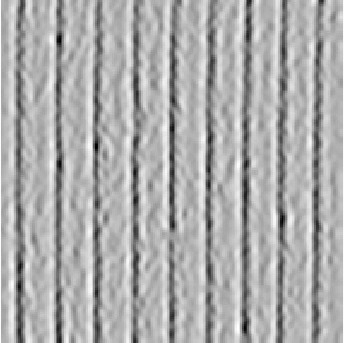}
           & \includegraphics[width=0.15\textwidth]{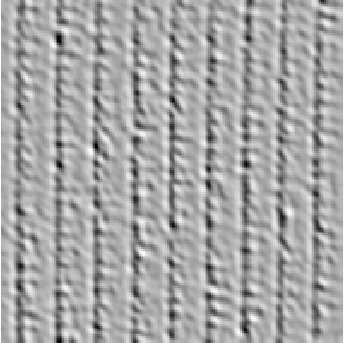}
           & \includegraphics[width=0.15\textwidth]{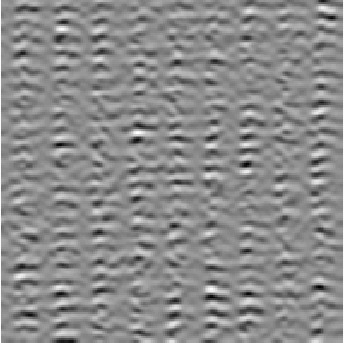}
           & \includegraphics[width=0.15\textwidth]{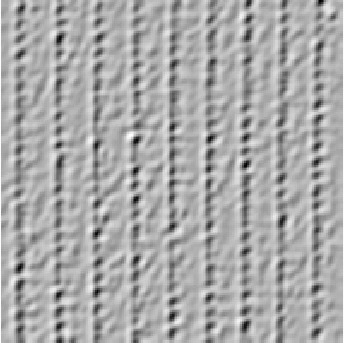} 
           & $\overline{\partial}_{\varphi_2 \varphi_2} {\cal Q}_{\varphi_1} L$ \\
       & \includegraphics[width=0.15\textwidth]{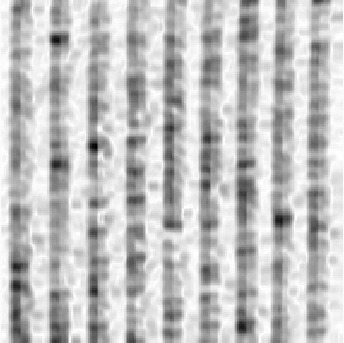}
           & \includegraphics[width=0.15\textwidth]{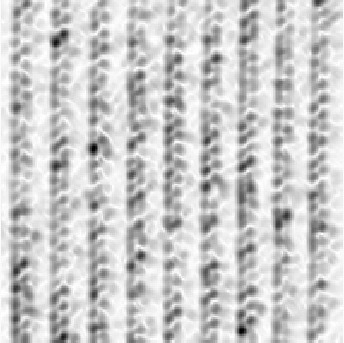}
           & \includegraphics[width=0.15\textwidth]{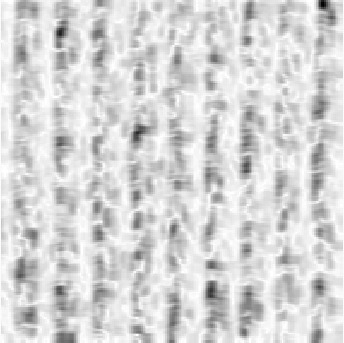}
           & \includegraphics[width=0.15\textwidth]{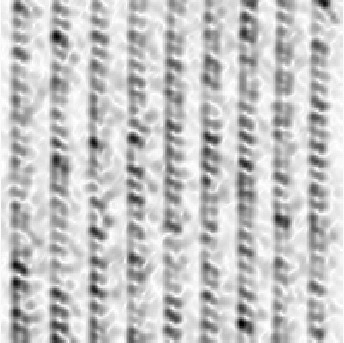} 
           & ${\cal Q}_{\varphi_2} {\cal Q}_{\varphi_1} L$ \\   
        & {\footnotesize\em $\varphi_1 = 0$, $\varphi_3 = 0$\/} 
           & {\footnotesize\em $\varphi_1 = 0$, $\varphi_3 = \tfrac{\pi}{4}$\/} 
           & {\footnotesize\em $\varphi_1 = 0$, $\varphi_3 = \tfrac{\pi}{2}$\/}
           & {\footnotesize\em $\varphi_1 = 0$, $\varphi_3 = \tfrac{3 \pi}{4}$\/} \\
      & \includegraphics[width=0.15\textwidth]{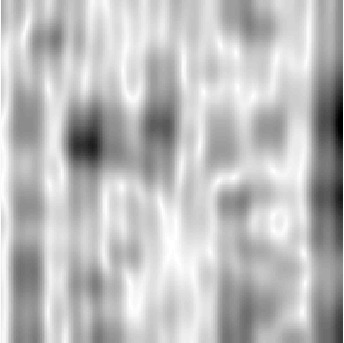}
           & \includegraphics[width=0.15\textwidth]{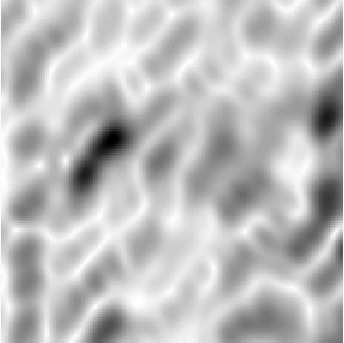}
           & \includegraphics[width=0.15\textwidth]{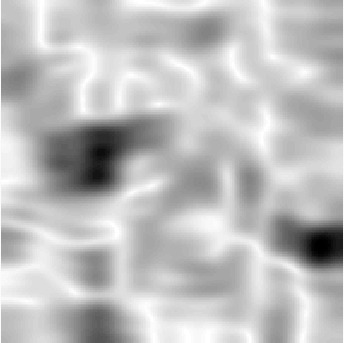}
           & \includegraphics[width=0.15\textwidth]{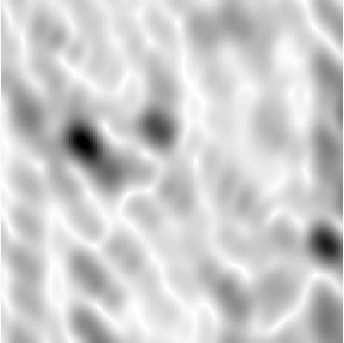} 
           & ${\cal Q}_{\varphi_3} {\cal P}_{\varphi_2} {\cal Q}_{\varphi_2} {\cal Q}_{\varphi_1} L$ \\      
\end{tabular} 
  \end{center}
  \caption{Subset of features in a hierarchy of directional derivatives and quasi quadrature measures
    computed from a texture image (corduroy/sample\_a/42a-scale\_4\_im\_5\_col.png) from
  the KTH-TIPS2 dataset \cite{KTH-TIPS2} for different
  combinations of angles in layers 1 and 2 for $s_0 = 2$.
  (top row) Original image and first-order directional derivatives
  $L_{\varphi}$.
  (second row) Second-order directional derivatives
  $L_{\varphi\varphi}$.
  (third row) Oriented quasi quadrature measures ${\cal Q}_{\varphi_1}
  L$ in layer 1.
  (fourth row) First-order directional derivatives computed from
  ${\cal Q}_{\varphi_1} L$ for $\varphi_1 = 0$.
  (fifth row) Second-order directional derivatives computed from
  ${\cal Q}_{\varphi_1} L$ for $\varphi_1 = 0$.
  (sixth row) Oriented quasi quadrature measures ${\cal Q}_{\varphi_2} {\cal Q}_{\varphi_1} L$ in layer 2 
  for $\varphi_1 = 0$.
  (bottom row) Oriented quasi quadrature measures 
  ${\cal Q}_{\varphi_3} {\cal P}_{\varphi_2} {\cal Q}_{\varphi_2} {\cal Q}_{\varphi_1} L$ in layer 3 
  for $\varphi_1 = 0$.
  (The contrast has been reversed for the quasi quadrature measures so
that high values are shown as dark and low values as bright.) (Image
size: $200 \times 200$ pixels.)}
  \label{fig-corduroy-quasiquadnet}
\end{figure*}

\begin{figure*}[hbtp]
  \begin{center}
    \begin{tabular}{cccccc}
        {\footnotesize\em image\/} 
           & {\footnotesize\em $\varphi_1 = 0$\/} 
           & {\footnotesize\em $\varphi_1 = \tfrac{\pi}{4}$\/} 
           & {\footnotesize\em $\varphi_1 = \tfrac{\pi}{2}$\/}
           & {\footnotesize\em $\varphi_1 = \tfrac{3 \pi}{4}$\/} \\
        \includegraphics[width=0.15\textwidth]{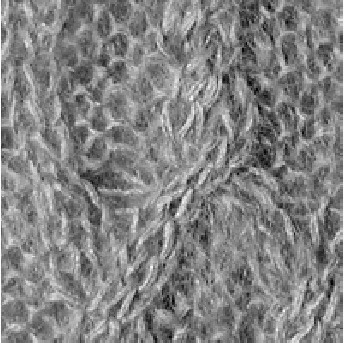}
           & \includegraphics[width=0.15\textwidth]{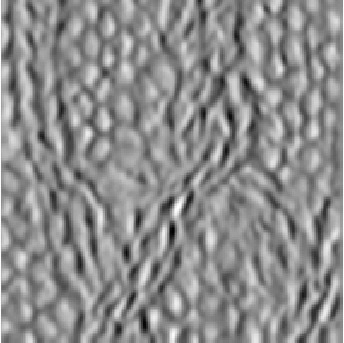}
           & \includegraphics[width=0.15\textwidth]{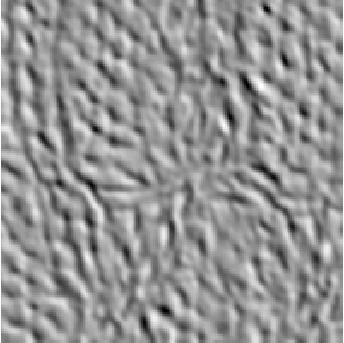}
           & \includegraphics[width=0.15\textwidth]{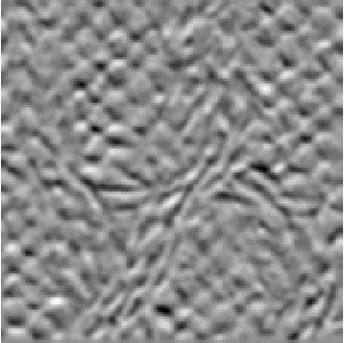}
           & \includegraphics[width=0.15\textwidth]{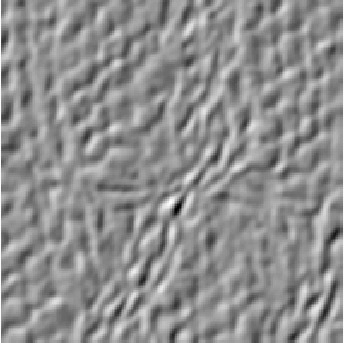} 
           & $L_{\varphi_1}$ \\
       & \includegraphics[width=0.15\textwidth]{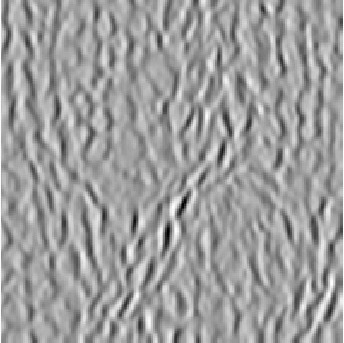}
           & \includegraphics[width=0.15\textwidth]{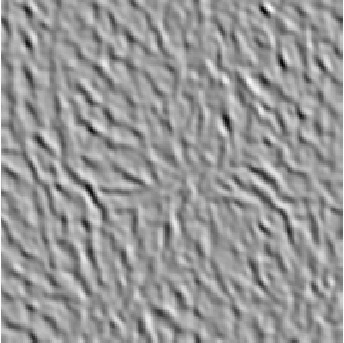}
           & \includegraphics[width=0.15\textwidth]{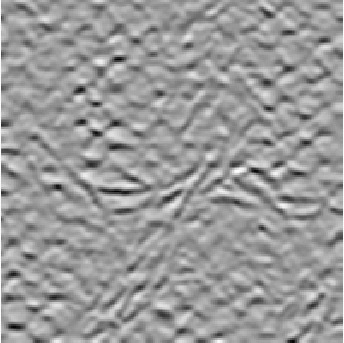}
           & \includegraphics[width=0.15\textwidth]{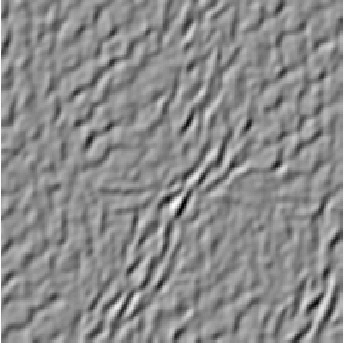} 
           & $L_{\varphi_1\varphi_1}$ \\
       & \includegraphics[width=0.15\textwidth]{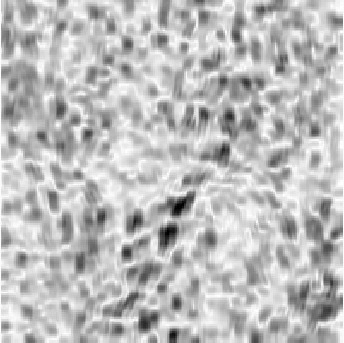}
           & \includegraphics[width=0.15\textwidth]{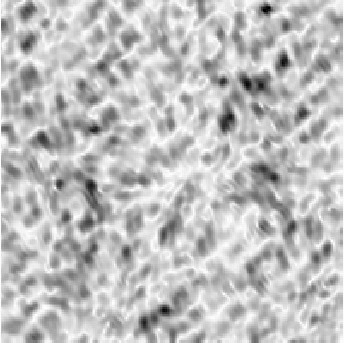}
           & \includegraphics[width=0.15\textwidth]{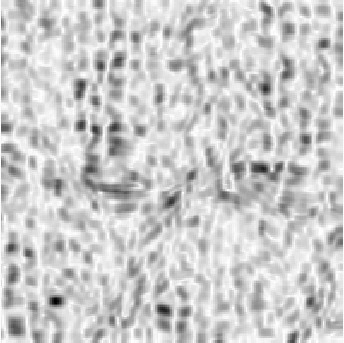}
           & \includegraphics[width=0.15\textwidth]{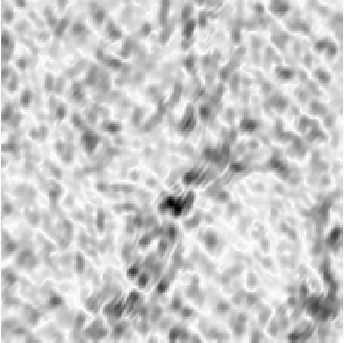} 
           & ${\cal Q}_{\varphi_1} L$ \\
       & {\footnotesize\em $\varphi_1 = 0$, $\varphi_2 = 0$\/} 
           & {\footnotesize\em $\varphi_1 = 0$, $\varphi_2 = \tfrac{\pi}{4}$\/} 
           & {\footnotesize\em $\varphi_1 = 0$, $\varphi_2 = \tfrac{\pi}{2}$\/}
           & {\footnotesize\em $\varphi_1 = 0$, $\varphi_2 = \tfrac{3 \pi}{4}$\/} \\
       & \includegraphics[width=0.15\textwidth]{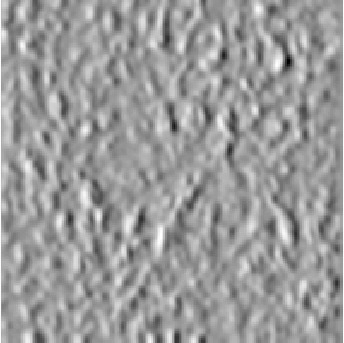}
           & \includegraphics[width=0.15\textwidth]{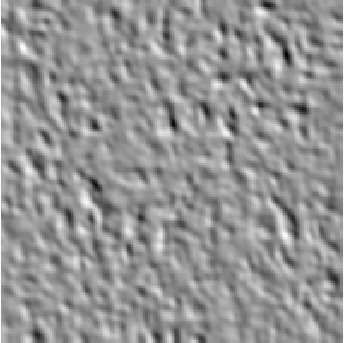}
           & \includegraphics[width=0.15\textwidth]{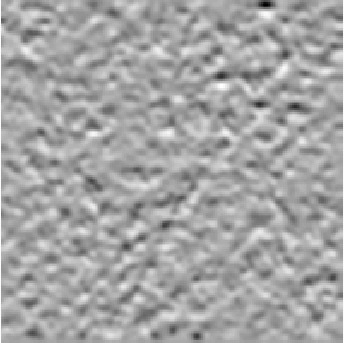}
           & \includegraphics[width=0.15\textwidth]{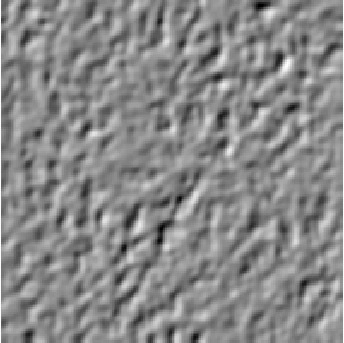} 
           & $\overline{\partial}_{\varphi_2} {\cal Q}_{\varphi_1} L$ \\
       & \includegraphics[width=0.15\textwidth]{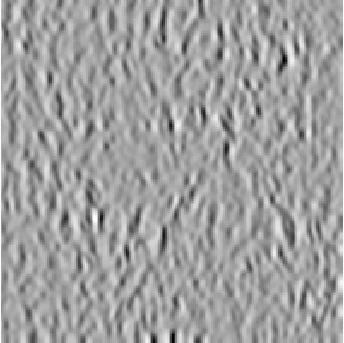}
           & \includegraphics[width=0.15\textwidth]{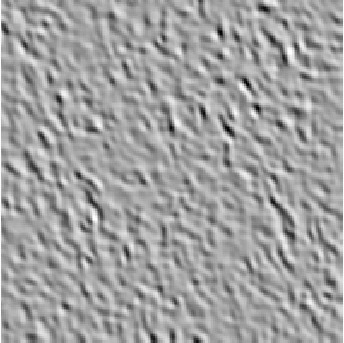}
           & \includegraphics[width=0.15\textwidth]{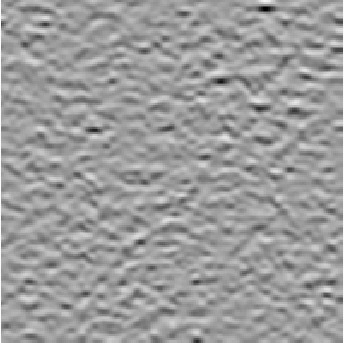}
           & \includegraphics[width=0.15\textwidth]{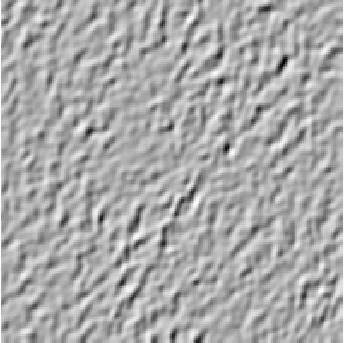} 
           & $\overline{\partial}_{\varphi_2 \varphi_2} {\cal Q}_{\varphi_1} L$ \\
       & \includegraphics[width=0.15\textwidth]{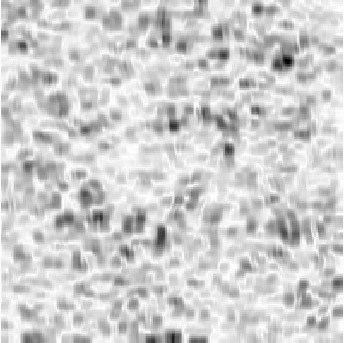}
           & \includegraphics[width=0.15\textwidth]{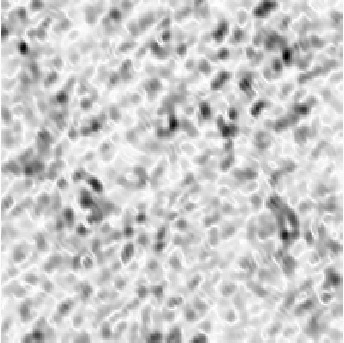}
           & \includegraphics[width=0.15\textwidth]{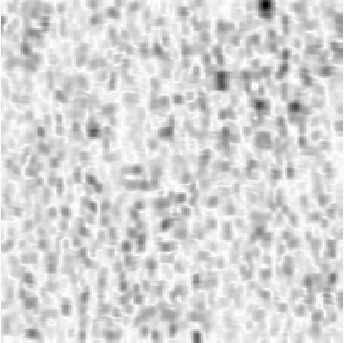}
           & \includegraphics[width=0.15\textwidth]{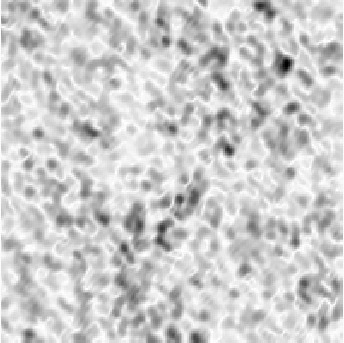} 
           & ${\cal Q}_{\varphi_2} {\cal Q}_{\varphi_1} L$ \\   
       & {\footnotesize\em $\varphi_1 = 0$, $\varphi_3 = 0$\/} 
           & {\footnotesize\em $\varphi_1 = 0$, $\varphi_3 = \tfrac{\pi}{4}$\/} 
           & {\footnotesize\em $\varphi_1 = 0$, $\varphi_3 = \tfrac{\pi}{2}$\/}
           & {\footnotesize\em $\varphi_1 = 0$, $\varphi_3 = \tfrac{3\pi}{4}$\/} \\
       & \includegraphics[width=0.15\textwidth]{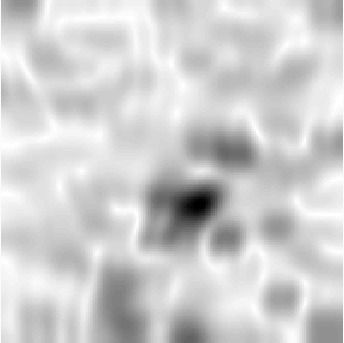}
           & \includegraphics[width=0.15\textwidth]{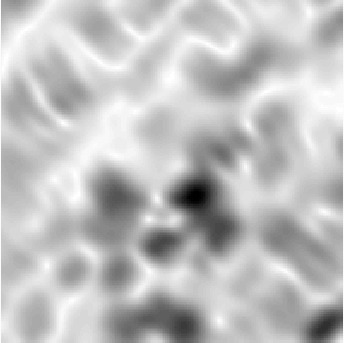}
           & \includegraphics[width=0.15\textwidth]{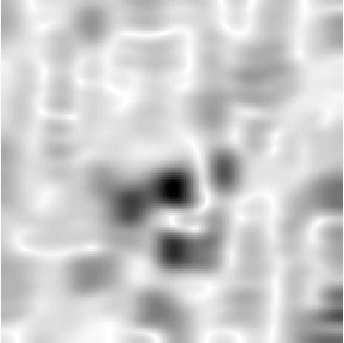}
           & \includegraphics[width=0.15\textwidth]{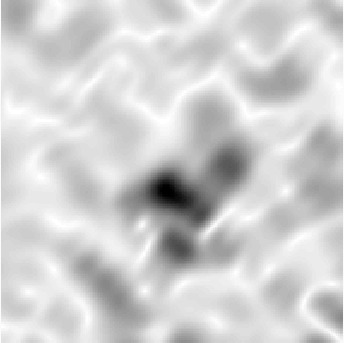} 
           & ${\cal Q}_{\varphi_3} {\cal P}_{\varphi_2} {\cal Q}_{\varphi_2} {\cal Q}_{\varphi_1} L$ \\ 
    \end{tabular} 
  \end{center}
  \caption{Subset of features in a hierarchy of directional derivatives and quasi quadrature measures
    computed from a texture image (wool/sample\_a/22a-scale\_7\_im\_3\_col.png) from
  the KTH-TIPS2 dataset \cite{KTH-TIPS2} for different
  combinations of angles in layers 1 and 2 for $s_0 = 2$.
  (top row) Original image and first-order directional derivatives
  $L_{\varphi}$.
  (second row) Second-order directional derivatives
  $L_{\varphi\varphi}$.
  (third row) Oriented quasi quadrature measures ${\cal Q}_{\varphi_1}
  L$ in layer 1.
  (fourth row) First-order directional derivatives computed from
  ${\cal Q}_{\varphi_1} L$ for $\varphi_1 = 0$.
  (fifth row) Second-order directional derivatives computed from
  ${\cal Q}_{\varphi_1} L$ for $\varphi_1 = 0$.
  (sixth row) Oriented quasi quadrature measures 
  ${\cal Q}_{\varphi_2} {\cal Q}_{\varphi_1} L$ in layer 2
  for $\varphi_1 = 0$.
  (bottom row) Oriented quasi quadrature measures 
  ${\cal Q}_{\varphi_3} {\cal P}_{\varphi_2} {\cal Q}_{\varphi_2} {\cal Q}_{\varphi_1} L$ in layer 3
  for $\varphi_1 = 0$.
  (The contrast has been reversed for the quasi quadrature measures so
that high values are shown as dark and low values as bright.) (Image
size: $200 \times 200$ pixels.)}
  \label{fig-wool-quasiquadnet}
\end{figure*}

\begin{figure*}[hbtp]
  \begin{center}
    \begin{tabular}{cccccc}
        {\footnotesize\em image\/} 
           & {\footnotesize\em $\varphi_1 = 0$\/} 
           & {\footnotesize\em $\varphi_1 = \tfrac{\pi}{4}$\/} 
           & {\footnotesize\em $\varphi_1 = \tfrac{\pi}{2}$\/}
           & {\footnotesize\em $\varphi_1 = \tfrac{3 \pi}{4}$\/} \\
        \includegraphics[width=0.16\textwidth]{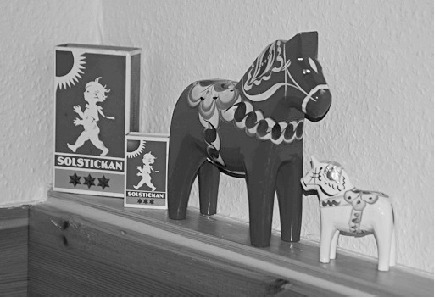}
           & \includegraphics[width=0.16\textwidth]{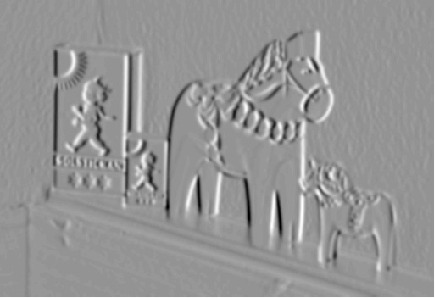}
           & \includegraphics[width=0.16\textwidth]{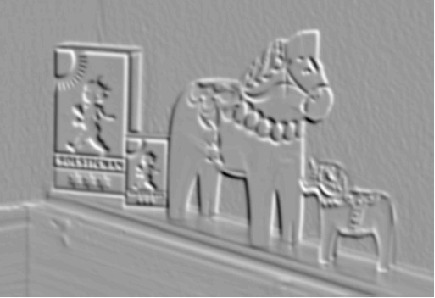}
           & \includegraphics[width=0.16\textwidth]{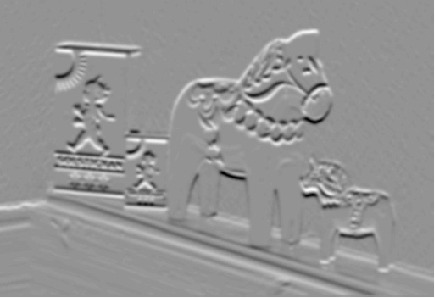}
           & \includegraphics[width=0.16\textwidth]{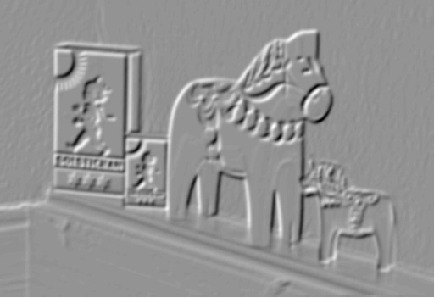} 
           & $L_{\varphi_1}$ \\
       & \includegraphics[width=0.16\textwidth]{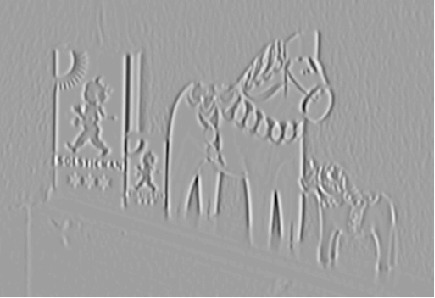}
           & \includegraphics[width=0.16\textwidth]{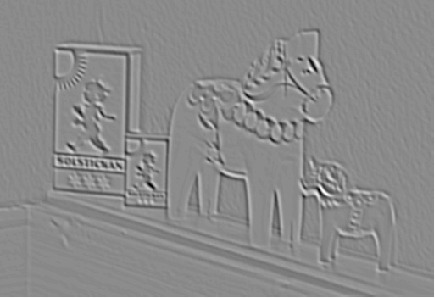}
           & \includegraphics[width=0.16\textwidth]{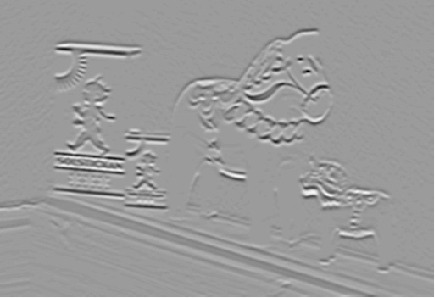}
           & \includegraphics[width=0.16\textwidth]{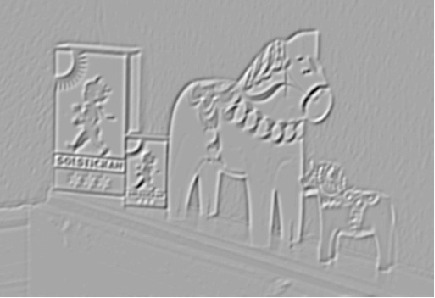} 
           & $L_{\varphi_1\varphi_1}$ \\
       & \includegraphics[width=0.16\textwidth]{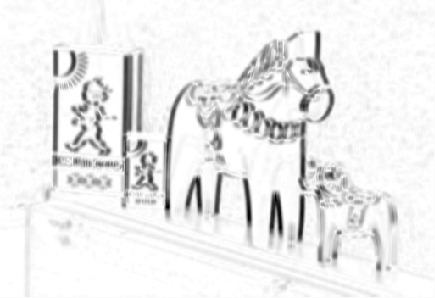}
           & \includegraphics[width=0.16\textwidth]{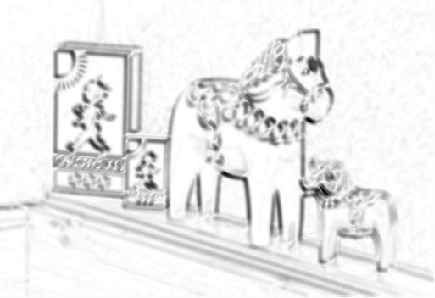}
           & \includegraphics[width=0.16\textwidth]{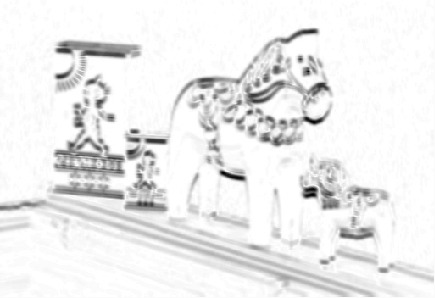}
           & \includegraphics[width=0.16\textwidth]{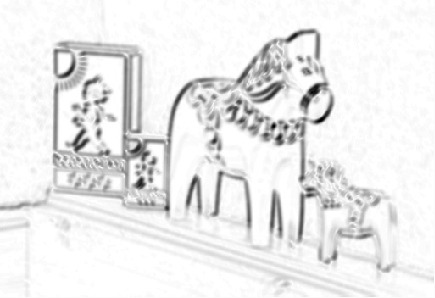} 
           & ${\cal Q}_{\varphi_1} L$ \\
       & {\footnotesize\em $\varphi_1 = 0$, $\varphi_2 = 0$\/} 
           & {\footnotesize\em $\varphi_1 = 0$, $\varphi_2 = \tfrac{\pi}{4}$\/} 
           & {\footnotesize\em $\varphi_1 = 0$, $\varphi_2 = \tfrac{\pi}{2}$\/}
           & {\footnotesize\em $\varphi_1 = 0$, $\varphi_2 = \tfrac{3 \pi}{4}$\/} \\
       & \includegraphics[width=0.16\textwidth]{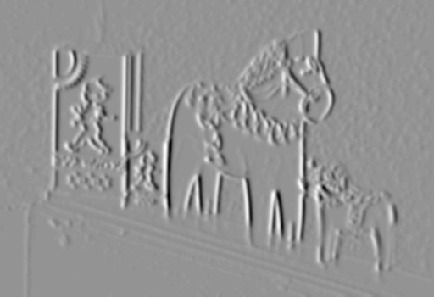}
           & \includegraphics[width=0.16\textwidth]{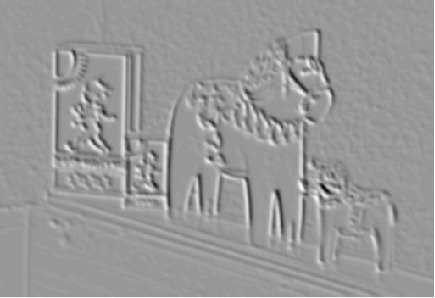}
           & \includegraphics[width=0.16\textwidth]{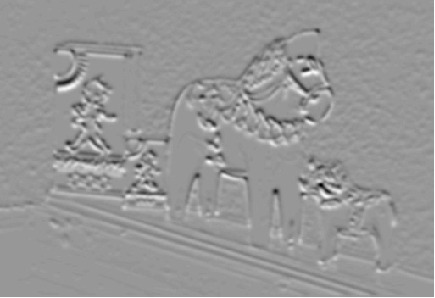}
           & \includegraphics[width=0.16\textwidth]{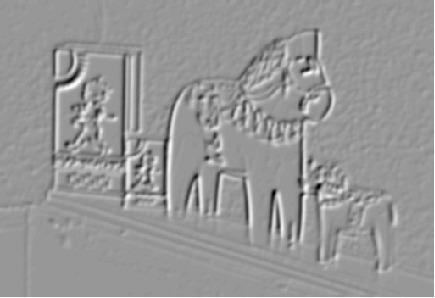} 
           & $\overline{\partial}_{\varphi_2} {\cal Q}_{\varphi_1} L$ \\
       & \includegraphics[width=0.16\textwidth]{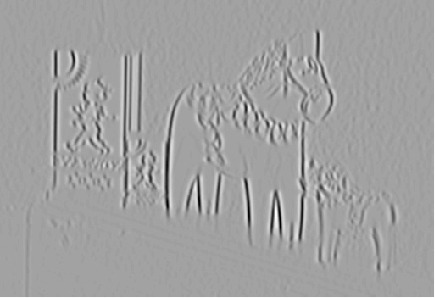}
           & \includegraphics[width=0.16\textwidth]{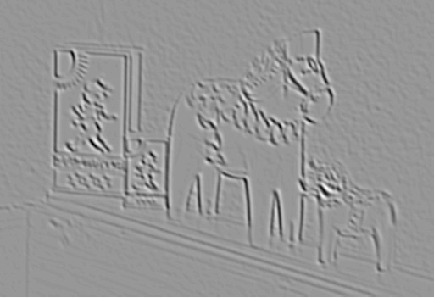}
           & \includegraphics[width=0.16\textwidth]{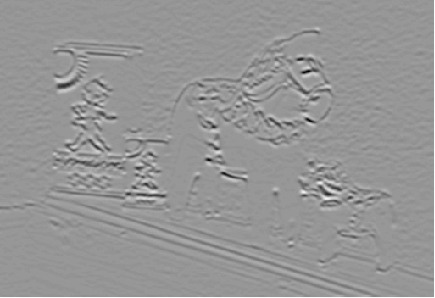}
           & \includegraphics[width=0.16\textwidth]{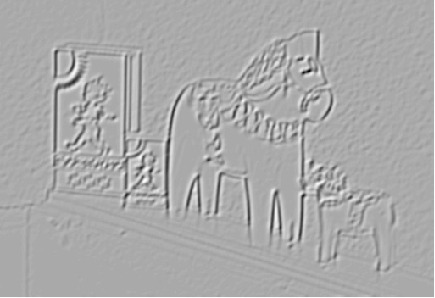} 
           & $\overline{\partial}_{\varphi_2 \varphi_2} {\cal Q}_{\varphi_1} L$ \\
       & \includegraphics[width=0.16\textwidth]{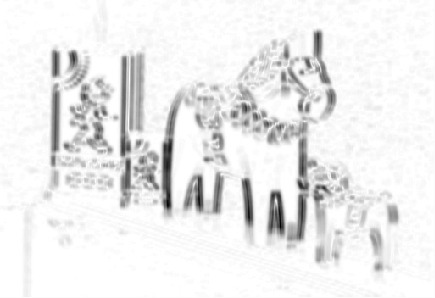}
           & \includegraphics[width=0.16\textwidth]{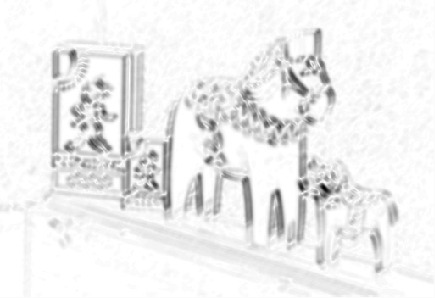}
           & \includegraphics[width=0.16\textwidth]{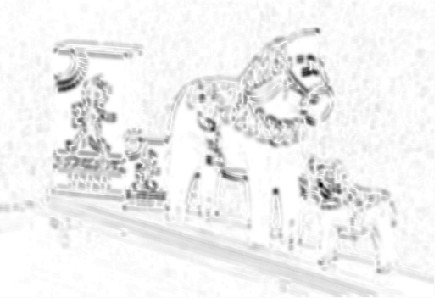}
           & \includegraphics[width=0.16\textwidth]{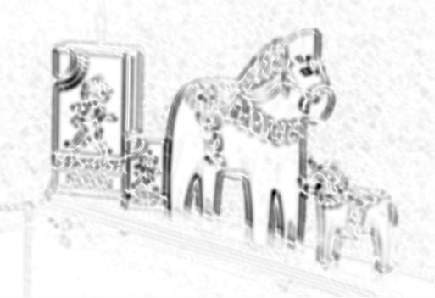} 
           & ${\cal Q}_{\varphi_2} {\cal Q}_{\varphi_1} L$ \\   
           & {\footnotesize\em $\varphi_1 = \tfrac{\pi}{2}$, $\varphi_2 = 0$\/} 
           & {\footnotesize\em $\varphi_1 = \tfrac{\pi}{2}$, $\varphi_2 = \tfrac{\pi}{4}$\/} 
           & {\footnotesize\em $\varphi_1 = \tfrac{\pi}{2}$, $\varphi_2 = \tfrac{\pi}{2}$\/}
           & {\footnotesize\em $\varphi_1 = \tfrac{\pi}{2}$, $\varphi_2 = \tfrac{3 \pi}{4}$\/} \\
       & \includegraphics[width=0.16\textwidth]{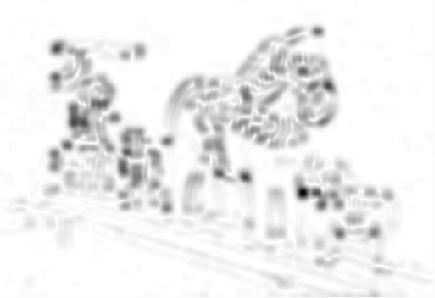}
           & \includegraphics[width=0.16\textwidth]{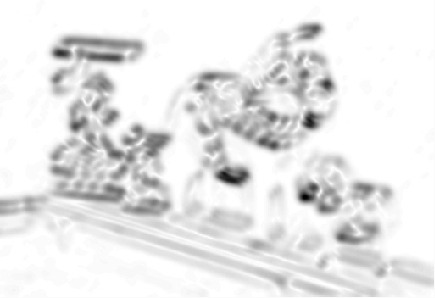}
           & \includegraphics[width=0.16\textwidth]{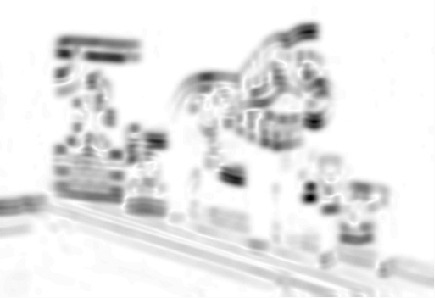}
           & \includegraphics[width=0.16\textwidth]{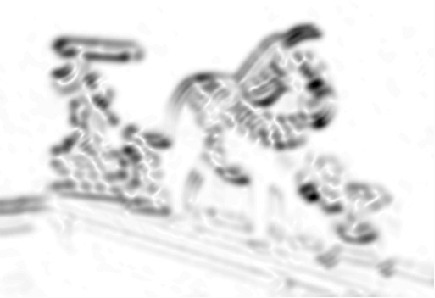} 
           & ${\cal Q}_{\varphi_2} {\cal Q}_{\varphi_1} L$ \\   
       & {\footnotesize\em $\varphi_1 = 0$, $\varphi_3 = 0$\/} 
           & {\footnotesize\em $\varphi_1 = 0$, $\varphi_3 = \tfrac{\pi}{4}$\/} 
           & {\footnotesize\em $\varphi_1 = 0$, $\varphi_3 = \tfrac{\pi}{2}$\/}
           & {\footnotesize\em $\varphi_1 = 0$, $\varphi_3 = \tfrac{3\pi}{4}$\/} \\
       & \includegraphics[width=0.16\textwidth]{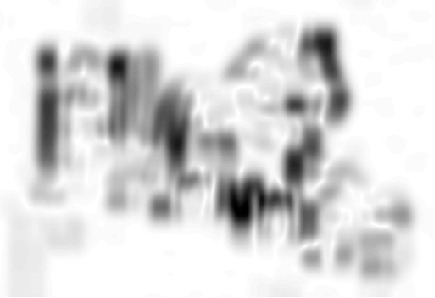}
           & \includegraphics[width=0.16\textwidth]{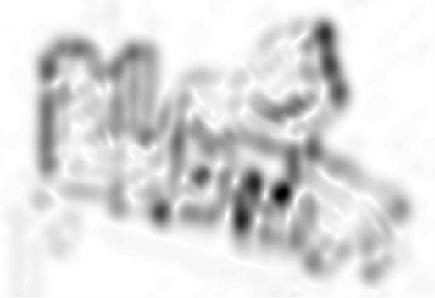}
           & \includegraphics[width=0.16\textwidth]{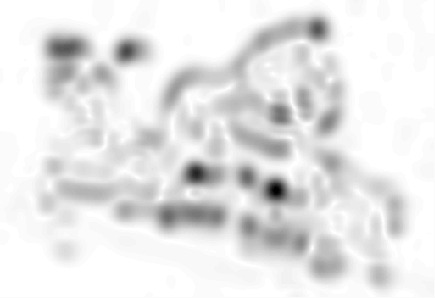}
           & \includegraphics[width=0.16\textwidth]{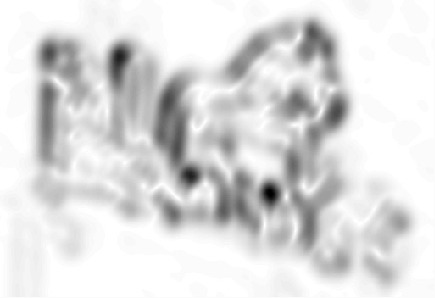} 
           & ${\cal Q}_{\varphi_3} {\cal P}_{\varphi_2} {\cal Q}_{\varphi_2} {\cal Q}_{\varphi_1} L$  \\ 
\end{tabular} 
  \end{center}
  \caption{Subset of features in a hierarchy of directional derivatives and quasi quadrature measures
    computed from an indoor image for different
  combinations of angles in layers 1 and 2 for $s_0 = 2$.
  (top row) Original image and first-order directional derivatives
  $L_{\varphi}$.
  (second row) Second-order directional derivatives
  $L_{\varphi\varphi}$.
  (third row) Oriented quasi quadrature measures ${\cal Q}_{\varphi_1}
  L$ in layer 1.
  (fourth row) First-order directional derivatives computed from
  ${\cal Q}_{\varphi_1} L$ for $\varphi_1 = 0$.
  (fifth row) Second-order directional derivatives computed from
  ${\cal Q}_{\varphi_1} L$ for $\varphi_1 = 0$.
  (sixth row) Oriented quasi quadrature measures 
  ${\cal Q}_{\varphi_2} {\cal Q}_{\varphi_1} L$ in layer 2 
  for $\varphi_1 = 0$.
  (seventh row) Oriented quasi quadrature measures 
  ${\cal Q}_{\varphi_2} {\cal Q}_{\varphi_1} L$ in layer 2 
  for $\varphi_1 = \pi/2$.
  (bottom row) Oriented quasi quadrature measures 
  ${\cal Q}_{\varphi_3} {\cal P}_{\varphi_2} {\cal Q}_{\varphi_2} {\cal Q}_{\varphi_1} L$ in layer 3 
  for $\varphi_1 = 0$.
  (The contrast has been reversed for the quasi quadrature measures so
that high values are shown as dark and low values as bright.) (Image
size: $512 \times 350$ pixels.)}
  \label{fig-dala-quasiquadnet}
\end{figure*}

\subsection{Scale covariance}
\label{sec-sc-cov-proof}

A theoretically attractive property of this family of networks is that
the networks are provably scale covariant. Given two images $f$ and $f'$ that
are related by a uniform scaling transformation,
\begin{equation}
   f'(x') = f(x) \quad\quad \mbox{with} \quad\quad x' = S x
\end{equation}
for some $S > 0$, their corresponding scale-space representations $L$ and
$L'$ will be equal 
\begin{equation}
   L'(x';\; s') = L(x;\; s) 
\end{equation}
and so will the
scale-normalized derivatives 
\begin{equation}
  s'^{n/2} \, L'_{{x_i'}^n}(x';\; s') = s^{n/2} \, L_{x_i^n}(x;\; s)
\end{equation}
based on $\gamma = 1$ if the scale levels are matched according to $s' = S^2 s$ 
\cite[Eqns.~(16) and (20)]{Lin97-IJCV}. 

This implies that if the initial scale levels
$s_0$ and $s_0'$ underlying the construction in 
(\ref{eq-hier-quasi-quad-line1}) and (\ref{eq-hier-quasi-quad-line2})
are related according to $s_0' = S^2 s_0$,
then the
first layers of the feature hierarchy will be related according to \cite[Eqns.~(55) and (63)]{Lin18-SIIMS}
\begin{equation}
   F_1'(x', \varphi_1) = S^{-\Gamma} \, F_1(x, \varphi_1).
\end{equation}
Higher
layers in the feature hierarchy are in turn related according to
\begin{equation}
   F_k'(x', \varphi_1, ..., \varphi_{k-1}, \varphi_k) 
   = S^{-k \Gamma} \, F_k(x, \varphi_1, ..., \varphi_{k-1}, \varphi_k)  
\end{equation}
and are specifically equal if $\Gamma = 0$. This means that it will be
possible to perfectly match such hierarchical representations
under uniform scaling transformations.

\subsection{Rotation covariance} 
\label{sec-rot-cov-proof}

Under a rotation of image space
by an angle $\alpha$, 
\begin{equation}
  f'(x') = f(x) \quad\quad \mbox{with} \quad\quad  x'= R_{\alpha} x,
\end{equation}
the
corresponding feature hierarchies
are in turn equal
if the orientation angles are related according to $\varphi'_i = \varphi_i + \alpha$ ($i = 1..k$)
\begin{equation}
   F_k'(x', \varphi'_1, ..., \varphi'_{k-1}, \varphi'_k) 
   = F_k(x, \varphi_1, ..., \varphi_{k-1}, \varphi_k).
\end{equation}

\subsection{Exact {\em vs.\/} approximate covariance (or invariance) in a practical
  implementation}

The
architecture of the quasi quadrature network has been 
designed to support scale covariance based on image primitives
(receptive fields) that obey the general scale covariance property
(\ref{eq-scale-covariance-prop-Fk}) and to support rotational covariance by an
explicit expansion over image rotations of the form (\ref{eq-hier-quasi-quad-line2}).

\paragraph{Scale covariance.}

The statement about true scale covariance in
Section~\ref{sec-sc-cov-proof} holds in the
  continuous case, provided that we can represent a continuum of scale
  parameters. 

In a practical implementation, it is natural to sample
  this space into a set of discrete scale levels with a constant scale
  ratio between adjacent scale levels. Then, the scale-covariant property
  will be restricted to spatial scaling factors that can be perfectly
  matched between these scale levels. If the scale levels are
  expressed in units of $\sigma = \sqrt{s}$ and if the scale ratio
  between adjacent scale levels in these units is $r$, then exact scale
  covariance will hold for all scaling factors that are integer powers
  of $r$, provided that the image resolution and the image size is
  sufficient to resolve the relevant image structures. For scaling
  factors in between these discrete values, there will be an
  approximation error, which could possibly be reduced by
  a complementary scale interpolation mechanism. 

For a discrete
  implementation with limited image resolution and limited image size, there
will be additional restrictions on how well the discrete
implementation approximates the continuous theory.
For the implementations underlying this paper, we use a scale-space
concept specially designed for discrete signals computed by separable
convolution with the discrete analogue of the Gaussian kernel
$T(n;\; s) = e^{-s} I_n(s)$ \cite{Lin90-PAMI}, which is
defined in terms of the modified Bessel functions of integer order
$I_n(s)$ \cite{AS64}. 
This discrete scale-space concept constitutes a numerical
approximation of the continuous scale-space concept via a spatial
discretization of the diffusion equation, which governs the evolution
properties over scale of the Gaussian scale-space concept.

\paragraph{Rotational covariance.}

The statement about true rotational covariance in
Section~\ref{sec-rot-cov-proof} holds
  provided that we can represent a continuum of rotation angles. 
For a continuum of orientation angles, the summation over image orientations in the pooling stage 
 (\ref{eq-orient-pooling}) should be replaced by an integral over all the
 image orientations to guarantee exact covariance to hold for all rotation angles.

In a practical implementation, it is natural to sample the orientation
  angles on the unit circle into a set of discrete angles with a
  constant increment between. Then, the rotation-covariant property will be
  restricted to the set of discrete rotation angles that are spanned by this
  discretization. For rotation angles in between, there will be an
  approximation error, which could possibly be reduced by a suitable
  interpolation mechanism. 

With regard to a discrete
  implementation, there may be additional deviations in how well the
  discrete approximations of directional derivatives numerically approximate their
continuous counterparts. For the implementation underlying this paper,
we complement the discrete scale-space concept in \cite{Lin90-PAMI}
with discrete derivative approximations with scale-space properties
\cite{Lin93-JMIV}, where small support discrete derivative
approximations $\delta_x = (-1/2, 0, 1/2)$ and $\delta_{xx} = (1, -2,
1)$ are applied to the discrete scale-space smoothed image data and
directional derivative approximations are then computed from the
continuous relationships
(\ref{eq-1st-dir-der-scsp}) and (\ref{eq-2nd-dir-der-scsp}).

\paragraph{Numerical approximation of a truly covariant continuous
  theory.}

By all steps in the discrete implementation constituting numerical approximations
of their corresponding counterparts in the continuous theory, it follows that the discrete implementation
will also numerically approximate the desirable covariance properties (or as an extension
invariance properties) with respect to scaling transformations and
rotations in the image domain.
The accuracy of approximation of the combined system will then be a composed
effect of the numerical accuracy of the different primitives.

\subsection{Experiments}

Figures~\ref{fig-corduroy-quasiquadnet}--\ref{fig-dala-quasiquadnet} 
show examples of computing different layers in such a
quasi quadrature network for two texture images and an indoor image,
with the combinatorial angular expansion for higher layers delimited
at layer $K = 3$.

For the quite regular corduroy image in Figure~\ref{fig-corduroy-quasiquadnet}, 
we can see that we get clear responses to the
stripes in the cloth in layers 1 and 2, with only a minor dominant response in the third layer
corresponding to the slight irregularity in the mid left of the
original image.

For the mixed regular/irregular wool image in Figure~\ref{fig-wool-quasiquadnet}, we get clear responses to
the crochet work in layer 1, with additional clear responses to the
different types of repeated crochet structures in different subparts of the image in layer
2, whereas in layer 3 the main strong response is due to the
intentional overall irregularity in the pattern.

For the indoor scene in Figure~\ref{fig-dala-quasiquadnet}, we can note that 
the responses are strongest
along the edges in the scene for all the layers, with some locally
stronger responses in layers 2 and 3 assumed near corners or
end-stoppings, especially when the orientations of the oriented quasi
quadrature measures at higher levels in the hierarchy are orthogonal
to the orientation of the oriented quasi quadrature measure in the first layer
($\varphi_2 \orth \varphi_1$ or $\varphi_3 \orth \varphi_1$).
For this image, which is not in any way stationary over image
space, we can observe that the spatial structure of the scene can 
be perceived from the pure magnitude responses of the quasi quadrature
measure in layer 3 in the hierarchy.

In these qualitative respects, we can see how the proposed quasi
quadrature hierarchy is able to reflect non-linear hierarchical relations between
image structures over different scales.

\begin{figure*}[hbtp]
 \begin{center}
    \begin{tabular}{cccccc}
        \includegraphics[width=0.15\textwidth]{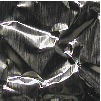} \hspace{-2mm}
        & \includegraphics[width=0.15\textwidth]{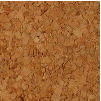} \hspace{-2mm}
        & \includegraphics[width=0.15\textwidth]{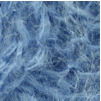} \hspace{-2mm}
        & \includegraphics[width=0.15\textwidth]{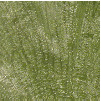} \hspace{-2mm}
        & \includegraphics[width=0.15\textwidth]{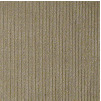} \hspace{-2mm} 
        & \includegraphics[width=0.15\textwidth]{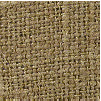} \hspace{-2mm} \\
        \includegraphics[width=0.15\textwidth]{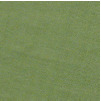} \hspace{-2mm}
        & \includegraphics[width=0.15\textwidth]{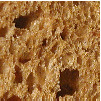} \hspace{-2mm}
        & \includegraphics[width=0.15\textwidth]{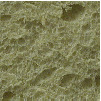} \hspace{-2mm}
        & \includegraphics[width=0.15\textwidth]{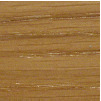} \hspace{-2mm}
        & \includegraphics[width=0.15\textwidth]{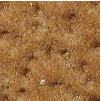} \hspace{-2mm} \\
    \end{tabular} 
  \end{center}
  \caption{Sample images from the KTH-TIPS2b texture dataset
    \cite{KTH-TIPS2}. This dataset consists of images of 11 classes of
    textures with 4 samples from each class. Each sample has been
    photographed from 9 distances leading to 9 relative scales, with
    additionally 12 different pose and illumination conditions for
    each scale, implying a total number of $11 \times 4 \times 9
    \times 12 = 4752$ images.
    This figure shows one sample from each class, with varying scale, pose and illumination
    conditions between the samples. 
    (Most images of size $200 \times 200$ pixels.)}
  \label{fig-kthtips2}
\end{figure*}

\begin{figure*}[hbtp]
 \begin{center}
    \begin{tabular}{cccccccc}
        \includegraphics[width=0.125\textwidth]{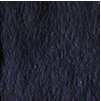} \hspace{-2mm}
        & \includegraphics[width=0.125\textwidth]{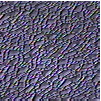} \hspace{-2mm}
        & \includegraphics[width=0.125\textwidth]{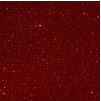} \hspace{-2mm} 
        & \includegraphics[width=0.125\textwidth]{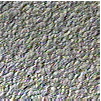} \hspace{-2mm}
        & \includegraphics[width=0.125\textwidth]{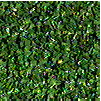} \hspace{-2mm}
        & \includegraphics[width=0.125\textwidth]{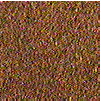} \hspace{-2mm} 
        & \includegraphics[width=0.125\textwidth]{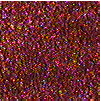} \hspace{-2mm} \\
        \includegraphics[width=0.125\textwidth]{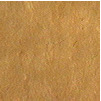} \hspace{-2mm}
        & \includegraphics[width=0.125\textwidth]{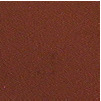} \hspace{-2mm}
        & \includegraphics[width=0.125\textwidth]{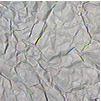} \hspace{-2mm} 
        & \includegraphics[width=0.125\textwidth]{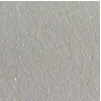} \hspace{-2mm}
        & \includegraphics[width=0.125\textwidth]{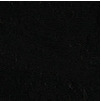} \hspace{-2mm} 
        & \includegraphics[width=0.125\textwidth]{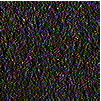} \hspace{-2mm}
        & \includegraphics[width=0.125\textwidth]{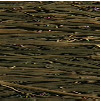} \hspace{-2mm} \\
        \includegraphics[width=0.125\textwidth]{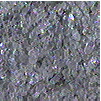} \hspace{-2mm}
        & \includegraphics[width=0.125\textwidth]{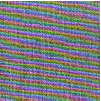} \hspace{-2mm}
        & \includegraphics[width=0.125\textwidth]{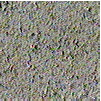} \hspace{-2mm}
        & \includegraphics[width=0.125\textwidth]{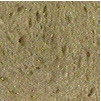}  \hspace{-2mm} 
        & \includegraphics[width=0.125\textwidth]{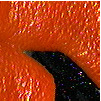} \hspace{-2mm}
        & \includegraphics[width=0.125\textwidth]{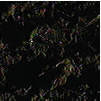} \hspace{-2mm}
        & \includegraphics[width=0.125\textwidth]{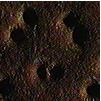} \hspace{-2mm} \\
   \end{tabular} 
  \end{center}
  \caption{Sample images from the CUReT texture dataset \cite{DanGinNayKoe99-TOG}.
    This dataset consists of images of 61 materials, with a single
    sample for each class, and each sample viewed under different
    viewing and illumination conditions. Here, we use the selection of
    92 viewing and illumination conditions chosen in \cite{VarZis09-PAMI} 
    leading to a total number of $61 \times 92 = 5 612$ images.
    This figure shows one image of about every third sample, with varying
    viewing and illumination conditions between the samples.
    (All images of size $200 \times 200$ pixels.)}
  \label{fig-curet}
\end{figure*}

\begin{figure*}[hbtp]
 \begin{center}
    \begin{tabular}{cccccc}
        \includegraphics[width=0.17\textwidth]{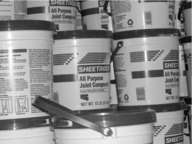} \hspace{-2mm}
        & \includegraphics[width=0.17\textwidth]{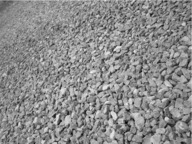} \hspace{-2mm}
        & \includegraphics[width=0.17\textwidth]{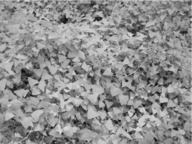} \hspace{-2mm}
        & \includegraphics[width=0.17\textwidth]{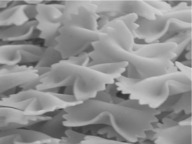} \hspace{-2mm}
        & \includegraphics[width=0.17\textwidth]{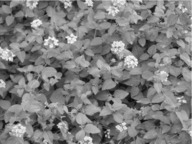} \hspace{-2mm} \\
        \includegraphics[width=0.17\textwidth]{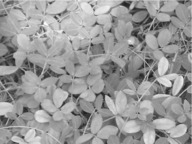} \hspace{-2mm} 
        & \includegraphics[width=0.17\textwidth]{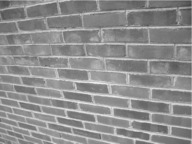} \hspace{-2mm} 
        & \includegraphics[width=0.17\textwidth]{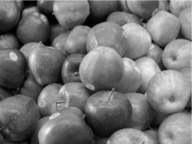} \hspace{-2mm} 
        & \includegraphics[width=0.17\textwidth]{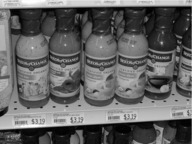} \hspace{-2mm}
        & \includegraphics[width=0.17\textwidth]{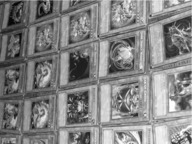} \hspace{-2mm} \\
        \includegraphics[width=0.17\textwidth]{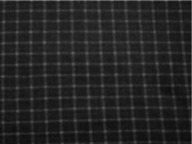} \hspace{-2mm}
        & \includegraphics[width=0.17\textwidth]{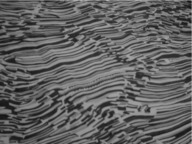} \hspace{-2mm} 
        & \includegraphics[width=0.17\textwidth]{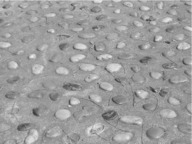} \hspace{-2mm}
        & \includegraphics[width=0.17\textwidth]{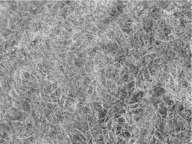} \hspace{-2mm} 
        & \includegraphics[width=0.17\textwidth]{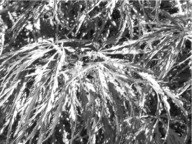} \hspace{-2mm} \\
   \end{tabular} 
  \end{center}
  \caption{Sample images from the UMD texture dataset \cite{XuYanLinJi10-CVPR}.
    This dataset consists of 25 texture classes, with 40 grey-level
    images from each class taken from a variety of different distances and viewing
    directions, thus a total number $25 \times 40 = 1000$ images. 
    This figure shows one sample from each of the first 15 classes.
    (All images of size $1280 \times 960$ pixels.)}
  \label{fig-umd}
\end{figure*}

\section{Application to texture analysis}
\label{sec-appl-texture-anal}

In the following, we will use a substantially reduced version of the
proposed quasi quadrature network for building an application to texture
analysis.

\subsection{Mean-reduced texture descriptors}

If we make the assumption that a spatial texture should obey certain
stationarity properties over image space, we may regard it as reasonable
to construct texture descriptors by accumulating statistics of
feature responses over the image domain, in terms of {\em e.g\/} mean values or histograms.

Inspired by the way the SURF descriptor \cite{BayEssTuyGoo08-CVIU} accumulates mean values and
mean absolute values of derivative responses and the way Bruna and Mallat
\cite{BruMal13-PAMI} and Hadji and
Wildes \cite{HadWil17-ICCV} compute mean values of their hierarchical
feature representations, we will initially explore reducing
the QuasiQuadNet to just the mean values over the image domain of the following 5 features
\begin{equation}
\{ \partial_{\varphi} F_{k}, |\partial_{\varphi}
F_{k}|, \partial_{\varphi\varphi} F_{k}, |\partial_{\varphi\varphi}
F_{k}|, {\cal Q}_{\varphi} F_{k} \}.
\end{equation}
These types of features are computed for all layers in the feature
hierarchy (with $F_0 = L$), which leads to a 4000-D descriptor%
\footnote{With $M = 8$ orientations in image space and 5 basic types
  of features  $\{ \partial_{\varphi} F_{k}, |\partial_{\varphi}
F_{k}|, \partial_{\varphi\varphi} F_{k}, |\partial_{\varphi\varphi}
F_{k}|, {\cal Q}_{\varphi} F_{k} \}$,
  there are $8 \times 5 = 40$ features in layer 1 at a single scale,
$8 \times 8 \times 5 = 320$ features in layer 2 due to the additional
combinatorial expansion in and similar numbers
of $320$ features in layers 3 and 4 due to the limitation on
combinational complexity at layer $K = 3$. For any initial scale level
$\sigma_0$, there are therefore a total number of $40 + 3 \times 320 =
1000$ features. Expanded over 4 initial scale levels $\sigma_0 = \sqrt{s_0} \in \{ 1, 2, 4, 8
\}$, this leads to a total number 4000 feature dimensions, which we
here represent by just their average values over image space.} 
based on $M = 8$ uniformly distributed orientations in $[0, \pi[$, 4
layers in the hierarchy delimited in complexity by directional pooling for $K = 3$
with 4 initial scale levels  $\sigma_0 = \sqrt{s_0} \in \{ 1, 2, 4, 8 \}$.

\begin{table*}
  \begin{center}
  \begin{tabular}{lllccc} 
      \hline
      & KTH-TIPS2b & CUReT & UMD & Feat & Class \\
      \hline
      FV-VGGVD \cite{CimMajVed15-CVPR} (SVM) & 88.2 & 99.0 & 99.9 & L & L \\ 
      FV-VGGM \cite{CimMajVed15-CVPR} (SVM) & 79.9 & 98.7 & 99.9 & L & L \\ 
      MRELBP \cite{LiuLaoFieGuoWanPie16-TIP}  (SVM) & 77.9 & 99.0 & 99.4 & F & L \\ 
      FV-AlexNet \cite{CimMajVed15-CVPR}  (SVM) & 77.9 & 98.4 & 99.7 & L & L \\ 
      {\em mean-reduced QuasiQuadNet LUV (SVM)}  & 78.3 & 98.6 & & F & L \\ 
      {\em mean-reduced QuasiQuadNet grey (SVM)}  & 75.3 & 98.3 & 97.1 & F & L \\ 
      ScatNet \cite{BruMal13-PAMI} (PCA) & 68.9 & 99.7 & 98.4 & F & L \\ 
      MRELBP \cite{LiuLaoFieGuoWanPie16-TIP}  & 69.0 & 97.1 & 98.7 & F & F \\ 
      BRINT \cite{LiuLonFieLaoZha14-TIP}  & 66.7 & 97.0 & 97.4 & F & F \\ 
      MDLBP \cite{SchDos12-ICPR} & 66.5 & 96.9 & 97.3 & F & F \\ 
      {\em mean-reduced QuasiQuadNet LUV (NNC)}  & 72.1 & 94.9 & & F & F \\ 
      {\em mean-reduced QuasiQuadNet grey (NNC)}  & 70.2 & 93.0 & 93.3 & F & F \\ 
      LBP \cite{OjaPieMae02-PAMI} & 62.7 & 97.0 & 96.2 & F & F \\ 
      ScatNet \cite{BruMal13-PAMI} (NNC) & 63.7 & 95.5 & 93.4 & F & F \\ 
      PCANet \cite{ChaJiaGaoLuZenMa15-TIP} (NNC) & 59.4 & 92.0 & 90.5 & L & F \\ 
      RandNet \cite{ChaJiaGaoLuZenMa15-TIP} (NNC) & 56.9 & 90.9 & 90.9 & F & F \\ 
      \hline
   \end{tabular}
   \end{center}
   \caption{Performance results of the mean-reduced QuasiQuadNet in
     comparison with a selection of among the better methods in the
     extensive performance evaluation by Liu {\em et al.\/}\ \cite{LiuFieGuoWanPei17-PattRecogn}
 (our results in slanted font). (The column labelled ``Feat'' states
 whether the image features are fixed (``F'') or learnt (``L''). The
 column labelled ``Class'' states whether the the classification
 criterion is fixed (``F'') or learnt (``L'').)}
   \label{tab-perf-KTH-TIPS2b}
\end{table*}

\begin{figure}[hbtp]
  \begin{center}
     \begin{tabular}{ccc}
         \includegraphics[width=0.29\columnwidth]{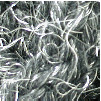}
         & \includegraphics[width=0.29\columnwidth]{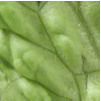}
         & \includegraphics[width=0.29\columnwidth]{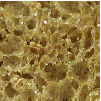} \\
         \includegraphics[width=0.29\columnwidth]{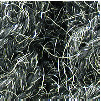}
         & \includegraphics[width=0.29\columnwidth]{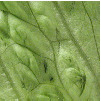}
         & \includegraphics[width=0.29\columnwidth]{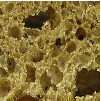} \\
         \includegraphics[width=0.29\columnwidth]{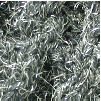}
         & \includegraphics[width=0.29\columnwidth]{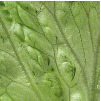}
         & \includegraphics[width=0.29\columnwidth]{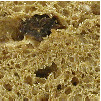} \\
         \includegraphics[width=0.29\columnwidth]{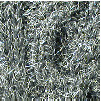}
         & \includegraphics[width=0.29\columnwidth]{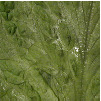}
         & \includegraphics[width=0.29\columnwidth]{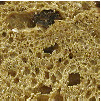} \\
         \includegraphics[width=0.29\columnwidth]{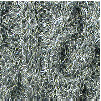}
         & \includegraphics[width=0.29\columnwidth]{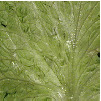}
         & \includegraphics[width=0.29\columnwidth]{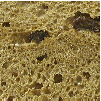} \\
     \end{tabular}
  \end{center}
  \caption{Examples of the scaling variations in the KTH TIPS2 dataset
    for one sample each from the classes ``wool'',
    ``lettuce'' and ``brown bread'' at a subset of five of the scales in the dataset
   (the sizes labelled ``2'', ``4'', ``6'', ``8'' and ``10'' from top
   to bottom.)}
  \label{fig-sc-var-KTH-TIPS2}
\end{figure}

\begin{figure}
  \begin{center}
     \begin{tabular}{c}
        {\footnotesize\em Scaling dependency over scale-matched
       subsets of training/testing data} \\
        \includegraphics[width=0.97\columnwidth]{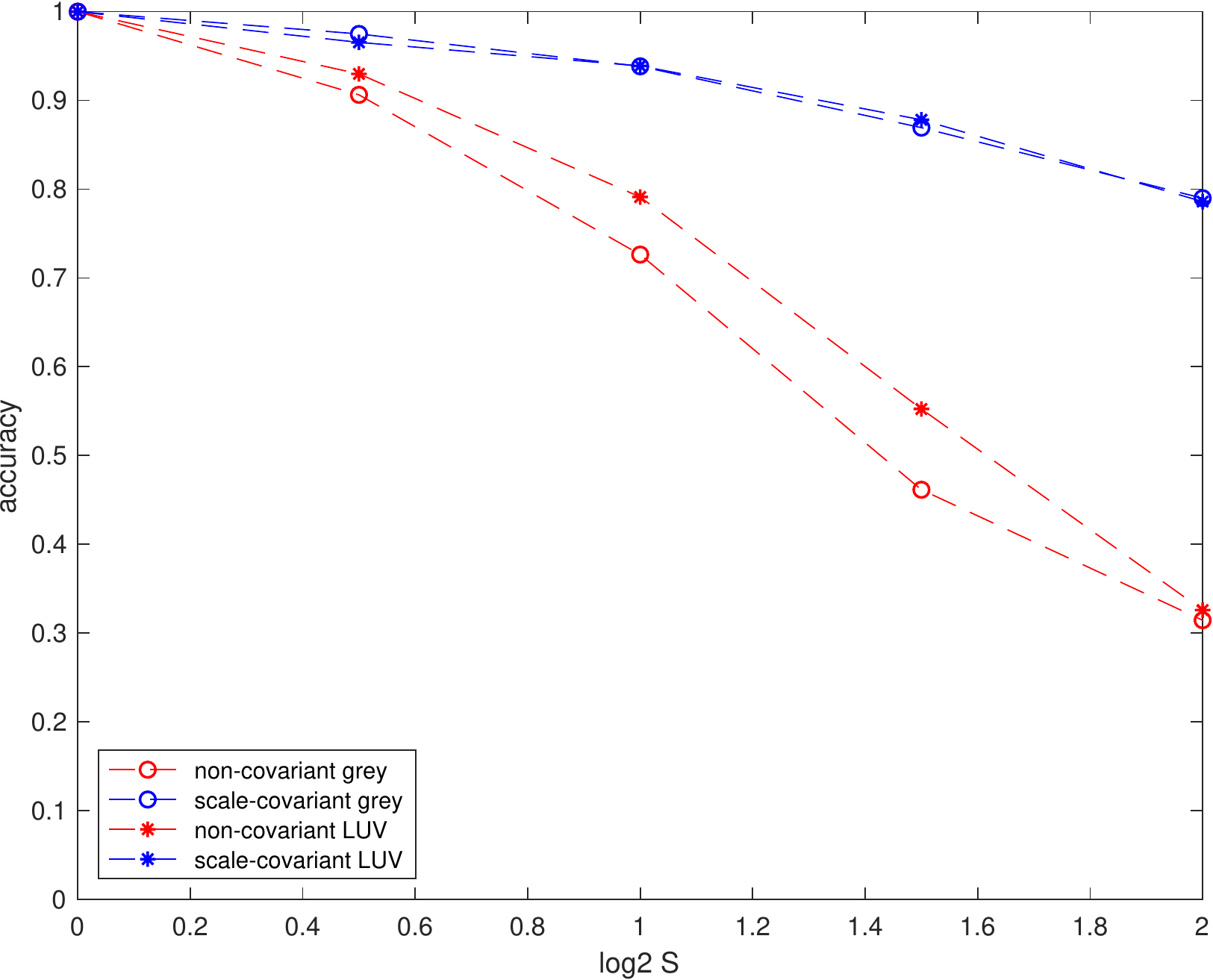}
     \end{tabular}
   \end{center}
   \caption{Comparison between scale-covariant matching {\em vs.\/}\ non-covariant
     matching of texture descriptors on the KTH-TIPS2b dataset \cite{KTH-TIPS2}.
     In the experiments underlying this figure, we have used the scale
   variations in the dataset to perform matching over spatial scaling
   factors of $S = \sqrt{2}$, $2$, $2\sqrt{2}$ and $4$, here
   represented as $\log_2 S$ on the horizontal axis.
   For non-covariant matching, here represented as red curves, we have
 used the same scale parameters for the image descriptors in the training data and the test
 data.
  For scale-covariant matching, here represented as blue curves, we have adapted the scale levels of
  the image descriptors to the known scale factor between the training
data and the test data. As can be seen from the results, in the
presence of substantial scaling variations, the use of scale-covariant
matching, as enabled by the provably scale-covariant networks proposed
in this article, improves the performance substantially if there are
significant scaling variations in the data.
(All these results have been computed with SVM
classification of mean-reduced image descriptors from QuasiQuadNets
computed from either pure grey-level images or colour images. 
The results for the pure grey-level descriptors are indicated by 'o', whereas
the results for the LUV colour descriptors are indicated by '*'.)}
   \label{fig-multi-sc-match-KTH-TIPS2}
\end{figure}

\begin{table}[hbtp]
  \begin{center}
    \begin{tabular}{lrrrr}
     \hline
      S        & $\sqrt{2}$ & $2$ & $2 \sqrt{2}$ & $4$ \\
     \hline
     non-covariant grey & 90.6 & 72.6 & 46.1 & 31.4 \\
     scale-covariant grey & 97.5 & 93.8 & 86.9 & 79.0 \\
     non-covariant LUV & 93.0 & 79.1 & 55.2 & 32.6 \\
     scale-covariant LUV & 96.5 & 93.9 & 87.8 & 78.6 \\
     \hline
     \end{tabular} 
  \end{center}
  \caption{Numerical performance values underlying the graphs in
    Figure~\ref{fig-multi-sc-match-KTH-TIPS2}, which quantify the
    performance of texture classification based on mean-reduced
    texture descriptors from QuasiQuadNets over
    scaling transformations with scaling factors of $\sqrt{2}$, $2$, $2
    \sqrt{2}$ and $4$ for the KTH TIPS2b dataset.}
  \label{tab-multi-sc-match-KTH-TIPS2}
\end{table}

\subsection{Texture classification on the KTH-TIPS2b dataset}
\label{sec-texture-class-kth-tips2}

The second column in Table~\ref{tab-perf-KTH-TIPS2b} shows the result of applying this approach to the
KTH-TIPS2b dataset \cite{KTH-TIPS2} for texture classification, see
Figure~\ref{fig-kthtips2} for sample images from this dataset.
The KTH-TIPS2b dataset contains images of 11 texture classes (``aluminium foil'', ``cork'',
``wool'', ``lettuce leaf'', ``corduroy'', ``linen'', ``cotton'',
``brown bread'', ``white bread'', ``wood'' and ``cracker") with 4
physical samples from each class and photographs of each sample taken from
9 distances leading to 9 relative scales labelled ``2'', \dots, ``10''
over a factor of 4 in scaling transformations and additionally 12 different pose and illumination conditions for each
scale, leading to a total number of $11 \times 4 \times 9 \times 12 = 4752$ images. The regular benchmark setup
implies that the images from 3 samples in each class are used for
training and the remaining sample in each class for testing over 4 permutations. Since
several of the samples from the same class are quite different from each other in appearance, this implies
a non-trivial benchmark which has not yet been saturated.

When using nearest-neighbour classification on the mean-reduced grey-level
descriptor, we get 70.2~\% accuracy, and 72.1~\% accuracy when 
computing corresponding features from the LUV channels of a
colour-opponent representation. When using SVM classification \cite{ChaLin11-TIST}, the
accuracy becomes 75.3~\% and 78.3~\%, respectively. Comparing with the results
of an extensive set of other methods in Liu {\em et al.\/}
\cite{LiuFieGuoWanPei17-PattRecogn}, out of which a selection of
the better results is listed in Table~\ref{tab-perf-KTH-TIPS2b},
the results of the mean-reduced QuasiQuadNet are better than
classical texture classification methods such as local binary patterns (LBP) \cite{OjaPieMae02-PAMI}, 
binary rotation-invariant noise tolerant texture descriptors
\cite{LiuLonFieLaoZha14-TIP}
 and multi-dimensional local binary patterns (MDLBP) \cite{SchDos12-ICPR}  
and also better than
other hand-crafted networks,
such as ScatNet \cite{BruMal13-PAMI}, 
PCANet \cite{ChaJiaGaoLuZenMa15-TIP} and 
RandNet \cite{ChaJiaGaoLuZenMa15-TIP}.
The performance of the mean-reduced QuasiQuadNet descriptor does,
however, not reach the performance of applying SVM classification to
Fischer vectors of the filter output in learned convolutional networks 
(FV-VGGVD, FV-VGGM \cite{CimMajVed15-CVPR}).

By instead performing the training on every second scale in the
dataset (scales ``2'', ``4'', ``6'', ``8'', ``10'') and the testing on the other scales
(``3'', ``5'', ``7'', ``9''), such that the benchmark does not primarily test the
generalization properties between the different very few samples in each class,
the classification performance is 98.8~\% for the
grey-level descriptor and 99.6~\% for the LUV descriptor.

\subsection{Scale-covariant matching of image descriptors on the KTH-TIPS2b dataset}

An attractive property of the KTH-TIPS2 dataset is that we can use the
controlled scaling variations in this dataset (see
Figure~\ref{fig-sc-var-KTH-TIPS2}) to investigate the
influence of scale covariance with respect to image descriptors
defined from a provably scale-covariant network.
To test this property, we constructed partitionings of the dataset into training sets and
test sets with known scaling variations between the data.

The scales in the datasets, which we will henceforth refer to as
sizes, labelled from ``2'' to ``10'', span a
scaling factor of 4, with a relative scaling factor of $\sqrt[4]{2}$
between adjacent sizes. To cover a set of relative scaling factors 
$S \in \{ \sqrt{2}, 2, 2\sqrt{2}, 4 \}$, we partitioned the
dataset and adapted the scale parameters of the QuasiQuadNet
to the relative scaling factors in the following way:
\begin{itemize}
\item
  Relative scaling factor $\sqrt{2}$: Training data at the sizes labelled $\{5, 6, 9, 10\}$
  with image descriptors computed at the scales $\sigma_0 \in \{1, 2, 4, 8\}$.
  Test data at the sizes labelled $\{3, 4, 7, 8\}$
  with image descriptors computed at the scales $\sigma_0 \in \{\sqrt{2}, 2\sqrt{2}, 4\sqrt{2}, 8\sqrt{2}\}$.
\item
  Relative scaling factor $2$: Training data at the sizes labelled $\{7, 8, 9, 10\}$
  with image descriptors computed at the scales $\sigma_0 \in \{1, 2, 4, 8\}$.
  Test data at the sizes labelled $\{3, 4, 5, 6\}$
  with image descriptors computed at the scales $\sigma_0 \in \{2, 4, 8, 16\}$.
\item
  Relative scaling factor $2\sqrt{2}$: Training data at the sizes labelled $\{8, 9, 10\}$
  with image descriptors computed at the scales $\sigma_0 \in \{1, 2, 4, 8\}$.
  Test data at the sizes labelled $\{2, 3, 4\}$
  with image descriptors computed at the scales $\sigma_0 \in \{2\sqrt{2}, 4\sqrt{2}, 8\sqrt{2}, 16\sqrt{2}\}$.
\item
  Relative scaling factor $4$: Training data at the size labelled $\{10\}$
  with image descriptors computed at the scales $\sigma_0 \in \{1, 2, 4, 8\}$.
  Test data at the size labelled $\{2\}$
  with image descriptors computed at scales $\sigma_0 \in \{4, 8, 16, 32\}$.
\end{itemize}
These partitionings between training sets and test sets have thus been
constructed in such a way that there for each image descriptor
computed from an image in the test set should exist a corresponding
scale-matched image descriptor in the training set.

To measure the influence relative to not adapting the scale levels to
scale covariance, we also performed non-covariant classification with
all the image descriptors, both in the training data and the test
data, computed at the scales $\sigma_0 \in \{1, 2, 4, 8\}$.
 
The result of this experiment is shown in
Figure~\ref{fig-multi-sc-match-KTH-TIPS2}, which
shows graphs of how the
accuracy of the texture classification depends on the logarithm of the
relative scaling factor $\log_2 S$ between the training data and
the test data. As can be seen from the graphs, the performance is
substantially higher for scale-covariant classification compared to
non-covariant classification.
Although this task is not influenced by the generalization ability of
the image descriptors, as measured in the regular experimental setup
for the KTH-TIPS2 dataset in the
sense that images from all the samples are here included in both the
training sets and the test sets, there are nevertheless reasons why
the image data cannot be perfectly matched: 
(i) The support regions for the texture descriptors differ in size due to
the scaling transformation, which implies that new image details
appear in one of the images relative to the other (see
Figure~\ref{fig-sc-var-KTH-TIPS2} for an illustration), which in turn
challenges the stationarity assumption underlying the image texture
descriptor, here represented by mean values only.
(ii) The boundary effects at the image boundaries are different
between the two image domains, which in particular affect the image
features at coarser spatial scales.

Notwithstanding these effects, due to the fact that the addition of
new image structures during the scaling transformations leads to a
violation of full scale covariance because of the {\em a priori\/}
delimited image domains in the already given dataset, the primary purpose of this experiment
is to conceptually demonstrate how substantial gains in performance can be obtained by
having a scale-covariant network, and how such scale-covariant
networks are conceptually easier to construct using a continuous model
of the filtering operations in the network. Specifically, scale-space
theory, which underpins this treatment, has been developed to handle
such scaling variations in a theoretically well-founded manner.

\begin{figure}
  \begin{center}
     \begin{tabular}{c}
        {\footnotesize\em Scaling dependency relative to a single reference} \\
        \includegraphics[width=0.97\columnwidth]{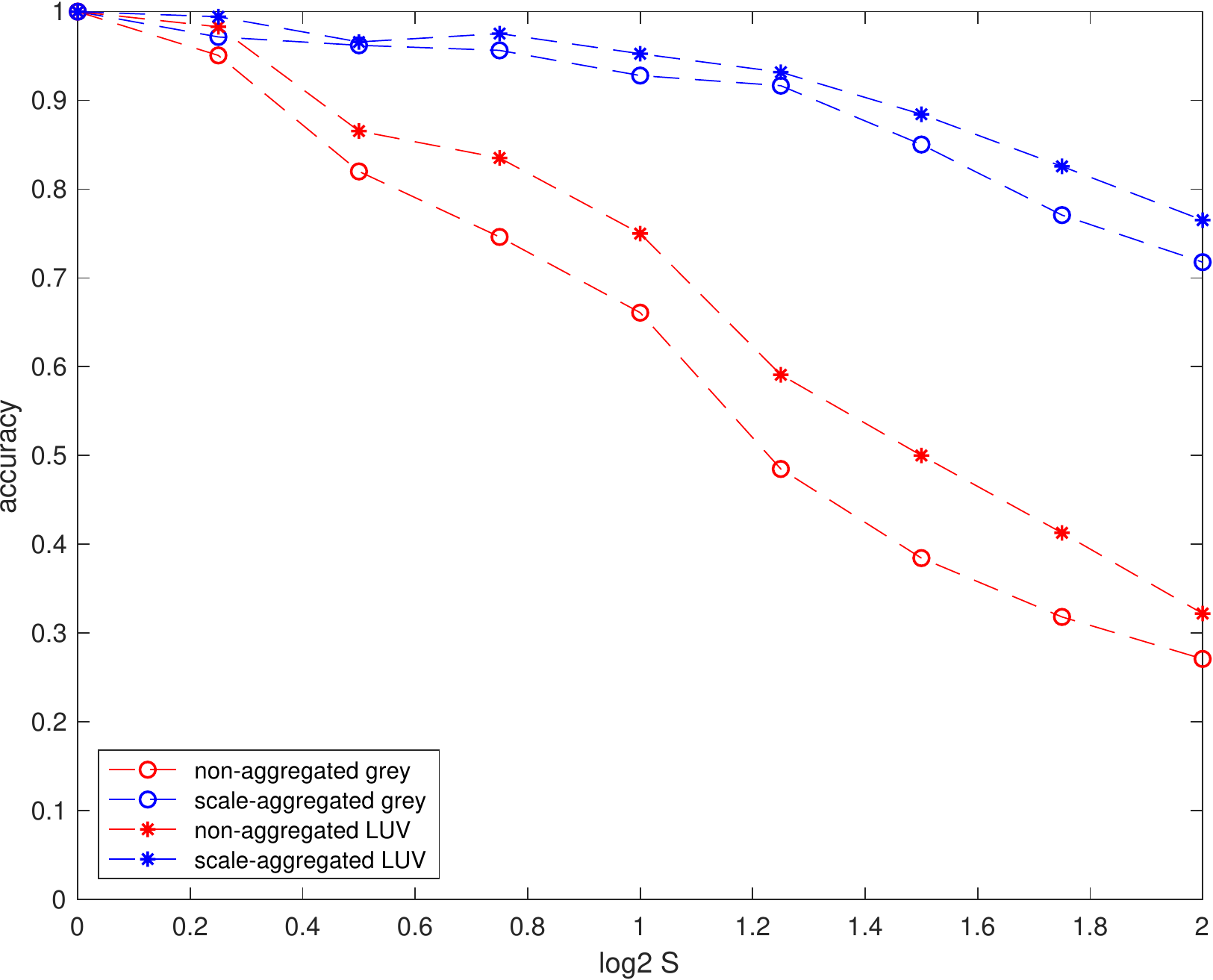} \\
        \medskip\\
        {\footnotesize\em Average scaling dependency over multiple references} \\
        \includegraphics[width=0.97\columnwidth]{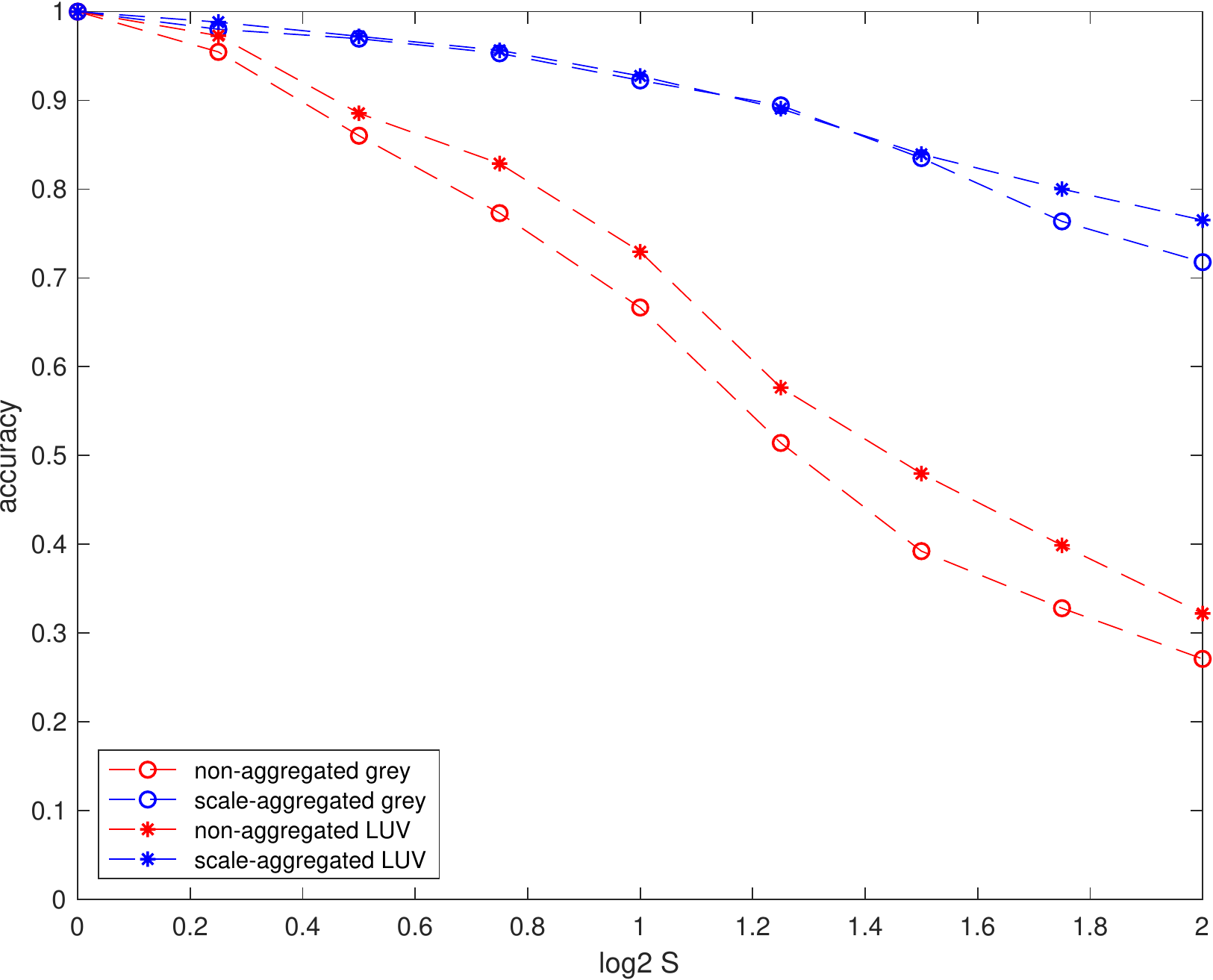}
     \end{tabular}
   \end{center}
   \caption{Comparison between scale-aggregated matching {\em vs.\/}\ non-aggregated
     matching of texture descriptors on the KTH-TIPS2b dataset \cite{KTH-TIPS2}.
     In the experiments underlying this figure, we have used the scale
   variations in the dataset to perform matching over spatial scaling
   factors of $S$ between $\sqrt[4]{2}$ and $4$ in steps of factors of
   $\sqrt[4]{2}$, here
   represented as $\log_2 S$ on the horizontal axis.
   For non-aggregated matching, here represented as red curves, we have
 used the image descriptors at the same single scale $\sigma_0 \in
 \{1, 2, 4, 8\}$  in the training data and the test data.
  For scale-aggregated matching, here represented as blue curves, we
  have extended the training data with image descriptors over the set
  of scale levels $\sigma_0 \in \{1, 2, 4, 8\}$,
$\sigma_0 \in \{\sqrt{2}, 2\sqrt{2}, 4\sqrt{2}, 8\sqrt{2}\}$,
$\sigma_0 \in \{2, 4, 8, 16\}$,
$\sigma_0 \in \{2\sqrt{2}, 4\sqrt{2}, 8\sqrt{2}, 16\sqrt{2}\}$,
$\sigma_0 \in \{4, 8, 16, 32\}$ to span scaling variations up to a
factor of 4 in steps of $\sqrt{2}$. 
In the top figure, the experiments have been made relative to training
data at size ``2'' only, and corresponding test data at the sizes
``3'', ``4'', \dots, ``10''. In the bottom figure, a set of multiple
experiments have been performed with training data at each one of the sizes ``2'',
``3'', \dots ``9'', with testing data for the sizes with greater
number labels. The curve in the bottom figure shows the average value
of all these experiments as averaged over equal relative scaling
factors. As can be seen from the results, in the
presence of substantial scaling variations, the use of scale-aggregated
matching, as enabled by the provably scale-covariant networks proposed
in this article, improves the performance substantially if there are
significant scaling variations in the data.
(All these results have been computed with SVM
classification of mean-reduced image descriptors from QuasiQuadNets
computed from either pure grey-level images or colour images. 
The results for the pure grey-level descriptors are indicated by 'o', whereas
the results for the LUV colour descriptors are indicated by '*'.)}
   \label{fig-scale-aggr-match-KTH-TIPS2}
\end{figure}

\subsection{Matching with scale-aggregated covariant image descriptors
  on the KTH-TIPS2b dataset}

In previous section, we used {\em a priori\/} known information about
the structured amounts of scaling transformations in the KTH TIPS2 dataset for demonstrating
the importance of using scale-covariant image descriptors as opposed
to non-covariant image descriptors in situations where the scaling
transformations are substantial.

A more realistic scenario is that the amount of scaling
transformation between the training data and the test data is not {\em a priori\/} known. A useful approach in
such a situation is to complement the image descriptors in the
training set by scale aggregation, meaning that multiple copies of
image descriptors are computed over some set of scale levels,
to enable scale-covariant matching of the image descriptors in the
sense that for any image descriptor computed from the test set
we should as far as possible increase the likelihood for the
classification scheme to be able to find a corresponding scale-matched image descriptor in the
training set.

To test the scale sensitivity of a composed texture classification scheme to
such a scenario, we computed image descriptors for the training data
at the following scales $\sigma_0 \in \{1, 2, 4, 8\}$,
$\sigma_0 \in \{\sqrt{2}, 2\sqrt{2}, 4\sqrt{2}, 8\sqrt{2}\}$,
$\sigma_0 \in \{2, 4, 8, 16\}$,
$\sigma_0 \in \{2\sqrt{2}, 4\sqrt{2}, 8\sqrt{2}, 16\sqrt{2}\}$,
$\sigma_0 \in \{4, 8, 16, 32\}$
and computed the test data at the single scale
$\sigma_0 \in \{1, 2, 4, 8\}$.
As training data we used the images at the single size $\{2\}$
and as test data the images from a single one of each of the sizes 
$\{3, 4, 5, 6, 7, 8, 9, 10\}$, to study the sensitivity to variations
in scaling transformation in steps of $\sqrt[4]{2}$ between adjacent sizes.

\begin{table*}[hbtp]
  \begin{center}
    \begin{tabular}{lrrrrrrrr}
     \hline
      $S$ & $2^{1/4}$ & $2^{1/2}$ & $2^{3/4}$ & $2^{1}$ & $2^{5/4}$ & $2^{3/2}$ & $2^{7/4}$ & $2^{2}$ \\
     \hline
     non-aggregated grey   & 95.5 & 86.0 & 77.3 & 66.7 & 51.4 & 39.2 & 32.8 & 27.1 \\
     scale-aggregated grey & 98.0 & 97.0 & 95.3 & 92.2 & 89.4 & 83.5 & 76.4 & 71.8 \\
     non-aggregated LUV    & 97.3 & 88.6 & 82.9 & 73.0 & 57.6 & 48.0 & 39.9 & 32.2 \\
     scale-aggregated LUV  & 98.8 & 97.2 & 95.6 & 92.8 & 89.1 & 83.9 & 80.0 & 76.5 \\
     \hline
     \end{tabular} 
  \end{center}
  \caption{Numerical performance values underlying the bottom graphs in
    Figure~\ref{fig-scale-aggr-match-KTH-TIPS2}, which quantify the
    performance of texture classification based on mean-reduced
    texture descriptors from QuasiQuadNets over
    scaling transformations with different scaling factors $S$.}
  \label{tab-multi-sc-match-KTH-TIPS2}
\end{table*}

The result of this experiment is shown in top figure in
Figure~\ref{fig-scale-aggr-match-KTH-TIPS2}, which shows graphs of how
the accuracy of the texture classification depends on the logarithm of
the relative scaling factor $\log_2 S$ between the training data and
the test data. 
In the top figure, the experiments have been made relative to training
data the single size ``2'' only, and corresponding test data for each
one of the sizes ``3'', ``4'', \dots, ``10'' in the dataset.
In the bottom figure, the average result of a set of more
extensive experiments is shown, where each one of the sizes ``2'',
``3'', \dots ``9'' has been used for defining scale-aggregated
training data and the testing data has then been taken from a single size
with number label greater than the label for the training data.
The graphs in the bottom figure show the average values over all those
graphs for equal relative scaling factors between the training data
and the test data. As can be seen from the graphs, the performance is
substantially higher for scale-aggregated matching compared to
non-aggregated matching. In this way, the experiment demonstrates how 
the use of a scale-covariant network enables significantly better performance in situations
where there are substantial scaling transformations in the test data
that are not spanned by corresponding scaling variations in the
training data. 

A similar way of handling scale variations between
training data and test data by computing the image descriptors over a
range of scales has also been used for texture classification by Crosier and Griffin
\cite{CroGri10-IJCV}.%
\footnote{The approach by Crosier and Griffin
\cite{CroGri10-IJCV} is also scale covariant, however, not
hierarchical or deep. They do furthermore not test for scale
prediction or scale generalization over large scaling factors as we
target here.}
This type of scale matching constitutes an integrated part of the
scale-space methodology for relating image
descriptors computed from image structures that have been subject to
scaling transformations in the image domain.
Here, we extend this approach for scale generalization to
hierarchical or deep networks, where the scale covariance property of
our networks makes such scale matching possible.

\subsection{Texture classification on the CUReT dataset}

The third column in Table~\ref{tab-perf-KTH-TIPS2b} shows the result
of applying a similar texture classification approach as was used in
Section~\ref{sec-texture-class-kth-tips2} to the
CUReT texture dataset \cite{DanGinNayKoe99-TOG}, see
Figure~\ref{fig-curet} for sample images from this dataset.
The CUReT dataset consists of images of 61 materials, with a single sample for
each material, and each sample viewed under 205 different viewing and illumination
conditions.
For our experiments, we use the selection of 92 cropped images of size
$200 \times 200$ pixels chosen in
\cite{VarZis09-PAMI} from the criterion that a sufficiently large region
of texture should be visible for all the materials.
This implies a total number of $61 \times 92 = 5 612$ images.
Following the standard for this dataset, we measure the average value of
a set of random partitionings into training and testing data of equal size.

With SVM classification on the mean-reduced QuasiQuadNet,
we get $98.3~\%$ accuracy for the grey-level descriptor and 
$98.6~\%$ for the colour descriptor.
This performance is better than hand-crafted 
PCANet \cite{ChaJiaGaoLuZenMa15-TIP} 
and RandNet \cite{ChaJiaGaoLuZenMa15-TIP}
and better than some pure texture descriptors such as 
local binary patterns \cite{OjaPieMae02-PAMI},
multi-dimensional local binary patterns (MDLBP) \cite{SchDos12-ICPR},
binary rotation-invariant noise tolerant texture descriptors
\cite{LiuLonFieLaoZha14-TIP} and near the learned networks
FV-AlexNet and FV-VGGM \cite{CimMajVed15-CVPR}.
For this dataset, the hand-crafted ScatNet \cite{BruMal13-PAMI} does,
however, perform better and so do the learned networks 
FV-VGGVD \cite{CimMajVed15-CVPR} and 
median robust extended local binary patterns \cite{LiuLaoFieGuoWanPie16-TIP}.

\subsection{Texture classification on the UMD dataset}

The fourth column in Table~\ref{tab-perf-KTH-TIPS2b} shows the result
of applying a similar texture classification approach to the
UMD texture dataset \cite{XuYanLinJi10-CVPR}, see Figure~\ref{fig-umd}
for sample images from this dataset.
The UMD dataset consists of 25 texture classes with 40 grey-level images of size
$1280 \times 900$ pixels from each class, taken from different
distances and viewpoints, thus a total number of $25 \times 40 = 1000$
images.
Following the standard for this dataset, we measure the average of
random partitions in training and testing data of equal size.
When using the same scale levels $\sigma_0 \in \{1, 2, 4, 8\}$ 
for the training data and the test data, we get $97.1~\%$ accuracy of 
our mean-reduced grey-level descriptor, which is
better than local binary patterns \cite{OjaPieMae02-PAMI},
PCA\-Net \cite{ChaJiaGaoLuZenMa15-TIP} and RandNet
\cite{ChaJiaGaoLuZenMa15-TIP}.

Noting that this dataset contains significant unstructured scaling
variations, which are not taken into account when computing all the
image descriptors at the same scale,
we also did an experiment with scale-covariant matching, where we expanded the training data
to the following scale combinations
$\sigma_0 \in \{1, 2, 4, 8\}$,
$\sigma_0 \in \{\sqrt{2}, 2\sqrt{2}, 4\sqrt{2}, 8\sqrt{2}\}$,
$\sigma_0 \in \{2, 4, 8, 16\}$,
$\sigma_0 \in \{2\sqrt{2}, 4\sqrt{2}, 8\sqrt{2}, 16\sqrt{2}\}$,
$\sigma_0 \in \{4, 8, 16, 32\}$
and computed the test data at the single scale
$\sigma_0 \in \{2, 4, 8, 16\}$.
The intention behind this data aggregation over scales is to make it easier to
find a match between the training data and the test data over
situations where there are significant scaling transformations between
the training data and the test data, with specifically a lack of
matching training data at a similar scale as for a given test data.
Then, the performance increased from $93.3~\%$ to $95.9~\%$ using NN classification and
from $97.1~\%$ to $98.1~\%$ using SVM classification on the UMD dataset.

A corresponding expansion of the training data to cyclic permutations
over the underlying angles in the image descriptors in the training
data, to achieve rotation-covariant matching, did, however, not improve the results.

\section{Summary and discussion}
\label{sec-summ-disc}

We have presented a theory for defining hand-crafted or structured hierarchical
networks by combining linear and non-linear scale-space operations 
in cascade. After presenting a general sufficiency condition to
construct networks based on continuous scale-space operations
that guarantee provable scale covariance, we have then in more detail developed
one specific example of such a network constructed by
applying quasi quadrature responses of first- and
second-order directional Gaussian derivatives in cascade.

A main purpose behind this study has been to investigate if we could
start building a bridge between the well-founded theory of scale-space
representation and the recent empirical developments in deep learning, while
at the same time being inspired by biological vision.
The present work is intended as initial work in this direction,
where we propose the family of quasi quadrature networks as a new
baseline for hand-crafted networks with associated provable covariance properties
under scaling and rotation transformations.

Specifically, by constructing the network from linear and non-linear
filters defined over a continuous domain, we avoid the restriction
to discrete $3 \times 3$ or $5 \times 5$ filters in most current
deep net approaches, which implies an implicit assumption about
a preferred scale in the data, as defined by the grid spacing 
in the deep net. If the input data to the
deep net are rescaled by external factors, such as from varying
the distance between an observed object and the observer,
the lack of true scale covariance as arising from such preferred
scales in the network implies that the non-linearities in the deep net
may affect the data in different ways, depending on the size of a
projected object in the image domain.

By early experiments with a substantially mean-reduced representation of
our provably scale-covariant QuasiQuadNet, we have demonstrated that it is possible 
to get quite promising performance on texture classification,
and comparable or better than other hand-crafted networks,
although not reaching the performance of 
applying more refined statistical classification methods on 
learned CNNs.

By inspection of the full non-reduced feature maps, we have also observed
that some representations in higher layers may respond to irregularities in
regular textures (defect detection) or corners or end-stoppings in
regular scenes.

Concerning extensions of the approach with quasi quadrature networks,
we propose to:
\begin{itemize}
\item
  relax the restriction to isotropic covariance matrices with $\Sigma
  = I$ in
  Section~\ref{sec-hier-quasi-quad-net} to construct hierarchical
  networks based on more general affine quasi quadrature measures based
  on affine Gaussian derivatives that are computed with varying eccentricities of
  the underlying affine Gaussian kernel to enable affine covariance,
  which will then also enable affine invariance,
\item
  complement the computation of quasi quadrature responses by a
  mechanism for divisive
  normalization \cite{CarHee12-NatureRevNeuroSci} to enforce a
  competition between multiple feature responses and thus increase the
  selectivity of the image features,
\item
  explore the spatial relationships in the full feature maps that
  are suppressed in the mean-reduced representation to make it possible
  for the resulting image descriptors to encode hierarchical relations
  between image features over multiple positions in the image domain and
\item
  incorporate learning mechanisms into the representation.
\end{itemize}
Specifically, it would be interesting to formulate learning mechanisms
that can learn the parameters of a parameterized model for divisive
normalization and to formulate learning mechanisms that can combine
quasi quadrature responses over different positions in the image
domain to support more general object recognition mechanisms than
those that can be supported by a stationarity assumption as explored
in the prototype application to texture classification developed in Section~\ref{sec-appl-texture-anal}.

For the specific application to texture classification in this work,
it does also seem possible that using more advanced statistical
classification methods on the QuasiQuadNet, such as Fischer vectors,
could lead to gains in performance compared to the mean-reduced
representation that we used here, based on just the mean values and
the mean absolute values of the filter responses in our hierarchical
representation.

Concerning more general developments, the general arguments about
scale-covariant continuous networks in
Section~\ref{sec-sc-cov-cont-hier-nets} open up for studying wider
classes of continuous hierarchical networks that guarantee
provable scale covariance. 
We plan to study such extensions in future work.

\section*{Acknowledgements}

I would like to thank Ylva Jansson and the anonymous reviewers for valuable comments that improved
the presentation of some of the topics in the article.

\vspace{-4mm}

$\,$ 

{\footnotesize
\bibliographystyle{splncs}
\bibliography{defs,tlmac}}

\end{document}